\documentclass[10pt,twocolumn,letterpaper]{article}
\newcommand{\real}[1]{\mathbb{R}^{#1}}
\usepackage[accsupp]{axessibility}
\usepackage{bm}
\usepackage{float}
\usepackage{graphicx}
\usepackage{multicol}
\usepackage{multirow}
\usepackage{acro}
\usepackage{setspace}
\usepackage[pagenumbers]{cvpr} 

\usepackage[symbol]{footmisc}

\DeclareAcroEnding{possessive}{'s}{'s}
\NewAcroCommand\acg{m}{\acropossessive\UseAcroTemplate{first}{#1}}
\NewAcroCommand\acsg{m}{\acropossessive\UseAcroTemplate{short}{#1}}
\NewAcroCommand\aclg{m}{\acropossessive\UseAcroTemplate{long}{#1}}

\DeclareAcronym{sota}{
    long=state-of-the-art,
    short=SOTA,
}
\DeclareAcronym{vit}{
    long=Vision Transformer,
    short=ViT
}
\DeclareAcronym{ctp}{
    long=CTransPath,
    short=CTP
}
\DeclareAcronym{wsi}{
    long=whole-slide image,
    short=WSI
}
\DeclareAcronym{mil}{
    long=multiple-instance learning,
    short=MIL
}
\DeclareAcronym{HE}{
    long=hematoxylin and eosin,
    short=H\&E
}
\DeclareAcronym{srcl}{
    long=semantically-relevant contrastive learning,
    short=SRCL
}
\DeclareAcronym{tcga}{
    long=The Cancer Genome Atlas,
    short=TCGA
}
\DeclareAcronym{moco}{
    long=momentum contrast,
    short=MoCo
}
\DeclareAcronym{cptac}{
    long=Clinical Protemic Tumor Analysis Consortium,
    short=CPTAC
}
\DeclareAcronym{brca}{
    long=breast cancer,
    short=BRCA
}
\DeclareAcronym{mpp}{
    long=microns per pixel,
    short=MPP
}
\DeclareAcronym{auroc}{
    long=area under the receiver operating characteristic,
    short=AUC
}
\DeclareAcronym{auprc}{
    long=area under the precision recall characteristic,
    short=AUPRC
}
\DeclareAcronym{ssl}{
    long=self-supervised learning,
    short=SSL
}
\DeclareAcronym{umap}{
    long=uniform manifold approximation and projection,
    short=UMAP
}
\DeclareAcronym{cobra}{
    long=COntrastive Biomarker Representation Alignment,
    short=\textsc{Cobra}
}
\DeclareAcronym{fm}{
    long=foundation model,
    short=FM
}
\DeclareAcronym{cpath}{
    long=Computational Pathology,
    short=CPath
}
\DeclareAcronym{IHC}{
    long=immunohistochemistry,
    short=IHC
}
\DeclareAcronym{ssd}{
    long=state space dual,
    short=SSD
}
\definecolor{cvprblue}{rgb}{0.21,0.49,0.74}
\usepackage[pagebackref,breaklinks,colorlinks,allcolors=cvprblue]{hyperref}

\def\paperID{16125} 
\def\confName{CVPR}
\def\confYear{2025}

\title{Unsupervised Foundation Model-Agnostic Slide-Level Representation Learning}

\author{Tim Lenz$^{1}$\,\thanks{Equal contribution}\;, Peter Neidlinger$^{1\:*}$, Marta Ligero$^{1}$, Georg Wölflein$^{1,2}$, Marko van Treeck$^{1}$, \\ Jakob N.\ Kather$^{1,3,4}$ \\
${^1}$EKFZ for Digital Health TU Dresden, 
${^2}$University of St Andrews, \\
${^3}$Heidelberg University Hospital, ${^4}$University Hospital Dresden  \\
{\tt\small \{tim.lenz,peter.neidlinger,jakob\_nikolas.kather\}@tu-dresden.de}
}

\begin{document}
\maketitle
\begin{abstract}
Representation learning of pathology whole-slide images (WSIs) has primarily relied on weak supervision with Multiple Instance Learning (MIL). 
This approach leads to slide representations highly tailored to a specific clinical task. 
Self-supervised learning (SSL) has been successfully applied to train histopathology foundation models (FMs) for patch embedding generation.
However, generating patient or slide level embeddings remains challenging. 
Existing approaches for slide representation learning extend the principles of SSL from patch level learning to entire slides by aligning different augmentations of the slide or by utilizing multimodal data.
By integrating tile embeddings from multiple FMs, we propose a new single modality SSL method in feature space that generates useful slide representations.
Our contrastive pretraining strategy, called \textsc{Cobra}, employs multiple FMs and an architecture based on Mamba-2. \textsc{Cobra} exceeds performance of state-of-the-art slide encoders on four different public \ac{cptac} cohorts on average by at least $+4.4\%$ AUC, despite only being pretrained on 3048 WSIs from \ac{tcga}. Additionally, \textsc{Cobra} is readily compatible at inference time with previously unseen feature extractors. Code available at \href{https://github.com/KatherLab/COBRA}{https://github.com/KatherLab/COBRA}.
\end{abstract}
\section{Introduction}
\label{sec:intro}
\begin{figure*}[h!]
  \centering
  \begin{subfigure}{0.9\linewidth}
    \centering
    \includegraphics[width=0.95\linewidth]{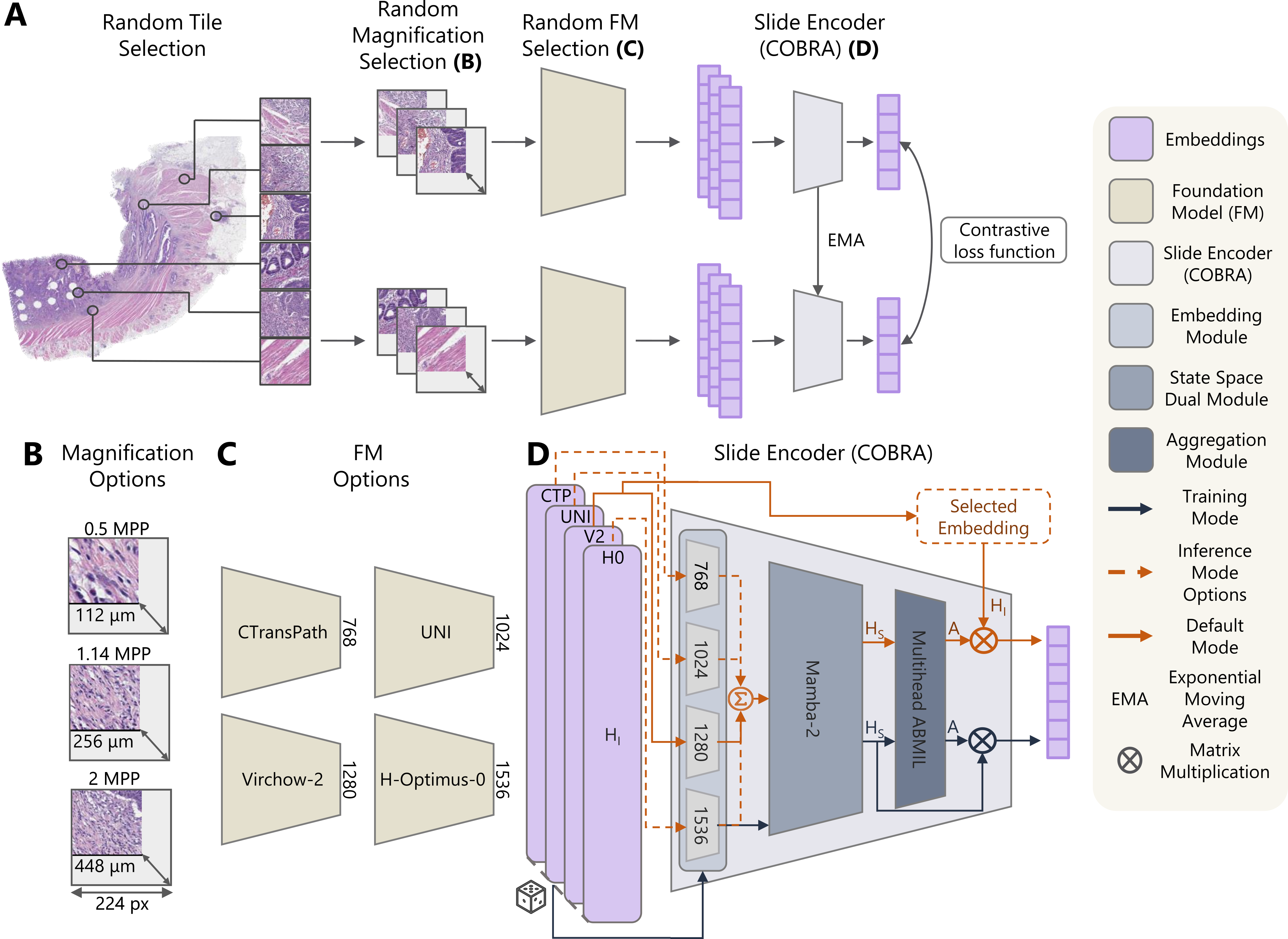}
  \end{subfigure}
  \caption{%
    \textbf{\Acs{cobra} overview} for self-supervised slide representation learning (\textbf{A}). 
    A \acs{wsi} is tessellated into patches at different magnifications (\textbf{B}) and encoded using different \acp{fm} (\textbf{C}) to produce tile embeddings.
    The magnifications (\textbf{B}) and \acp{fm} (\textbf{C}) serve as feature space augmentations to pretrain the \acs{cobra} slide encoder (\textbf{D}) using contrastive self-supervised learning.
  }
  \label{fig:cobra-ov}
  \hfill
\end{figure*}
In recent years, \ac{ssl} has emerged as a foundational approach in \ac{cpath}, providing the basis for weakly supervised models to achieve remarkable results in diagnostic, prognostic, and treatment response prediction tasks~\cite{Chen2022HIPT,wang2022ctp,ElNahhas2024regression,unger2024systematic,Kang2023benchmarking,Mukashyaka2024sampler,chen2024uni,ElNahhas2024stamp,vorontsov2024virchow,xu2024gigapath,lu2024conch,Filiot2023phikon,zimmermann2024virchow2,hoptimus0,nechaev2024hibou}. By capturing informative, low-dimensional representations from unannotated \acp{wsi}, \ac{ssl} has enabled weakly supervised models to use these features for downstream tasks, effectively bridging the gap between high-resolution data and the limited availability of fully annotated datasets. \Ac{ssl} excels in generating low-dimensional feature representations for gigapixel \acp{wsi}, which can reach dimensions of $\text{150,000} \times \text{150,000}$ pixels (px), making them challenging to process with \acp{vit}~\cite{dosovitskiy2021vit} due to memory constraints. Consequently, most \ac{cpath} approaches tessellate \acp{wsi} into smaller patches and extract low-dimensional embeddings for these patches using pretrained histopathology \acp{fm}~\cite{lu2021clam}. Typically, these patch embeddings are used in weakly-supervised models for downstream classification tasks via \ac{mil}~\cite{DIETTERICH1997milproblem, Ilse2018Abmil, shao2021transmil}. 

In addition to patch-based representations, \ac{ssl} can also generate slide-level embeddings without any human annotations~\cite{koohbanani2020selfpath, Lazard2023GigaSSL, yu2023slpd}. Pretrained \ac{ssl} models can be leveraged to achieve impressive results on downstream tasks with minimal labeled data for task-specific fine-tuning, offering practical advantages like reduced labeling costs, elimination of noisy labels inherent to inter-observer variability, and improved generalizability through label-free representations. Central to \ac{ssl} is the alignment of multiple representations of \acp{wsi} or related modalities (e.g., morphological text descriptions) into a shared latent space using contrastive learning or other similarity-based pretraining methods. However, generating effective augmentations to create these representations remains challenging. While image-level augmentations have been widely explored for patch-based learning, they may fail to produce diverse feature augmentations, as many modern FMs are designed to be invariant to these transformations~\cite{wölflein2024benchmarkingpathologyfeatureextractors,neidlinger2024benchmarkingfoundationmodelsfeature}. Other approaches, such as using different stainings (e.g., \ac{HE} combined with \ac{IHC}), have shown potential but are limited by the availability of multi-stained tissue samples~\cite{jaume2024madeleine}. Similarly, aligning multiple modalities, such as text or gene expression data, has produced promising results but is constrained by the limited availability of such datasets and requires additional compute to process the different modalities~\cite{shaikovski2024prism,Wang2024Chief,jaume2024tangle}.

To address these challenges, we propose a novel \ac{ssl} method for image-only slide representation learning called \ac{cobra}. \Ac{cobra} integrates tile embeddings from multiple FMs to generate augmentations directly in feature space, which can then be used to train a slide- or patient-level encoder. By employing Mamba-2~\cite{dao2024mamba2} followed by multi-head gated attention~\cite{jaume2024madeleine} and a contrastive loss objective, \ac{cobra} produces robust slide-level embeddings.
Our contributions are summarized as follows: 
\begin{itemize}
    \item We propose an unsupervised single-modality contrastive slide encoder framework (\ac{cobra}) that avoids the need for stochastic image augmentations as it is trained and deployed on frozen patch embeddings.
    Extensive evaluations across 15 downstream classification tasks on three tissue types with external validation demonstrate \acg{cobra} superiority over existing slide encoders.
    \item Our patient level encoder produces \ac{sota} unsupervised slide representations with unprecedented data efficiency, outperforming existing approaches with only a fraction of the pretaining data (3048 \acp{wsi} across four tissue types).
    \item We show that \ac{cobra} can turn patch level \acp{fm}, including ones not encountered during training, into better slide level feature extractors \emph{without any additional finetuning}, making it particularly valuable as new \acp{fm} emerge. 
    \item \Ac{cobra} can be deployed across different \ac{wsi} magnifications, where lower magnifications yield significant gains in computational efficiency with minimal sacrifice of downstream classification performance. 
\end{itemize}

\section{Related work}
\label{sec:rel_work}

\paragraph{Patch representation learning}
Most works applying \ac{ssl} focus on creating embeddings from image patches. Training a \ac{vit} with an \ac{ssl} method like Dino-v2~\cite{oquab2024dinov2learningrobustvisual} is now the preferred approach for learning task-agnostic image representations in \ac{cpath}. \ac{sota} \acp{fm} usually combine alignment- and reconstruction-based objectives trained with a student-teacher learning paradigm. 
These \acp{fm} are trained on increasingly large datasets and architectures (e.g. \ac{vit}-Giant~\cite{xu2024gigapath} or trained on up to 3M \acp{wsi}~\cite{zimmermann2024virchow2}). Besides image-only \acp{fm}, vision-language pathology \acp{fm} have recently emerged which rely on large-scale paired data~\cite{huang2023plip,lu2024conch}.
\paragraph{Multiple instance learning}
The \ac{sota} approach for \ac{wsi} classification is generating tile embeddings using \acp{fm} and then using these embeddings in a \ac{mil} approach to train an aggregator model for a specific downstream task. In particular, Attention-based MIL (ABMIL)~\cite{Ilse2018Abmil} and many extensions thereof have been proposed~\cite{lu2021clam,Wagner2023,ElNahhas2024stamp,shao2021transmil,yang2024mambamil}. While MIL approaches are prevalent for WSI classification, they are typically supervised and tailored to specific tasks.

\paragraph{Slide representation learning}

In contrast to \ac{mil}, slide representation learning constructs embeddings in an unsupervised manner and is task-agnostic. This next frontier in representation learning of histology images has been proposed in several works. In early work, Chen et al.\ proposed a hierarchical self-distillation approach for learning unsupervised \ac{wsi}-level representations~\cite{Chen2022HIPT}.
Lazard et al.\ used augmented patches to create many embeddings of the same input image to enable contrastive learning with slide embeddings~\cite{Lazard2023GigaSSL}.
In GigaPath, Xu et al.\ trained a masked autoencoder on the embeddings of their patch encoder to obtain slide representations~\cite{xu2024gigapath}.
More recent work applied vast amounts of multimodal data to pretrain aggregation models~\cite{shaikovski2024prism,Wang2024Chief,jaume2024madeleine}. 
Differing from previous methodologies, we achieve state-of-the-art \ac{wsi}-patient-level encoding by performing self-supervised contrastive learning on frozen vision features with a fraction of the data volume. 
None of the mentioned studies used less than 10K \acp{wsi} for WSI-level encoder pretraining~\cite{Chen2022HIPT,Lazard2023GigaSSL,jaume2024madeleine,Wang2024Chief}, while PRISM~\cite{shaikovski2024prism} and Gigapath~\cite{xu2024gigapath} were trained on over 100K WSIs. \ac{cobra} surpasses the performance of earlier work, even though it is trained on only 3K publicly available WSIs~(see \autoref{tab:se-overview}).
\begin{table}
\centering
\caption{\textbf{Slide encoder overview}. Abbreviations are as follows: \#~Ps refers to the number of parameter and \#~WSI refers to the number of \acp{wsi} the slide encoder was pretrained on.}
\label{tab:se-overview}
\resizebox{0.47\textwidth}{!}{%
\begin{tabular}{c|c|c|c}
\toprule
\textbf{Model} & \textbf{\# Ps}[M] & \textbf{\# WSI}[K] & \textbf{Patch FM} \\
\midrule
Gigapath-SE~\cite{xu2024gigapath}   & 86 & 171 & Gigapath~\cite{xu2024gigapath} \\
\hline
CHIEF~\cite{Wang2024Chief}      & 1 & 60 & CTransPath~\cite{wang2022ctp} \\
\hline
PRISM~\cite{shaikovski2024prism}      & 513 & 587 & Virchow~\cite{vorontsov2024virchow} \\
\hline
MADELEINE~\cite{jaume2024madeleine}  & 5 & 21 & CONCH~\cite{lu2024conch} \\
\hline
\multirow{4}{*}{\ac{cobra}} & \multirow{4}{*}{15} & \multirow{4}{*}{3} & CTransPath~\cite{wang2022ctp}, \\
          &     &  & UNI~\cite{chen2024uni}, \\
          &     &  & Virchow2~\cite{zimmermann2024virchow2},\\
          &     &  & H-Optimus-0~\cite{hoptimus0} \\
\bottomrule
\end{tabular}
}
\end{table}

\section{Method}

\Ac{cobra} is an unsupervised slide representation learning framework.
Given a set of \acp{wsi} $\{\mathbf{X}_i|\mathbf{X}_i\in\mathbb{R}^{d_x \times d_y \times 3}\}$ belonging to a single patient, it produces a single $d$-dimensional feature vector $\bm{z}\in\real{d}$ representing that patient. 
We provide a brief overview of \ac{cobra} below and in \cref{fig:cobra-ov}, before going into detail in the following subsections.

\Ac{cobra} operates on preprocessed patch embeddings (\cref{sec:preprocessing}) from a set of CPath \acp{fm}. 
Its architecture consists of a Mamba-2~\cite{dao2024mamba2} encoding module, a multi-head attention-based pooling module for learning a patient-level slide embedding (\cref{sec:method-arch}) and an embedding module that learns to align multiple \acp{fm} into the same embedding space. \Ac{cobra} can be deployed in various different modes, which makes it very flexible to adapt to different \acp{fm} (see \cref{sec:inference}). We train \ac{cobra} using a contrastive loss~\cite{Oord2019} (\cref{sec:method-loss}) and evaluate it on a variety of external validation tasks (\cref{sec:results}).

\subsection{Preprocessing}\label{sec:preprocessing}
Given a histology slide $(\mathbf{X}_i\in\mathbb{R}^{d_x \times d_y \times 3})$, we tessellate the slide into ($224 \times 224$)~px patches and remove background tiles by employing Canny background detection~\cite{rong2014canny}. Next, we extract patch embeddings with pretrained \acp{fm} and pool the resulting feature vectors into a slide embedding. We use \(fe_n\) to refer to the \(n\textsuperscript{th}\) \ac{fm}, \(fe_n \in \{ \text{CTP}, \text{UNI}, \text{V2}, \text{H0} \}\) denoting CTransPath~\cite{wang2022ctp}, UNI~\cite{chen2024uni}, Virchow2~\cite{zimmermann2024virchow2}, and H-optimus-0~\cite{hoptimus0}, respectively. By integrating \acp{fm} of different sizes and with different strengths, we aim to capture a diverse set of morphological features and ensure that our slide representations are robust and that \ac{cobra} is adaptable to other \acp{fm}. 
We obtain the patch embeddings $\bm{H}^{fe_n} \in \real{N_t \times d_n}$ with \(N_t\) and \(d_n\) denoting the number of tiles and the embedding dimension \(d_n \in ds=\{ \text{768}, \text{1024}, \text{1280}, \text{1536} \}\).
We extract patch embeddings at 0.5, 1.14 and 2~\ac{mpp} using 3048 \acp{wsi} from 2848 patients in \ac{tcga} BRCA, CRC, LUAD, LUSC and STAD. The use of multiple magnifications acts as a form of data augmentation in feature space, enriching the model's learning by providing multiscale contextual information. This approach enhances the model's ability to learn scale-invariant representations and improves its generalization across different tasks.

\subsection{Architecture}
\label{sec:method-arch}
The slide encoder consists of individual embedding MLPs for the different \acp{fm} and two Mamba-2 layers~\cite{dao2024mamba2} followed by multihead gated attention~\cite{jaume2024madeleine,Ilse2018Abmil}. 
The embedding module is a layer norm~\cite{ba2016layernormalization} followed by an MLP with one hidden layer and SiLU activation~\cite{hendrycks2023gelu}. It projects the different embedding dimensions of the \acp{fm} to the shared embedding space of the slide encoder. Inspired by MambaMIL~\cite{yang2024mambamil}, we use two Mamba~\cite{gu2024mambalineartimesequencemodeling} layers to efficiently encode the feature embeddings. We opt for the Mamba-2 \ac{ssd} modules as they scale substantially better for higher state-space dimensions compared to original Mamba modules~\cite{dao2024mamba2}. Additional information on the hyperparameters used can be found in \cref{sec:impl_deats}.

Formally, the architecture may be described as follows:\\
Let $f_{SE}: \real{N_t \times ds}\to\real{d}$ denote the slide encoder consisting of three submodules $f_E: \real{N_t\times ds}\to\real{N_t\times d}$, $f_S: \real{N_t\times d}\to\real{N_t\times d}$ and $f_A: \real{N_t\times d}\to\real{d}$, given by
\begin{equation}\label{eq:cobra-tr}
\bm{z} = f_{SE}(\bm{H}^{fe_n}) = f_A(f_S(f_E(\bm{H}^{fe_n}))),~\bm{H}^{fe_n}\in\real{N_t \times d_n}, 
\end{equation}
where $f_E$,$f_S$,$f_A$ denote the \textit{embedding module}, the \textit{state-space dual module} and the \textit{aggregation module}, respectively, and $d_n\in ds=\{ \text{768}, \text{1024}, \text{1280}, \text{1536} \}$ and $\bm{H}^{fe_n}$ refers to the patch embedding of the \(n\textsuperscript{th}\) \ac{fm}. The \textit{embedding module} $f_E$ is defined as follows:
\begin{equation}\label{eq:emb}
    \bm{H}_E = f_E(\bm{H}^{fe_n}) = \text{Lin}(\text{SiLU}(\text{Lin}(\text{LN}(\bm{H}^{fe_n})))),
\end{equation}
where Lin denotes a linear layer and LN denotes layer norm. The \textit{state-space dual module} $f_S$ is specified as:
\begin{equation}\label{eq:mamba}
    \bm{H}_S = f_S(\bm{H}_E) = \text{Lin}(\text{SSD}(\text{SSD}(\bm{H}_E)+\bm{H}_E)+\bm{H}_E).
\end{equation}
The \textit{aggregation module} $f_A$ consists of multi-head gated attention~\cite{jaume2024madeleine,Ilse2018Abmil} to aggregate the input embeddings into a single feature vector via a weighted average.
For multi-head gated attention, the encoded embeddings are split into $M$ parts for the $M$ heads:
$\bm{H}_S=\{\bm{H}_S^m\}_{m\in\{1,\dots,M\}}$ with $\bm{H}^m_S\in \real{N_t\times\frac{d}{M}}$. 
The \textit{aggregation module} $f_A$ is given by
\begin{equation}
    \label{eq:aggregation_module}
    \begin{split}
        \bm{z} = f_A(\bm{H}_{S}) &= \sum\limits_{k=1}^{N_t} a_k(\bm{H}_{S,k}) \cdot \bm{H}_{S,k}; \\
        a_k(\bm{H}_{S,k}) &= \frac{1}{M} \sum\limits_{m=1}^M a^m_k(\bm{H}_{S,k}^{m}),
    \end{split}
\end{equation}
with $\bm{H}_{S,k}\in\real{d}$ and $a_k^m\in\mathbb{R}$ is defined as:
\begin{align}\label{eq:abmil}
    a_k^m(\bm{H}_{S,k}^{m})=\hspace{6.5cm}& \notag \\ 
    \quad\frac{\exp\bigg(\bm{w}_m^{\top}\Big(\tanh\big(\bm{V}_m(\bm{H}_{S,k}^{m\top})\big)\odot\sigma\big(\bm{U}_m\bm{H}_{S,k}^{m\top}\big)\Big)\bigg)}
    {\sum\limits_{i}^{N_t}\exp\bigg(\bm{w}_m^{\top}\Big(\tanh\big(\bm{V}_m\bm{H}_{S,i}^{m\top}\big)\odot\sigma\big(\bm{U}_m\bm{H}_{S,i}^{m\top}\big)\Big)\bigg)},
\end{align}
with $\sigma$ denoting the sigmoid function and $\bm{w}\in \real{p \times 1}, \bm{U}\in\real{p \times d}, \bm{V}\in\real{p \times d}$ as learnable parameters and $p$ being the attention dimension.

\subsection{Inference modes}~\label{sec:inference}
During self-supervised pretraining, the slide encoder learns to map the patch embeddings ($\bm{H}^{fe_n}$) of different slides, patches, \acp{fm} and magnifications from the same patient to be close in slide embedding space ($\bm{z}$). For this purpose, encoded embeddings are aggregated to a single feature vector. 
\paragraph{Single-\ac{fm} inference mode}
In line with Wang et al.~\cite{Wang2024Chief}, we found it beneficial at inference time to compute the weighted average in \cref{eq:aggregation_module} using the original patch embeddings ($\bm{H}^{fe_n}$) instead of the encoded embeddings ($\bm{H}_{S}$) to obtain the slide-level representation (see \cref{sec:supp-mlp-results}).
Importantly, we still use the encoded embeddings to compute the weighting $a_k(\bm{H}_{S,k})$ of that average. 
Specifically, at inference time, \cref{eq:aggregation_module} becomes
\begin{equation}
    \label{eq:aggregation_module_inference}
    \bm{z} = f_{A_{\text{inf}}}(\bm{H}_{S},\bm{H}^{fe_n}) = \sum\limits_k^{N_t} a_k(\bm{H}_{S,k})\cdot \bm{H}_{k}^{fe_n}.
\end{equation}
We refer to this as the \textit{single-\ac{fm} inference mode} of \ac{cobra} and provide an ablation for the choice of \cref{eq:aggregation_module} \vs \cref{eq:aggregation_module_inference} in \cref{sec:supp-mlp-results}.
Unless stated otherwise, we will denote as \ac{cobra} the \textit{single-\ac{fm} inference mode} version using Virchow2 patch embeddings as input, which is given by 
\begin{equation}\label{eq:cobra}
\bm{z}=f_{{SE}_{\text{inf}}}\big(\bm{H}^{V2},\bm{H}^{V2}\big).
\end{equation}

\paragraph{Multi-\ac{fm} inference mode}
Additionally, one can use feature vectors from multiple different \acp{fm} and average the embeddings after the embedding module to extract patient-level features which incorporate the knowledge of the different \acp{fm} simultaneously with $f_{{SE}_{\text{inf}}}^\dag: \real{N_t \times ds}\times\real{N_t \times d_k}\to\real{d}$ ($d_k\in ds$):
\begin{equation}\label{eq:cobra-dagger}
    \begin{split}
    \bm{z}^\dag &= f_{{SE}_{\text{inf}}}^\dag\big(\{\bm{H}^{fe_n}\}_{n\in \{1,\dots,N_{FM}\}},\bm{H}^{fe_l}\big) \\
    &= f_{A_{\text{inf}}}\Bigg(f_S\bigg(\frac{\sum^{N_{FM}}_{n}f_E^\dag(\bm{H}^{fe_n})}{N_{FM}}\bigg),\bm{H}^{fe_l}\Bigg).
    \end{split}
\end{equation}
Here, $N_{FM}$ denotes the number of \acp{fm} used for pretraining and $\bm{H}^{fe_l}$ refers to the patch embeddings that are aggregated during inference. Additional information on the inference modes can be found in \cref{sec:fm-modes}.

\subsection{Contrastive loss function}
\label{sec:method-loss}
Following He et al.~\cite{He2020moco}, we interpret contrastive learning as training an encoder for a \textit{dictionary look-up task}:

Consider a set of encoded samples, denoted as $K=\{\bm{k}_1, \bm{k}_2, \dots, \bm{k}_N\}$, which represent the keys of a dictionary. For a given query $\bm{q}$, there exists exactly one matching key $\bm{k}^+\in K$. The contrastive loss is minimized when $\bm{q}$ closely matches $\bm{k}^+$ and diverges from all other keys. The InfoNCE~\cite{Oord2019} loss function is defined as
\begin{equation}\label{eq:moco}
\mathcal{L}_{\mathbf{q}} = -\log \frac{\psi(\mathbf{q},\mathbf{k^+})}{\sum\limits_{i=1}^N\psi(\mathbf{q},\mathbf{k_i})}, 
\end{equation}
where $\bm{q}$ and the corresponding $\bm{k}^+$ represent feature vectors produced by a randomly selected pretrained encoder, sampling patches from \acp{wsi} of the same patient and $N$ is the batch size or the length of the memory queue.
The function $\psi$ is defined as follows:
\begin{equation}\label{eq:psi}
    \psi(\mathbf{x_1},\mathbf{x_2}) = \exp(\text{sim}(\mathbf{x_1},\mathbf{x_2})/\tau),
\end{equation}
where $\tau$ denotes the temperature parameter and the cosine similarity function is depicted as $\text{sim}(\cdot)$.
To avoid feature collapse, the keys and queries should be generated by distinct encoders. Let $\theta_q$ denote the parameters of the query encoder with the dense projection head, then the parameters of the key encoder $\theta_k$ are updated as follows: 
\begin{equation}\label{eq:ke}
    \theta_k\leftarrow m\theta_k+(1-m)\theta_q,
\end{equation}
where $m\in [0,1)$ is the momentum coefficient. With the key encoder as the exponential average of the query encoder, the key representations stay more consistent, which enables a more stabilized training process. We adapted the public MoCo-v3~\cite{chen2021mocov3} repository for our experiments to align the embedding space of the slide embeddings generated with tile embeddings from different \acp{fm}.

\begin{table*}
\centering
\caption{\textbf{Comparison of different slide encoders and mean baselines.} AUC performance of downstream tasks trained on TCGA and deployed on CPTAC. ST denotes Subtyping, SE denotes Slide Encoder. $\overline{{\text{{Overline}}}}$ indicates mean over patch embeddings, $\overline{\text{Concatenated}}$ refers to concatenated mean embeddings of all FMs involved in \textsc{{Cobra}}’s pretraining, and $\overline{\text{Ensemble Prediction}}$ refers to the average of predictions from the mean patch embeddings of the training FMs. \textbf{{Bold}} indicates the best performance, and $\underline{\text{underline}}$ indicates the second-best performance.}
\label{tab:main_results}
\resizebox{\textwidth}{!}{%
\begin{tabular}{l|l|llll|llll|llllll|l}
\toprule
AUROC[\%] & NSCLC & \multicolumn{4}{c|}{LUAD} & \multicolumn{4}{c|}{BRCA} & \multicolumn{6}{c|}{COAD} & Average \\
Model & ST & STK11 & EGFR & TP53 & KRAS & ESR1 & PGR & ERBB2 & PIK3CA & MSI & BRAF & LN & KRAS & Side & PIK3CA &  \\
\midrule
$\overline{\text{CTransPath}}$~\cite{wang2022ctp} & $87.2_{1.5}$ & $62.8_{2.5}$ & $59.3_{7.4}$ & $70.1_{2.3}$ & $52.4_{5.9}$ & $68.1_{2.5}$ & $66.5_{2.1}$ & $48.6_{1.5}$ & $56.3_{3.0}$ & $76.1_{4.6}$ & $59.8_{2.3}$ & $59.8_{1.0}$ & $55.9_{7.7}$ & $52.5_{2.6}$ & $56.3_{6.3}$ & $62.1_{4.2}$ \\
$\overline{\text{Virchow}}$~\cite{vorontsov2024virchow} & $89.4_{0.6}$ & $76.5_{6.8}$ & $60.3_{2.0}$ & $70.7_{1.7}$ & $54.3_{6.9}$ & $66.9_{4.8}$ & $60.5_{3.9}$ & $51.3_{5.7}$ & $63.5_{3.7}$ & $62.1_{6.7}$ & $65.0_{2.6}$ & $58.2_{5.1}$ & $53.9_{7.1}$ & $52.3_{3.1}$ & $52.4_{5.7}$ & $62.5_{4.9}$ \\
$\overline{\text{CONCH}}$~\cite{lu2024conch} & $96.5_{0.3}$ & $66.0_{10.3}$ & $62.0_{7.6}$ & $74.6_{1.6}$ & $59.0_{7.4}$ & $85.3_{1.5}$ & $\underline{80.3}_{2.0}$ & $58.8_{11.0}$ & $63.2_{3.1}$ & $79.2_{0.5}$ & $57.5_{3.4}$ & $\mathbf{67.3}_{2.0}$ & $55.7_{8.6}$ & $53.4_{2.4}$ & $63.2_{5.3}$ & $68.1_{5.7}$ \\
$\overline{\text{UNI}}$~\cite{chen2024uni} & $95.8_{1.1}$ & $69.4_{2.4}$ & $70.1_{12.1}$ & $73.9_{0.8}$ & $50.7_{4.7}$ & $87.4_{3.1}$ & $74.9_{2.3}$ & $64.0_{3.4}$ & $62.3_{5.0}$ & $89.0_{1.6}$ & $73.0_{3.4}$ & $62.0_{7.8}$ & $\underline{63.5}_{2.8}$ & $59.4_{3.7}$ & $63.5_{6.2}$ & $70.6_{4.9}$ \\
$\overline{\text{H-Optimus}}$~\cite{hoptimus0} & $97.2_{0.4}$ & $78.5_{2.7}$ & $78.2_{3.5}$ & $71.3_{1.1}$ & $58.1_{3.7}$ & $85.2_{2.7}$ & $74.9_{3.3}$ & $51.4_{4.5}$ & $59.5_{4.9}$ & $94.7_{0.7}$ & $77.1_{7.1}$ & $55.5_{3.6}$ & $59.2_{4.2}$ & $\mathbf{62.2}_{3.4}$ & $62.4_{9.0}$ & $71.0_{4.3}$ \\
$\overline{\text{GigaPath}}$~\cite{xu2024gigapath} & $96.6_{0.7}$ & $71.3_{1.9}$ & $75.7_{6.8}$ & $75.4_{1.3}$ & $56.9_{6.2}$ & $85.7_{1.1}$ & $75.9_{1.8}$ & $64.5_{2.7}$ & $62.2_{5.9}$ & $93.3_{1.6}$ & $77.5_{2.6}$ & $61.7_{2.9}$ & $56.1_{5.0}$ & $60.0_{1.5}$ & $59.4_{7.9}$ & $71.5_{4.0}$ \\
$\overline{\text{Virchow2}}$~\cite{zimmermann2024virchow2} & $95.8_{0.7}$ & $79.6_{5.5}$ & $78.3_{4.6}$ & $72.1_{0.7}$ & $\mathbf{60.9}_{5.6}$ & $89.2_{2.8}$ & $79.3_{2.7}$ & $\underline{71.3}_{1.8}$ & $63.2_{4.8}$ & $\underline{94.9}_{1.2}$ & $81.6_{4.5}$ & $63.0_{1.9}$ & $59.3_{6.2}$ & $56.3_{3.8}$ & $62.7_{11.6}$ & $73.8_{4.7}$ \\
\midrule
$\overline{\text{Ensemble Prediction}}$ & $97.2_{0.3}$ & $77.2_{3.7}$ & $78.5_{4.1}$ & $73.3_{0.6}$ & $\underline{59.5}_{5.1}$ & $87.6_{2.8}$ & $77.2_{2.7}$ & $65.6_{2.0}$ & $63.3_{4.1}$ & $94.7_{1.0}$ & $78.9_{5.4}$ & $62.5_{3.9}$ & $\mathbf{64.1}_{3.0}$ & $60.5_{2.1}$ & $\underline{64.5}_{9.1}$ & $73.6_{4.0}$ \\
$\overline{\text{Concatenated}}$ & $97.4_{0.4}$ & $75.7_{3.0}$ & $\mathbf{80.2}_{2.2}$ & $72.5_{0.8}$ & $57.6_{4.8}$ & $\underline{89.6}_{1.4}$ & $79.1_{3.4}$ & $67.5_{3.9}$ & $61.8_{4.0}$ & $\mathbf{95.0}_{1.1}$ & $\underline{82.2}_{4.3}$ & $61.6_{2.5}$ & $59.7_{5.8}$ & $\underline{62.0}_{2.4}$ & $\mathbf{70.2}_{4.1}$ & $\underline{74.1}_{3.3}$ \\
\midrule
GigaPath-SE~\cite{xu2024gigapath} & $90.9_{1.3}$ & $67.0_{4.4}$ & $65.4_{4.4}$ & $73.7_{1.4}$ & $57.1_{5.2}$ & $72.9_{0.9}$ & $71.9_{3.3}$ & $55.4_{4.7}$ & $60.5_{4.6}$ & $66.2_{2.1}$ & $56.7_{4.5}$ & $54.6_{5.2}$ & $51.3_{2.9}$ & $45.8_{3.1}$ & $53.2_{5.7}$ & $62.8_{3.9}$ \\
MADELEINE~\cite{jaume2024madeleine} & $94.0_{0.6}$ & $72.2_{8.7}$ & $64.0_{6.7}$ & $72.0_{2.8}$ & $51.9_{3.9}$ & $80.1_{1.7}$ & $73.7_{1.3}$ & $66.7_{2.7}$ & $64.9_{1.6}$ & $68.6_{9.1}$ & $54.2_{6.7}$ & $60.3_{7.3}$ & $58.9_{6.6}$ & $50.5_{1.6}$ & $59.5_{8.6}$ & $66.1_{5.5}$ \\
CHIEF~\cite{Wang2024Chief} & $93.6_{0.8}$ & $64.2_{10.7}$ & $62.8_{10.9}$ & $73.4_{1.5}$ & $50.1_{5.0}$ & $83.0_{0.5}$ & $77.5_{0.3}$ & $63.4_{2.3}$ & $\underline{65.4}_{1.5}$ & $75.1_{4.8}$ & $63.6_{4.3}$ & $58.0_{1.7}$ & $58.4_{3.8}$ & $48.2_{4.2}$ & $56.6_{3.2}$ & $66.2_{4.9}$ \\
PRISM~\cite{shaikovski2024prism} & $\mathbf{99.2}_{0.1}$ & $\mathbf{87.6}_{1.6}$ & $70.7_{2.4}$ & $\underline{78.2}_{0.5}$ & $52.9_{8.5}$ & $\mathbf{92.2}_{0.7}$ & $\mathbf{84.2}_{0.5}$ & $64.5_{6.0}$ & $\mathbf{69.4}_{2.1}$ & $79.1_{1.5}$ & $59.9_{1.4}$ & $\underline{67.2}_{2.4}$ & $54.6_{6.2}$ & $52.2_{1.8}$ & $52.1_{6.8}$ & $70.9_{3.8}$ \\
\midrule
\textsc{{Cobra}} & $\underline{98.1}_{0.2}$ & $\underline{84.0}_{2.9}$ & $\underline{80.0}_{2.4}$ & $\mathbf{78.4}_{2.9}$ & $59.2_{6.2}$ & $\underline{89.6}_{2.0}$ & $79.2_{2.4}$ & $\mathbf{71.6}_{2.2}$ & $63.6_{6.2}$ & $94.1_{0.5}$ & $\mathbf{87.8}_{2.0}$ & $65.7_{2.5}$ & $62.1_{10.4}$ & $58.3_{1.9}$ & $57.6_{7.5}$ & $\mathbf{75.3}_{4.4}$ \\
\bottomrule
\end{tabular}
}
\end{table*}

\section{Experiments \& results}
\label{sec:results}

\subsection{Dataset}
\paragraph{TCGA} 
We collected 3048 WSIs from 2848 patients using the cohorts TCGA~\cite{TCGA2013} Breast Invasive Carcinoma (TCGA-BRCA, 1112~\acp{wsi}), TCGA Colorectal Carcinoma (TCGA-CRC, 566~\acp{wsi}), TCGA Lung Adenocarcinoma (TCGA-LUAD, 524~\acp{wsi}), TCGA Lung Squamous Cell Carcinoma (TCGA-LUSC, 496~\acp{wsi}), and TCGA Stomach Adenocarcinoma (TCGA-STAD, 350~\acp{wsi}). See \cref{sec:supp-data} for detailed information. These cohorts were used for pretraining \ac{cobra} and for training the downstream classifiers and linear regression models. We emphasize that neither \ac{cobra} nor any FMs used in this study were pretrained on datasets included in the evaluation of the downstream tasks, precluding any data leakage. 

\paragraph{CPTAC} 
We collected 1604 WSIs from 444 patients using the cohorts CPTAC~\cite{Edwards2015cptac} Breast Invasive Carcinoma (CPTAC-BRCA, 395~\acp{wsi}), CPTAC Colon Adenocarcinoma (CPTAC-COAD, 233~\acp{wsi}), CPTAC Lung Adenocarcinoma (CPTAC-LUAD, 498~\acp{wsi}), and CPTAC Lung Squamous Cell Carcinoma (CPTAC-LUSC, 478~\acp{wsi}). These cohorts were exclusively used for external validation.

\subsection{Pretraining setup}

We trained \ac{cobra} on patch embeddings derived from slides of 2848 patients, using a batch size of 1024 across four NVIDIA A100 GPUs for 2000 epochs, which took approximately 40 hours. In total, we used 36576 extracted feature embeddings consisting of 3048 \acp{wsi} for each of the four \acp{fm} models and each of the three magnifications included into the pretraining. Additional information about the hyperparameters used for the training of \ac{cobra} can be found in the Appendix \cref{tab:hyperparameters}.

\begin{table*}
\centering
\caption{\textbf{Ablation over different inference modes.} AUC performance of \textsc{Cobra} embeddings compared to mean embeddings of the FMs involved. $\overline{\text{Overline}}$  indicates mean over patch embeddings, $^\dag$ indicates that embeddings of all four training FMs were used to generate the weighting vector (\cref{eq:cobra-dagger}). For the other \ac{cobra} entries, we used the inference mode from (\cref{eq:aggregation_module_inference}). \textbf{Bold} indicates the best performance, and $\underline{\text{underline}}$ indicates the second-best performance. The abbreviations are as follows: CTP: CTransPath~\cite{wang2022ctp}, H0: H-Optimus-0~\cite{hoptimus0}, V2: Virchow-2~\cite{zimmermann2024virchow2}, GP: GigaPath~\cite{xu2024gigapath}. The different magnifications (5$\times$, 9$\times$, 20$\times$) indicate which magnification of the WSIs was used to extract the embeddings.}
\label{tab:ablation}
\resizebox{\textwidth}{!}{%
\begin{tabular}{l|l|llll|llll|llllll|l}
\toprule
AUC[\%] & NSCLC & \multicolumn{4}{c|}{LUAD} & \multicolumn{4}{c|}{BRCA} & \multicolumn{6}{c|}{COAD} & Average \\
Model & ST & STK11 & EGFR & TP53 & KRAS & ESR1 & PGR & ERBB2 & PIK3CA & MSI & BRAF & LN & KRAS & Side & PIK3CA &  \\
\midrule
$\overline{\text{CTransPath}}$~\cite{wang2022ctp} & $87.2_{1.5}$ & $62.8_{2.5}$ & $59.3_{7.4}$ & $70.1_{2.3}$ & $52.4_{5.9}$ & $68.1_{2.5}$ & $66.5_{2.1}$ & $48.6_{1.5}$ & $56.3_{3.0}$ & $76.1_{4.6}$ & $59.8_{2.3}$ & $59.8_{1.0}$ & $55.9_{7.7}$ & $52.5_{2.6}$ & $56.3_{6.3}$ & $62.1_{4.2}$ \\
$\overline{\text{UNI}}$~\cite{chen2024uni} & $95.8_{1.1}$ & $69.4_{2.4}$ & $70.1_{12.1}$ & $73.9_{0.8}$ & $50.7_{4.7}$ & $87.4_{3.1}$ & $74.9_{2.3}$ & $64.0_{3.4}$ & $62.3_{5.0}$ & $89.0_{1.6}$ & $73.0_{3.4}$ & $62.0_{7.8}$ & $63.5_{2.8}$ & $59.4_{3.7}$ & $\underline{63.5}_{6.2}$ & $70.6_{4.9}$ \\
$\overline{\text{H-Optimus}}$~\cite{hoptimus0} & $97.2_{0.4}$ & $78.5_{2.7}$ & $78.2_{3.5}$ & $71.3_{1.1}$ & $58.1_{3.7}$ & $85.2_{2.7}$ & $74.9_{3.3}$ & $51.4_{4.5}$ & $59.5_{4.9}$ & $94.7_{0.7}$ & $77.1_{7.1}$ & $55.5_{3.6}$ & $59.2_{4.2}$ & $\mathbf{62.2}_{3.4}$ & $62.4_{9.0}$ & $71.0_{4.3}$ \\
$\overline{\text{GigaPath}}$~\cite{xu2024gigapath} & $96.6_{0.7}$ & $71.3_{1.9}$ & $75.7_{6.8}$ & $75.4_{1.3}$ & $56.9_{6.2}$ & $85.7_{1.1}$ & $75.9_{1.8}$ & $64.5_{2.7}$ & $62.2_{5.9}$ & $93.3_{1.6}$ & $77.5_{2.6}$ & $61.7_{2.9}$ & $56.1_{5.0}$ & $60.0_{1.5}$ & $59.4_{7.9}$ & $71.5_{4.0}$ \\
$\overline{\text{Virchow2}}$~\cite{zimmermann2024virchow2} & $95.8_{0.7}$ & $79.6_{5.5}$ & $78.3_{4.6}$ & $72.1_{0.7}$ & $\mathbf{60.9}_{5.6}$ & $89.2_{2.8}$ & $79.3_{2.7}$ & $71.3_{1.8}$ & $63.2_{4.8}$ & $94.9_{1.2}$ & $81.6_{4.5}$ & $63.0_{1.9}$ & $59.3_{6.2}$ & $56.3_{3.8}$ & $62.7_{11.6}$ & $73.8_{4.7}$ \\
\hline
\textsc{{Cobra}}-CTP & $95.9_{0.6}$ & $65.0_{10.5}$ & $66.0_{5.5}$ & $74.8_{1.8}$ & $49.2_{4.4}$ & $78.6_{0.7}$ & $72.2_{0.6}$ & $62.0_{1.4}$ & $60.9_{3.6}$ & $80.3_{2.2}$ & $73.2_{3.0}$ & $61.3_{1.6}$ & $52.4_{4.2}$ & $48.1_{2.3}$ & $56.3_{4.3}$ & $66.4_{4.0}$ \\
\textsc{{Cobra}}-UNI & $98.8_{0.3}$ & $79.4_{2.5}$ & $76.5_{5.3}$ & $78.9_{1.2}$ & $52.2_{4.2}$ & $88.1_{1.7}$ & $80.5_{3.0}$ & $65.1_{4.3}$ & $63.9_{5.2}$ & $89.1_{1.1}$ & $82.8_{1.5}$ & $64.6_{2.2}$ & $59.0_{8.4}$ & $57.4_{2.1}$ & $\mathbf{64.3}_{5.8}$ & $73.4_{3.9}$ \\
\textsc{{Cobra}}-H0 & $\underline{\mathbf{99.4}}_{0.2}$ & $\underline{86.5}_{1.8}$ & $79.9_{2.9}$ & $80.1_{2.4}$ & $54.3_{4.7}$ & $87.1_{1.0}$ & $74.0_{4.2}$ & $64.2_{4.9}$ & $55.7_{2.3}$ & $\mathbf{96.0}_{0.6}$ & $86.2_{3.3}$ & $58.2_{2.5}$ & $62.2_{4.4}$ & $57.2_{1.9}$ & $62.9_{4.7}$ & $73.6_{3.2}$ \\
\textsc{{Cobra}}-V2 & $98.1_{0.2}$ & $84.0_{2.9}$ & $80.0_{2.4}$ & $78.4_{2.9}$ & $59.2_{6.2}$ & $\underline{\mathbf{89.6}}_{2.0}$ & $79.2_{2.4}$ & $\underline{71.6}_{2.2}$ & $63.6_{6.2}$ & $94.1_{0.5}$ & $\underline{87.8}_{2.0}$ & $\underline{65.7}_{2.5}$ & $62.1_{10.4}$ & $58.3_{1.9}$ & $57.6_{7.5}$ & $\underline{75.3}_{4.4}$ \\
\hline
\textsc{{Cobra}}$^\dag$-CTP & $95.9_{0.6}$ & $68.1_{5.1}$ & $69.2_{4.7}$ & $75.1_{1.6}$ & $46.7_{4.1}$ & $77.9_{0.8}$ & $71.3_{1.3}$ & $59.3_{1.5}$ & $59.2_{1.3}$ & $80.3_{1.5}$ & $73.5_{3.4}$ & $60.6_{2.5}$ & $55.2_{4.5}$ & $48.3_{3.4}$ & $54.3_{5.2}$ & $66.3_{3.2}$ \\
\textsc{{Cobra}}$^\dag$-UNI & $99.1_{0.2}$ & $79.1_{2.8}$ & $76.2_{4.6}$ & $\underline{80.2}_{0.7}$ & $55.0_{5.7}$ & $86.0_{1.7}$ & $78.1_{3.2}$ & $60.3_{4.9}$ & $62.3_{3.1}$ & $89.1_{0.7}$ & $83.5_{1.5}$ & $\underline{65.7}_{2.1}$ & $\underline{65.3}_{4.2}$ & $57.2_{2.1}$ & $61.9_{2.1}$ & $73.3_{3.1}$ \\
\textsc{{Cobra}}$^\dag$-H0 & $\underline{\mathbf{99.4}}_{0.1}$ & $\mathbf{86.9}_{2.0}$ & $\underline{80.9}_{3.4}$ & $79.9_{1.8}$ & $56.7_{3.6}$ & $87.8_{1.2}$ & $72.8_{3.4}$ & $59.9_{2.1}$ & $58.0_{0.9}$ & $\underline{95.2}_{1.1}$ & $84.9_{3.7}$ & $58.1_{2.6}$ & $59.7_{7.2}$ & $58.5_{2.4}$ & $61.2_{3.4}$ & $73.3_{3.1}$ \\
\textsc{{Cobra}}$^\dag$-V2-5$\times$ & $99.0_{0.2}$ & $79.0_{1.2}$ & $\mathbf{82.9}_{2.7}$ & $79.9_{1.8}$ & $\underline{59.4}_{3.1}$ & $89.0_{1.3}$ & $\underline{81.6}_{1.5}$ & $62.1_{4.7}$ & $\mathbf{67.2}_{2.9}$ & $94.1_{0.7}$ & $75.5_{2.7}$ & $\mathbf{68.1}_{8.5}$ & $\mathbf{69.6}_{3.4}$ & $51.4_{7.8}$ & $61.3_{1.5}$ & $74.7_{3.7}$ \\
\textsc{{Cobra}}$^\dag$-V2-9$\times$ & $98.9_{0.2}$ & $79.5_{1.2}$ & $76.7_{3.9}$ & $80.0_{1.8}$ & $53.0_{4.4}$ & $\underline{\mathbf{89.6}}_{1.6}$ & $\mathbf{83.6}_{1.7}$ & $70.6_{2.6}$ & $\underline{65.8}_{4.8}$ & $95.1_{0.9}$ & $82.5_{2.5}$ & $61.7_{0.5}$ & $61.2_{3.2}$ & $\underline{61.9}_{2.9}$ & $58.4_{12.8}$ & $74.6_{4.2}$ \\
\textsc{{Cobra}}$^\dag$-V2-20$\times$ & $98.4_{0.2}$ & $84.6_{1.9}$ & $78.9_{3.6}$ & $78.4_{2.6}$ & $55.9_{7.1}$ & $\underline{\mathbf{89.6}}_{1.7}$ & $80.0_{2.3}$ & $\mathbf{72.2}_{1.7}$ & $65.1_{4.5}$ & $94.2_{0.6}$ & $\mathbf{88.7}_{1.6}$ & $64.8_{2.3}$ & $60.6_{5.7}$ & $58.6_{1.7}$ & $61.5_{4.1}$ & $\mathbf{75.4}_{3.3}$ \\
\hline
\textsc{{Cobra}}$^\dag$-GP & $98.9_{0.3}$ & $81.5_{2.3}$ & $78.7_{4.2}$ & $\mathbf{80.9}_{1.1}$ & $56.9_{5.3}$ & $87.8_{1.2}$ & $77.5_{1.1}$ & $65.4_{1.3}$ & $64.6_{3.8}$ & $93.5_{1.2}$ & $85.6_{2.0}$ & $64.7_{2.4}$ & $59.2_{6.6}$ & $57.4_{2.1}$ & $56.9_{9.4}$ & $74.0_{3.8}$ \\
\bottomrule
\end{tabular}
}
\end{table*}

\begin{table*}
\centering
\caption{\textbf{Evaluation of the magnification augmentation during pretraining} AUC performance of downstream tasks trained on TCGA and deployed on CPTAC. $^\dag$ indicates that embeddings of all four training FMs were used to generate the weighting vector (\cref{eq:cobra-dagger}), $^\ddag$ indicates that \ac{cobra} was only pretrained on 0.5~MPP. ST denotes Subtyping. \textbf{Bold} indicates the best performance, and $\underline{\text{underline}}$ indicates the second-best performance.}
\label{tab:magn-ablation}
\resizebox{\textwidth}{!}{%
\begin{tabular}{c|l|l|llll|llll|llllll|l}
\toprule
&AUC[\%] & NSCLC & \multicolumn{4}{c|}{LUAD} & \multicolumn{4}{c|}{BRCA} & \multicolumn{6}{c|}{COAD} & Average \\
&Model & ST & STK11 & EGFR & TP53 & KRAS & ESR1 & PGR & ERBB2 & PIK3CA & MSI & BRAF & LN & KRAS & Side & PIK3CA &  \\
\midrule
\multirow{8}{*}{5$\times$}
&\textsc{{Cobra}}$^\ddag$-CTP & $88.7_{1.5}$ & $63.8_{10.5}$ & $66.2_{2.7}$ & $69.9_{2.0}$ & $51.7_{3.9}$ & $82.8_{1.2}$ & $75.0_{7.7}$ & $58.5_{3.1}$ & $61.5_{4.1}$ & $75.4_{1.4}$ & $62.5_{10.3}$ & $63.1_{2.5}$ & $53.8_{8.8}$ & $51.3_{1.4}$ & $51.7_{4.0}$ & $65.1_{5.4}$ \\
&\textsc{{Cobra}}$^\ddag$-UNI & $90.3_{0.5}$ & $72.3_{9.9}$ & $68.5_{4.0}$ & $69.2_{2.2}$ & $53.3_{7.5}$ & $81.4_{1.2}$ & $75.9_{1.4}$ & $56.0_{2.9}$ & $64.1_{4.8}$ & $77.8_{1.6}$ & $71.9_{2.6}$ & $60.4_{2.4}$ & $57.4_{8.0}$ & $55.3_{2.3}$ & $51.3_{10.4}$ & $67.0_{5.2}$ \\
&\textsc{{Cobra}}$^\ddag$-H0 & $89.5_{1.2}$ & $80.4_{4.2}$ & $69.1_{2.5}$ & $67.9_{1.6}$ & $52.5_{3.2}$ & $77.7_{1.8}$ & $73.6_{2.2}$ & $56.2_{2.8}$ & $62.0_{4.3}$ & $79.8_{2.8}$ & $71.4_{3.4}$ & $62.9_{1.9}$ & $56.3_{3.6}$ & $56.3_{2.1}$ & $58.0_{4.1}$ & $67.6_{2.9}$ \\
&\textsc{{Cobra}}$^\dag$-CTP & $96.6_{0.4}$ & $70.5_{3.9}$ & $70.5_{2.1}$ & $74.3_{1.0}$ & $53.2_{2.4}$ & $82.2_{0.9}$ & $77.1_{0.8}$ & $65.7_{2.9}$ & $66.0_{2.6}$ & $79.3_{1.3}$ & $67.9_{2.8}$ & $60.6_{3.2}$ & $53.0_{8.8}$ & $47.2_{2.4}$ & $51.6_{2.3}$ & $67.7_{3.2}$ \\
&\textsc{{Cobra}}$^\dag$-H0 & $97.6_{0.4}$ & $78.5_{4.0}$ & $71.3_{4.0}$ & $73.1_{1.3}$ & $51.8_{5.2}$ & $81.8_{1.1}$ & $74.9_{1.8}$ & $56.7_{2.0}$ & $63.1_{4.1}$ & $82.3_{1.6}$ & $71.4_{1.8}$ & $59.6_{3.1}$ & $56.0_{6.6}$ & $50.6_{2.4}$ & $55.7_{5.3}$ & $68.3_{3.5}$ \\
&\textsc{{Cobra}}$^\dag$-UNI & $97.1_{0.4}$ & $77.1_{1.6}$ & $71.5_{2.5}$ & $75.2_{1.1}$ & $\underline{57.3}_{2.0}$ & $82.7_{0.7}$ & $74.3_{1.5}$ & $57.1_{3.2}$ & $\mathbf{68.2}_{4.2}$ & $78.8_{1.9}$ & $70.5_{2.2}$ & $55.9_{4.3}$ & $54.6_{7.0}$ & $49.3_{6.8}$ & $58.4_{4.4}$ & $68.5_{3.5}$ \\
&\textsc{{Cobra}}$^\ddag$-V2 & $97.3_{0.4}$ & $81.3_{3.2}$ & $72.9_{2.6}$ & $77.0_{1.6}$ & $\mathbf{58.1}_{5.2}$ & $\mathbf{91.2}_{1.2}$ & $\mathbf{81.3}_{1.2}$ & $\underline{71.9}_{1.8}$ & $65.0_{3.1}$ & $87.2_{1.5}$ & $78.0_{1.8}$ & $64.2_{3.6}$ & $57.2_{12.4}$ & $60.3_{1.5}$ & $62.6_{8.8}$ & $73.7_{4.6}$ \\
&\textsc{{Cobra}}$^\dag$-V2 & $99.0_{0.2}$ & $81.6_{1.5}$ & $75.5_{2.7}$ & $79.9_{1.8}$ & $51.4_{7.8}$ & $89.0_{1.3}$ & $79.0_{1.2}$ & $67.2_{2.9}$ & $62.1_{4.7}$ & $94.1_{0.7}$ & $82.9_{2.7}$ & $61.3_{1.5}$ & $\mathbf{68.1}_{8.5}$ & $59.4_{3.1}$ & $\mathbf{69.6}_{3.4}$ & $\underline{74.7}_{3.7}$ \\
\hline
\multirow{8}{*}{9$\times$}
&\textsc{{Cobra}}$^\ddag$-CTP & $93.7_{1.1}$ & $68.7_{17.7}$ & $70.9_{3.8}$ & $72.3_{1.5}$ & $51.0_{6.8}$ & $81.0_{0.7}$ & $74.5_{0.8}$ & $50.2_{3.8}$ & $57.3_{2.2}$ & $72.4_{2.1}$ & $64.9_{1.7}$ & $58.9_{4.8}$ & $56.8_{5.5}$ & $54.2_{1.7}$ & $\underline{63.4}_{2.4}$ & $66.0_{5.6}$ \\
&\textsc{{Cobra}}$^\dag$-CTP & $96.4_{0.3}$ & $75.5_{3.5}$ & $69.4_{6.7}$ & $74.4_{1.4}$ & $52.1_{3.8}$ & $81.6_{0.6}$ & $76.2_{0.2}$ & $65.7_{1.5}$ & $62.0_{2.1}$ & $81.3_{1.0}$ & $72.1_{2.5}$ & $59.9_{2.2}$ & $52.6_{8.1}$ & $49.4_{4.4}$ & $55.4_{8.0}$ & $68.3_{4.0}$ \\
&\textsc{{Cobra}}$^\ddag$-H0 & $98.0_{0.6}$ & $83.9_{1.7}$ & $74.6_{4.0}$ & $75.9_{1.7}$ & $47.9_{3.1}$ & $83.0_{1.2}$ & $76.8_{1.1}$ & $61.1_{4.7}$ & $60.9_{2.0}$ & $76.5_{2.0}$ & $69.1_{2.3}$ & $\mathbf{66.4}_{2.5}$ & $58.6_{4.6}$ & $57.5_{1.7}$ & $58.4_{7.2}$ & $69.9_{3.2}$ \\
&\textsc{{Cobra}}$^\ddag$-UNI & $97.3_{0.5}$ & $80.2_{3.1}$ & $77.6_{4.6}$ & $75.7_{1.2}$ & $53.5_{3.5}$ & $86.3_{1.2}$ & $79.1_{0.8}$ & $60.0_{5.5}$ & $\underline{66.7}_{2.5}$ & $78.3_{2.2}$ & $72.0_{0.9}$ & $60.1_{3.0}$ & $56.9_{11.5}$ & $58.8_{1.9}$ & $59.7_{6.1}$ & $70.8_{4.3}$ \\
&\textsc{{Cobra}}$^\ddag$-V2 & $97.5_{0.7}$ & $84.3_{1.5}$ & $77.2_{4.2}$ & $77.7_{2.1}$ & $52.0_{6.8}$ & $88.4_{1.9}$ & $79.4_{1.3}$ & $68.6_{4.4}$ & $62.1_{2.2}$ & $81.3_{1.6}$ & $72.8_{0.4}$ & $62.6_{2.3}$ & $58.5_{9.7}$ & $60.3_{1.1}$ & $59.4_{11.1}$ & $72.1_{4.7}$ \\
&\textsc{{Cobra}}$^\dag$-H0 & $\underline{99.3}_{0.2}$ & $83.4_{2.3}$ & $73.6_{3.3}$ & $78.7_{2.1}$ & $52.4_{5.5}$ & $84.1_{2.0}$ & $77.2_{0.9}$ & $66.7_{1.7}$ & $62.3_{3.6}$ & $91.4_{0.7}$ & $82.5_{3.3}$ & $63.8_{2.7}$ & $56.2_{4.5}$ & $57.3_{2.8}$ & $58.3_{2.5}$ & $72.5_{2.9}$ \\
&\textsc{{Cobra}}$^\dag$-UNI & $98.9_{0.3}$ & $71.6_{16.2}$ & $74.8_{3.4}$ & $\mathbf{80.5}_{1.7}$ & $56.3_{4.1}$ & $87.2_{0.7}$ & $79.1_{0.9}$ & $65.5_{2.6}$ & $66.0_{3.8}$ & $89.5_{1.4}$ & $\underline{85.2}_{1.9}$ & $59.9_{4.9}$ & $62.0_{8.7}$ & $58.4_{3.9}$ & $56.5_{5.7}$ & $72.8_{5.6}$ \\
&\textsc{{Cobra}}$^\dag$-V2 & $98.9_{0.2}$ & $83.6_{1.7}$ & $76.7_{3.9}$ & $80.0_{1.8}$ & $53.0_{4.4}$ & $\underline{89.6}_{1.6}$ & $79.5_{1.2}$ & $70.6_{2.6}$ & $65.8_{4.8}$ & $\underline{95.1}_{0.9}$ & $82.5_{2.5}$ & $61.7_{0.5}$ & $58.4_{12.8}$ & $\underline{61.9}_{2.9}$ & $61.2_{3.2}$ & $74.6_{4.2}$ \\
\hline
\multirow{8}{*}{20$\times$}
&\textsc{{Cobra}}$^\ddag$-CTP & $94.1_{0.9}$ & $69.0_{4.7}$ & $67.9_{11.3}$ & $77.0_{0.9}$ & $51.3_{7.3}$ & $76.5_{1.7}$ & $70.7_{0.8}$ & $58.0_{1.5}$ & $51.3_{5.7}$ & $81.9_{0.7}$ & $67.1_{2.4}$ & $54.7_{4.6}$ & $54.6_{7.1}$ & $51.2_{1.6}$ & $59.4_{5.4}$ & $65.6_{4.8}$ \\
&\textsc{{Cobra}}$^\dag$-CTP & $95.9_{0.6}$ & $68.1_{5.1}$ & $69.2_{4.7}$ & $75.1_{1.6}$ & $46.7_{4.1}$ & $77.9_{0.8}$ & $71.3_{1.3}$ & $59.3_{1.5}$ & $59.2_{1.3}$ & $80.3_{1.5}$ & $73.5_{3.4}$ & $60.6_{2.5}$ & $55.2_{4.5}$ & $48.3_{3.4}$ & $54.3_{5.2}$ & $66.3_{3.2}$ \\
&\textsc{{Cobra}}$^\ddag$-UNI & $97.8_{0.6}$ & $75.5_{3.7}$ & $78.5_{4.9}$ & $\underline{80.4}_{1.7}$ & $53.6_{5.7}$ & $83.1_{2.4}$ & $73.6_{2.2}$ & $63.4_{3.7}$ & $59.2_{2.7}$ & $86.4_{1.6}$ & $76.5_{1.0}$ & $62.0_{3.2}$ & $57.5_{6.4}$ & $61.1_{1.4}$ & $60.6_{2.4}$ & $71.3_{3.3}$ \\
&\textsc{{Cobra}}$^\ddag$-H0 & $98.8_{0.2}$ & $\underline{85.1}_{2.9}$ & $78.0_{8.4}$ & $79.5_{1.0}$ & $55.8_{3.9}$ & $86.7_{1.2}$ & $75.5_{2.7}$ & $66.3_{4.2}$ & $50.9_{2.7}$ & $91.0_{0.8}$ & $79.0_{2.7}$ & $56.0_{4.1}$ & $56.3_{6.3}$ & $\mathbf{64.3}_{0.9}$ & $62.8_{3.4}$ & $72.4_{3.7}$ \\
&\textsc{{Cobra}}$^\dag$-H0 & $\mathbf{99.4}_{0.1}$ & $\mathbf{86.9}_{2.0}$ & $\underline{\mathbf{80.9}}_{3.4}$ & $79.9_{1.8}$ & $56.7_{3.6}$ & $87.8_{1.2}$ & $72.8_{3.4}$ & $59.9_{2.1}$ & $58.0_{0.9}$ & $\mathbf{95.2}_{1.1}$ & $84.9_{3.7}$ & $58.1_{2.6}$ & $59.7_{7.2}$ & $58.5_{2.4}$ & $61.2_{3.4}$ & $73.3_{3.1}$ \\
&\textsc{{Cobra}}$^\dag$-UNI & $99.1_{0.2}$ & $79.1_{2.8}$ & $76.2_{4.6}$ & $80.2_{0.7}$ & $55.0_{5.7}$ & $86.0_{1.7}$ & $78.1_{3.2}$ & $60.3_{4.9}$ & $62.3_{3.1}$ & $89.1_{0.7}$ & $83.5_{1.5}$ & $\underline{65.7}_{2.1}$ & $\underline{65.3}_{4.2}$ & $57.2_{2.1}$ & $61.9_{2.1}$ & $73.3_{3.1}$ \\
&\textsc{{Cobra}}$^\ddag$-V2 & $96.9_{0.3}$ & $83.4_{4.0}$ & $\underline{\mathbf{80.9}}_{3.4}$ & $78.8_{1.5}$ & $56.7_{4.4}$ & $88.3_{1.7}$ & $77.8_{1.9}$ & $70.7_{4.7}$ & $58.1_{1.3}$ & $91.8_{0.8}$ & $80.6_{2.3}$ & $62.2_{1.8}$ & $54.7_{11.9}$ & $61.3_{1.9}$ & $61.0_{7.2}$ & $73.5_{4.4}$ \\
&\textsc{{Cobra}}$^\dag$-V2 & $98.4_{0.2}$ & $84.6_{1.9}$ & $78.9_{3.6}$ & $78.4_{2.6}$ & $55.9_{7.1}$ & $\underline{89.6}_{1.7}$ & $\underline{80.0}_{2.3}$ & $\mathbf{72.2}_{1.7}$ & $65.1_{4.5}$ & $94.2_{0.6}$ & $\mathbf{88.7}_{1.6}$ & $64.8_{2.3}$ & $60.6_{5.7}$ & $58.6_{1.7}$ & $61.5_{4.1}$ & $\mathbf{75.4}_{3.3}$ \\
\bottomrule
\end{tabular}
}
\end{table*}

\subsection{Tasks}

\Ac{cpath} is used for different task categories. One important such category is biomarker prediction. Here, we focused on \textit{STK11}, \textit{EGFR}, \textit{KRAS} and \textit{TP53} mutation prediction in LUAD, \textit{ESR1}, \textit{PGR} and \textit{ERBB2} expression, and \textit{PIK3CA} mutation prediction in BRCA, and MSI status, \textit{BRAF}, \textit{KRAS}, \textit{PIK3CA} mutation prediction in COAD. We also included classification of phenotypic subtypes, Non-Small Cell Lung Cancer (NSCLC) Subtyping and Sidedness prediction of COAD. Finally, we added N-Status prediction in COAD, a task that goes beyond the tissue itself and tries to classify whether the tumor has infiltrated lymph nodes, thereby influencing prognostication. 
We report \ac{auroc} results in the main text, additional metrics such as F1 score, \ac{auprc} and the balanced accuracy of all experiments can be found in \cref{sec:supp-results}. Unless indicated otherwise, all results are reported for 0.5~\ac{mpp} (20$\times$ \ac{wsi} magnification). In general, we conducted our evaluation experiments for three different \ac{wsi} magnifications: 0.5~\ac{mpp} (20$\times$), 1.14~\ac{mpp} (9$\times$) and 2~\ac{mpp} (5$\times$). Additional information about the downstream experiments can be found in \cref{sec:app-eval}.

\subsection{Evaluation of patient embeddings}

\paragraph{MLP downstream classification}

We evaluated \textsc{Cobra}'s patient-level slide embeddings following standard practice in \ac{cpath} using 5-fold cross-validation on the \ac{tcga} training cohort followed by deploying all five classifiers on the full external validation set CPTAC. The classifier is a simple MLP. Generating a slide embedding and then training a small MLP is much more efficient than current \ac{mil} approaches using tile embeddings. We compare \ac{cobra} to all mean patch embeddings of \acp{fm} used in this study and to the slide encoders MADELEINE~\cite{jaume2024madeleine}, PRISM~\cite{shaikovski2024prism}, GigaPath~\cite{xu2024gigapath} and CHIEF~\cite{Wang2024Chief} (see \cref{tab:main_results}). All slide encoders except GigaPath and MADELEINE manage to outperform the mean patch embeddings of the \ac{fm} they are based upon. However, \ac{cobra} is the only model that manages to reach a higher macro-AUC than Virchow2 mean patch embeddings.
Nevertheless, it should be noted that MADELEINE was trained only on BRCA slides. 
Still, \ac{cobra} also substantially outperforms MADELEINE on most BRCA tasks (\textit{ESR1}: +9.5\%, \textit{PGR} +5.5\%, \textit{ERBB2} +4.9\%, \textit{PIK3CA} -1.3\% AUC). 
Overall, \ac{cobra} improves over PRISM by +4.4\% average \ac{auroc} and over the mean of the patch embeddings of Virchow2 by +1.5\%. Especially on the COAD downstream tasks, MSI and \textit{BRAF}, \ac{cobra} achieves substantial performance increases over the other slide encoders of at least +15\% and +24.2\% average AUC, respectively.
\begin{figure}[h!]
    \centering
    \includegraphics[width=0.9\linewidth]{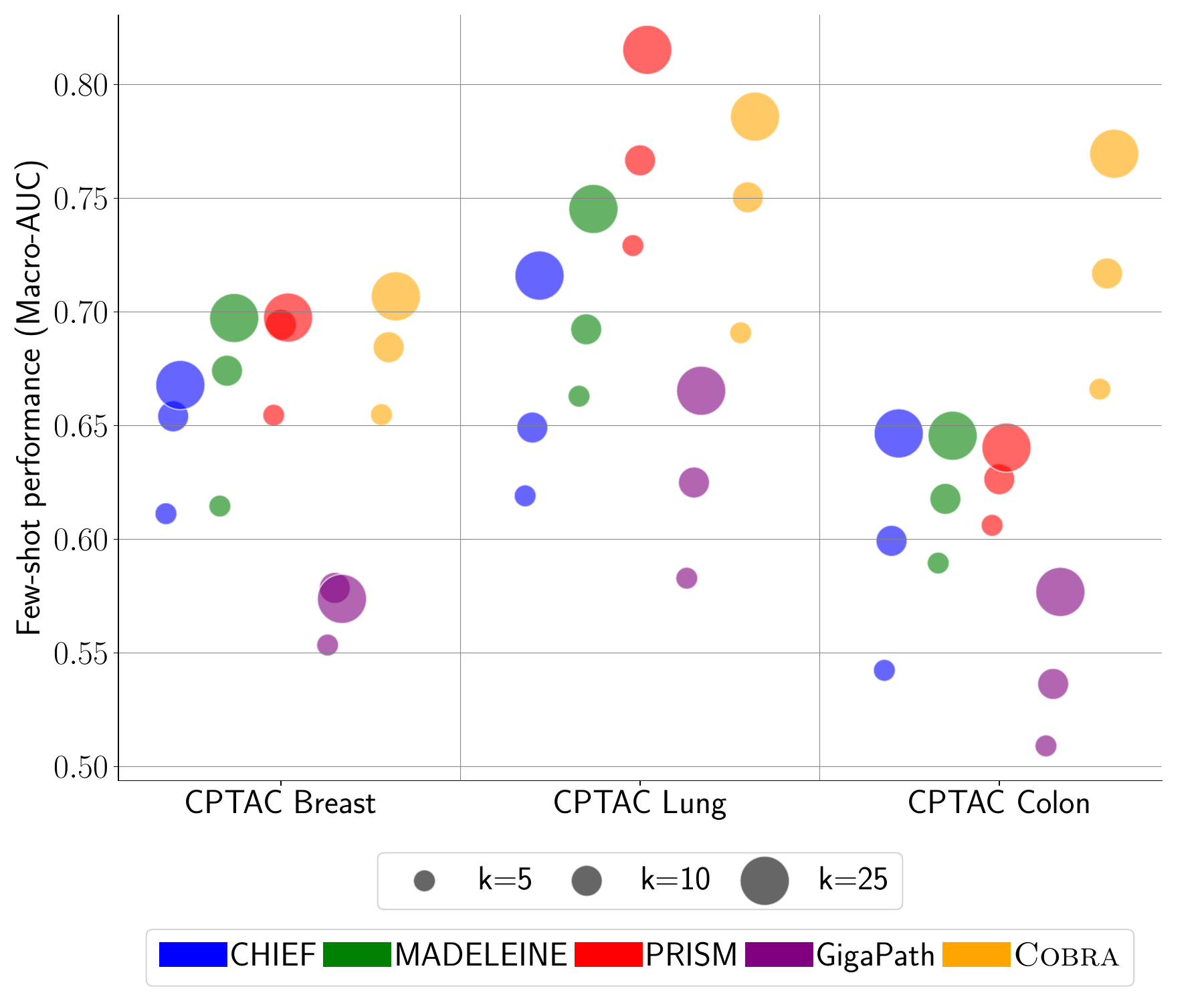}
    \caption{\textbf{Few shot linear probing classification}. Linear probing macro-AUC performance comparison for $k$ samples per class.}
    \label{fig:few-shot}
\end{figure}
\paragraph{Linear probing few-shot classification}
We also evaluate \ac{cobra} in a few-shot setting across 10 runs for high-performance tasks, where the mean patch embeddings of at least one \ac{fm} scores an average macro-AUC of $>$ 0.7 across the five folds of the full classification and where the \ac{tcga} cohorts contain at least 50 cases per class.
These tasks are NSCLC Subtyping, \textit{STK11}, \textit{EGFR} and \textit{TP53} mutation in LUAD, \textit{ESR1}, \textit{PGR} and \textit{ERBB2} expression in BRCA, and \textit{BRAF} mutation, Sidedness and MSI status prediction in COAD (see \cref{fig:few-shot}).
Although \ac{cobra} was only trained on very few samples and with only one modality, we observe that it is still robust enough to achieve high few-shot performance compared to the other slide encoders. On the BRCA tasks, it slightly outperforms the competition, while it substantially exceeds the results of the other models on the COAD tasks. We provide further results and information about the few shot experiments in \cref{sec:supp-lp-results}.
Interestingly, slide encoders demonstrate greater robustness with fewer training samples compared to their mean patch embedding baselines, as all slide encoders except GigaPath consistently outperform their corresponding baselines.

\begin{figure*}[h!]
    \centering
    \includegraphics[width=0.65\linewidth]{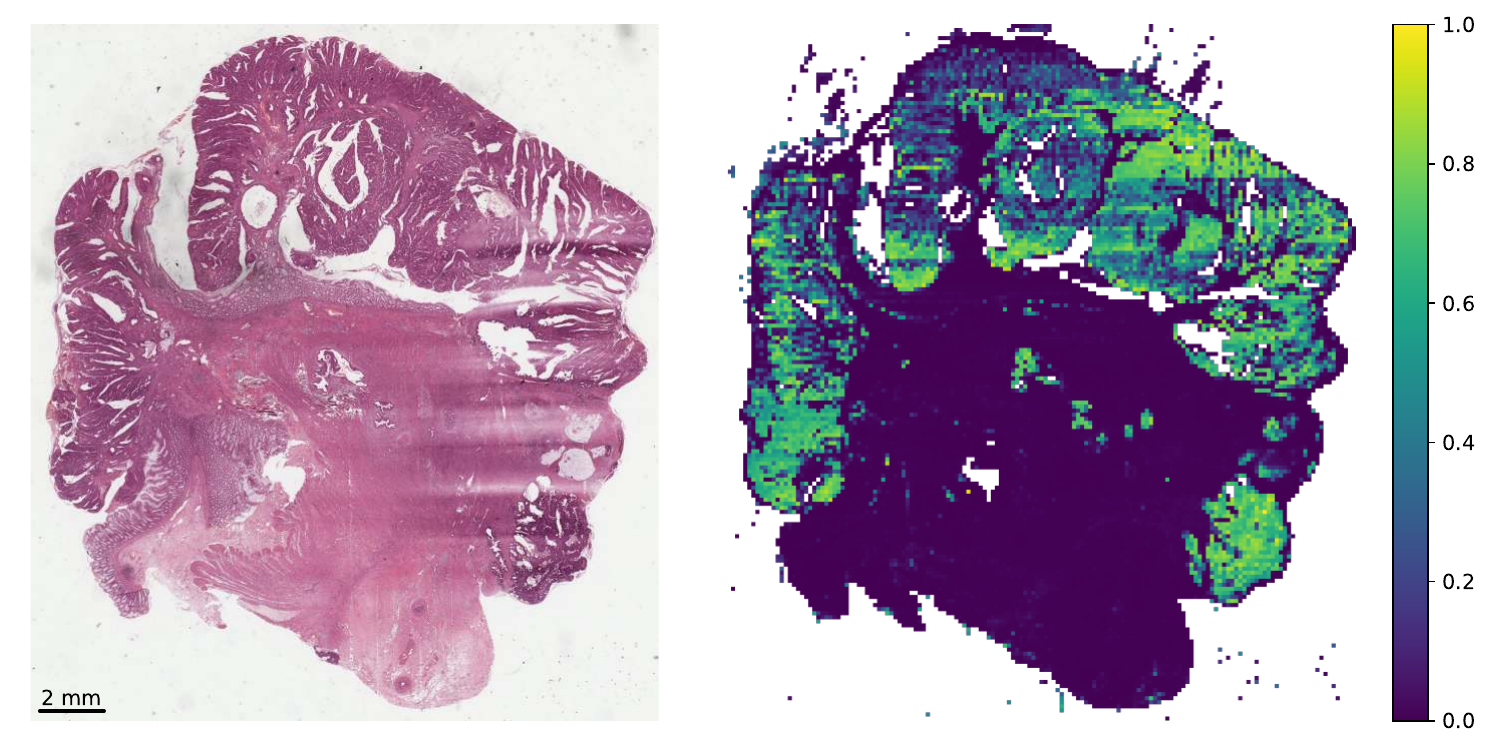}
    \caption{\textbf{\ac{cobra} Unsupervised Heatmap}. Visualization of the weighting scores for the tiles of a WSI generated by \textsc{Cobra} for Patient-ID TCGA-CA-6716 from TCGA-CRC.}
    \label{fig:heatmap-TCGA-CA-6716}
\end{figure*}
\subsection{Inference ablations}

\paragraph{Foundation models}
As \ac{cobra} is FM-agnostic, it can be used to improve small, inferior tile level \acp{fm} like CTransPath as \ac{cobra}-CTP improves over all slide encoders but PRISM (see \cref{tab:ablation,tab:main_results}). 
This substantially improves efficiency, as CTransPath has approximately 30M parameters, compared to more than 600M in Virchow2 and more than 1B in H-optimus-0.

\paragraph{Magnifications}
Another way to achieve efficiency improvements is to reduce the magnification of the \acp{wsi} for the patch embeddings, which in turn significantly reduces the number of tiles that need to be extracted and embedded. In particular, this change does not result in a significant performance drop as \ac{cobra}$^\dag$-V2-5$\times$ and \ac{cobra}$^\dag$-V2-9$\times$ achieve performance gains over the next best slide encoder PRISM at 0.5~MPP of +3.8\% and +3.7\% average AUC, respectively (see \cref{tab:ablation,tab:main_results}), which we attribute to our multiscale alignment during pretraining.

\paragraph{Combined inference and unseen FMs}
In a combined inference mode (indicated by $^\dag$ in \cref{tab:ablation}), where embeddings from all pretrained \acp{fm} are used, performance is slightly better for \ac{cobra}-V2, though it does not notably improve the downstream classification performance for 0.5~\ac{mpp}. 
In general, performance is comparable to the single-\ac{fm} mode. 
However, for smaller magnifications, the benefits of the combined inference mode become more notable (see Appendix \cref{tab:mlp_AUC-5x,tab:mlp_AUC-9x}).
Especially on 2~\ac{mpp}, \ac{cobra} exhibits gains over the single-\ac{fm} mode (on average +0.53\% AUC).
Furthermore, \ac{cobra} remains useful for future \acp{fm} as it can aggregate embeddings from unseen \acp{fm} and improve their performance over the mean baseline. We show evidence for this by deploying \ac{cobra} on GigaPath patch embeddings, which improves over the mean baseline of +2.5\% average AUC and +3.1\% average AUC over the next best slide encoder PRISM (\cref{tab:ablation,tab:main_results}).

\subsection{Pretraining ablation}

\paragraph{Single-magnification pretraining} We analyze the performance of a single magnification training on only 0.5~\ac{mpp} embeddings and find that the use of all three magnifications results in an average AUC improvement of +1.73\% AUC across all models when comparing the multi-FM inference mode (\cref{eq:cobra-dagger}). 
Furthermore, the three-magnification setup yields substantial gains in NSCLC subtyping at 5$\times$ magnification, with improvements of +6.8\% AUC for UNI, +7.9\% for CTransPath, +8.1\% for H-optimus-0 and +1.7\% for Virchow2 (\cref{tab:magn-ablation}). These results indicate that the use of multiple magnifications can enhance performance in certain cases and does not negatively impact model performance.

\subsection{Interpretability}
\Ac{cobra} enables unsupervised interpretability as it is an aggregation method of patch embeddings that calculates a weighted average by assigning each tile a softmaxed value, which can be interpreted as an attention value.
By visualizing these weightings for \acp{wsi}, we observe that the model shows high attention values for the tumor regions in the slide (see \cref{fig:heatmap-TCGA-CA-6716}). It is worth mentioning that for these heatmaps, no GradCam~\cite{Selvaraju_2019gradCam} is required, and they are generated only based on patch embeddings, so each tile only receives one value instead of pixel-level attention that can be achieved with other methods. However, this extremely simple approach is sufficient to identify important tumor regions in detail without any supervision such as targeted segmentation training. More examples and detailed explanations can be found in \cref{sec:supp-heatmaps}.

\begin{figure}
    \includegraphics[width=0.85\linewidth]{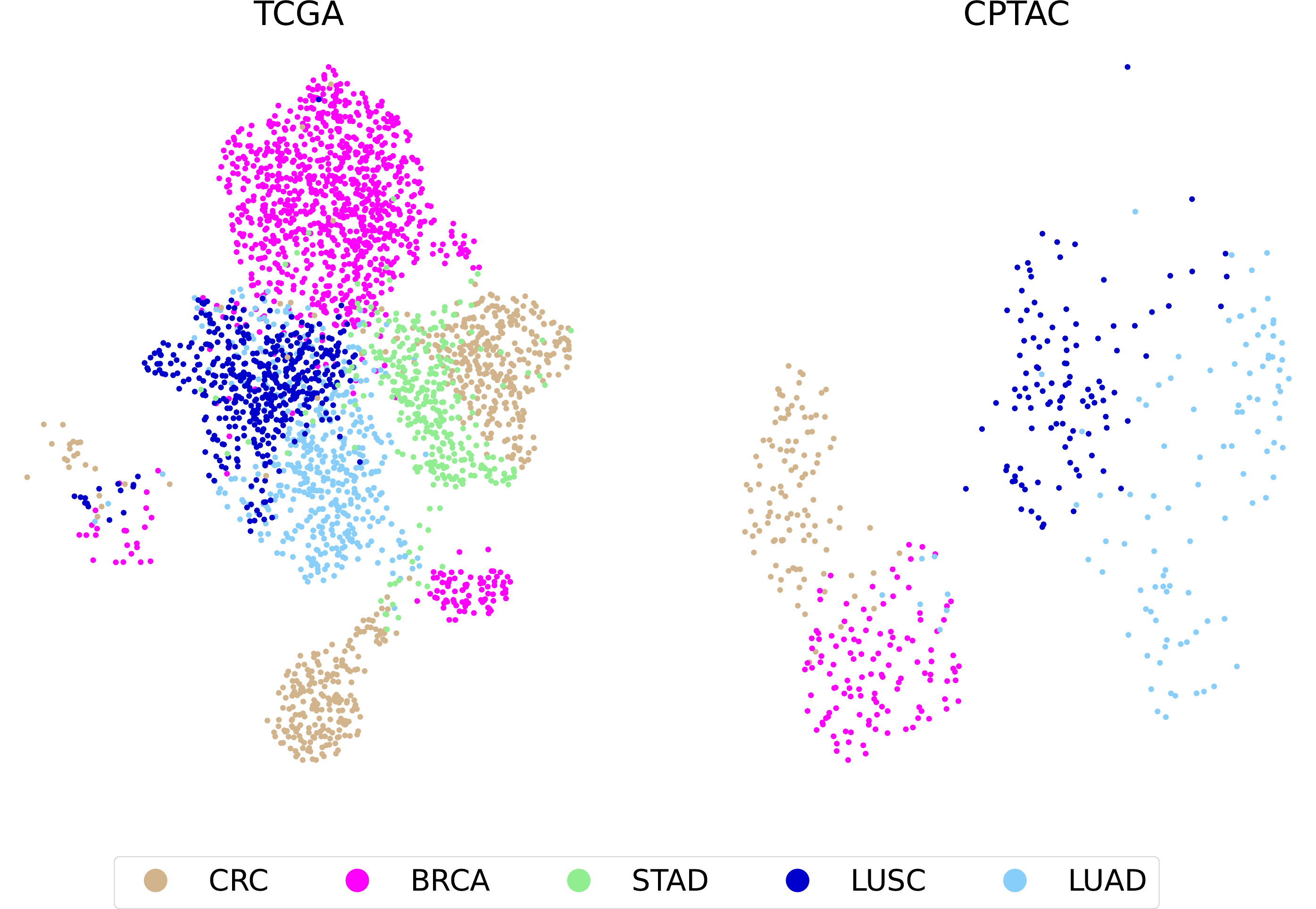}
  \caption{UMAP visualization of \ac{cobra}’s patient-level slide embeddings for TCGA and CPTAC datasets at $0.5~$\ac{mpp}. Each color represents a different tissue type, with five tissue types in total.}
  \label{fig:cobra-umap}
  \hfill
\end{figure}

Furthermore, we visualized \ac{cobra}'s embedding space using \ac{umap}~\cite{McInnes2018umap} plots of \ac{cobra}'s patient-level slide embeddings extracted at $0.5~$\ac{mpp} for TCGA and CPTAC (see \cref{fig:cobra-umap}). We observe a decent separation between the different tissue types involved in this study, indicating that \ac{cobra} learned meaningful representations that can distinguish between tissue types without supervision.

\section{Conclusion}
In this paper, we introduced \ac{cobra}, a novel FM- and task-agnostic approach for slide representation learning. Trained on only 3048 \acp{wsi} from \ac{tcga}, \ac{cobra} achieves \ac{sota} performance, even surpassing multimodal slide encoders. This is particularly valuable for medical imaging, where acquiring large annotated datasets is challenging due to privacy concerns and annotation costs. While additional data might enhance performance, our results indicate that \ac{cobra} is highly effective even in low-data regimes. These results highlight the potential of SSL in leveraging the strengths of histopathology FMs. Future work includes exploring SSL objectives that extend beyond contrastive approaches, as well as incorporating more cancer types, pretraining data and a larger variety of FMs into \ac{cobra}.
{\small\paragraph*{Acknowledgments} 
The authors gratefully acknowledge the GWK's support for funding this project by providing computing time through the Center for Information Services and HPC (ZIH) at TU Dresden. We also acknowledge the TCGA Research Network and the Clinical Proteomic Tumor Analysis Consortium (CPTAC), which generated the data on which the results shown in this study are based. GW is supported by Lothian NHS.}

\newpage
{
    \small
    \bibliographystyle{ieeenat_fullname}
    \bibliography{main}
}
\newpage
\appendix
\clearpage
\setcounter{page}{1}
\maketitlesupplementary
\setcounter{section}{0}

\section{Implementation details}\label{sec:impl_deats}

\paragraph{FM pretraining}
The detailed pretraining settings for \ac{cobra} can be found in \cref{tab:hyperparameters}. We used 25\% dropout in all MLPs.

\begin{table}
\centering
\begin{tabular}{l|l}
\textbf{Hyperparameter} & \textbf{Value} \\ 
\hline
Heads & 8 \\
Number of Mamba-2 layers & 2 \\
Embedding dimension & 768 \\
Input dimensions & 768, 1024, 1280, 1536 \\
Dropout & 0.25 \\
Attention hidden dimension & 96 \\ \hline
Teacher momentum & 0.99 \\
Contrastive loss temperature & 0.2 \\
Optimizer & AdamW~\cite{loshchilov2019adamw} \\
Learning rate & 5e-4 \\
Warmup epochs & 50 \\
Weight decay & 0.1 \\
Epochs & 2000 \\
Batch size & 1024 \\ 
Tile embeddings per patient & 768 \\ 
\end{tabular}
\caption{Hyperparameters for \ac{cobra} pretraining}
\label{tab:hyperparameters}
\end{table}

\subsection{Additional information on evaluation}\label{sec:app-eval}
\subsubsection{MLP downstream classification}
An MLP classifier is implemented using a two-layer architecture, with an input layer of 768 dimensions and a hidden layer of 256 dimensions. The hidden layer employs SiLU~\cite{hendrycks2023gelu} activation, followed by a dropout layer (50\%) for regularization. The output layer consists of a fully connected layer with the appropriate number of output classes. Cross-entropy loss with class weighting is applied to handle class imbalance. The classifier is trained using the AdamW~\cite{loshchilov2019adamw} optimizer with a learning rate of 0.0001 and weight decay of 0.01, employing a one-cycle policy for 32 epochs. Training is conducted in a 5-fold cross-validation setup, with early stopping and best model checkpoints monitored by validation loss.

\subsubsection{Linear probing}
Linear probing is implemented using a logistic regression objective based on sklearn.
We use the default sklearn L2 regularization (set to 1.0) with an lbfgs solver. We set the maximum iterations to 10,000 and apply balanced class weights. Training is conducted in a stratified sampling setting with 10 random runs, using 5, 10, and 25 cases per class in each run.

\section{Inference modes} \label{sec:fm-modes}
\ac{cobra} is designed to be flexible and versatile, supporting three primary inference modes: single-FM, multi-FM, and hybrid-FM. These modes allow for adaptability across various histology datasets and computational resources.

\paragraph{Single-FM inference mode} 
In this mode, \ac{cobra} utilizes patch embeddings extracted from a single feature extractor, such as Virchow2. The framework generates attention weights for the patches based on these embeddings and aggregates them to produce the final patient-level embedding. This mode is computationally efficient and achieves state-of-the-art performance with minimal overhead, making it ideal for scenarios requiring simplicity and resource efficiency.

\paragraph{Multi-FM inference mode} 
In multi-FM inference mode, \ac{cobra} integrates embeddings generated independently by multiple FMs. Each FM produces its own patch embeddings, which \textsc{Cobra}’s embedding module then projects into a shared embedding space. These projected embeddings from all FMs are averaged for each corresponding patch, resulting in unified patch embeddings that integrate diverse morphological representations. \ac{cobra} processes these averaged embeddings through its encoding and attention modules to produce attention weights. Finally, these attention weights are applied to the original embeddings from a selected primary FM to obtain the final patient-level representation. While this mode might improve robustness by leveraging multiple FMs simultaneously, performance gains compared to the single-FM mode appear marginal.

\paragraph{Hybrid-FM inference mode} 
The hybrid-FM inference mode allows \ac{cobra} to incorporate patch embeddings from previously unseen FMs without retraining the model. First, the patch embeddings (from one or more of \textsc{Cobra}'s pretraining \acp{fm}) are mapped into \textsc{Cobra}’s shared embedding space via the embedding module. Subsequently, the framework generates attention weights based on these encoded embeddings. Finally, these weights are applied to the original external patch embeddings of the previously unseen FM to generate a patient-level representation. This ability ensures that \ac{cobra} remains adaptable, allowing seamless integration and effective utilization of new FMs without requiring any retraining.

\paragraph{Handling different magnifications} 
\ac{cobra} is equipped to process patch embeddings extracted at various magnifications, including 0.5 MPP (20$\times$), 1.14 MPP (9$\times$), and 2 MPP (5$\times$). This flexibility ensures compatibility with a wide range of histology datasets, allowing for diverse applications without requiring adjustments to the core architecture.

\section{Data}
\label{sec:supp-data}
Overall, our study comprises a total of 4,652 WSIs from 3,292 patients, including the organs lung, stomach, breast and colon. We use 3,048 WSIs for pretraining \ac{cobra} and training the classifiers, and 1604 WSIs for external validation.  
The slides for TCGA are available at \href{https://portal.gdc.cancer.gov}{https://portal.gdc.cancer.gov}. 
The slides for CPTAC are available at \href{https://proteomics.cancer.gov/data-portal}{https://proteomics.cancer.gov/data-portal}. 
The molecular data for TCGA and CPTAC are available at \href{https://www.cbioportal.org}{https://www.cbioportal.org}~\cite{Cerami2012cbioportal}.

\paragraph{TCGA BRCA (training)}
We collected N=1,041 primary cases from the TCGA Breast Invasive Carcinoma (BRCA) cohort. For each case, we downloaded the corresponding molecular status: ER (N=1041; 770 positive, 271 negative), PR (N=1041; 704 positive, 337 negative), HER2 (N=1041; 125 positive, 916 negative), and PIK3CA driver mutation (N=1023; 687 WT, 336 MUT). We defined ER positive, PR positive, HER2 positive and PIK3CA MUT as positive classes for AUPRC and F1 scores.

\paragraph{TCGA CRC (training)}
We collected N=558 primary cases from the TCGA Colorectal Carcinoma (CRC) cohort. For each case, we downloaded the corresponding molecular status: MSI status (N=429; 368 MSS, 61 MSI), Lymph Node status (N=556; 318 N0, 238 N+), CRC sidedness (N=398; 230 left, 168 right), BRAF (N=501; 450 WT, 51 MUT), KRAS (N=501; 296 WT, 205 MUT), and PIK3CA driver mutation (N=501; 377 WT, 124 MUT). We defined MSI high, N+, right-sided CRC, BRAF MUT, KRAS MUT and PIK3CA MUT as positive classes for AUPRC and F1 scores.

\paragraph{TCGA LUAD (training)}
We collected N=461 primary cases from the TCGA Lung Adenocarcinoma (LUAD) cohort. For each case, we downloaded the corresponding molecular status: STK11 (N=461; 394 WT, 67 MUT), EGFR (N=461; 411 WT, 50 MUT), KRAS (N=461; 317 WT, 144 MUT), and TP53 driver mutation (N=461; 239 MUT, 222 WT). We defined STK11 MUT, EGFR MUT, KRAS MUT and TP53 MUT as positive classes for AUPRC and F1 scores.

\paragraph{TCGA NSCLC (training)}
We collected N=462 primary cases from the TCGA Lung Squamous Cell Carcinoma (LUSC) cohort and the aforementioned N=461 primary cases from the TCGA LUAD cohort. We defined LUAD as the positive class for AUPRC and F1 scores.

\paragraph{TCGA STAD (training)}
We collected N=326 primary cases from the TCGA Stomach Adenocarcinoma (STAD) cohort. They were only used for the training of \ac{cobra}.

\paragraph{CPTAC BRCA (testing)}
We collected N=120 primary cases from the CPTAC Breast Invasive Carcinoma (BRCA) cohort. For each case, we downloaded the corresponding molecular status: ER (N=120; 79 positive, 41 negative), PR (N=120; 70 positive, 50 negative), HER2 (N=120; 14 positive, 106
negative), and PIK3CA driver mutation (N=120; 82 WT, 38 MUT).

\paragraph{CPTAC COAD (testing)}
We collected N=110 primary cases from the CPTAC Colon Adenocarcinoma (COAD) cohort. For each case, we downloaded the corresponding molecular status: MSI status (N=105; 81 MSS, 24 MSI), Lymph Node status (N=110; 56 N0, 54 N+), CRC sidedness (N=108; 51 left, 57 right), BRAF (N=106; 91 WT, 15 MUT), KRAS (N=106; 71 WT, 35 MUT), and PIK3CA driver mutation (N=106; 87 WT, 19 MUT).

\paragraph{CPTAC LUAD (testing)}
We collected N=106 primary cases from the CPTAC Lung Adenocarcinoma (LUAD) cohort. For each case, we downloaded the corresponding molecular status: STK11 (N=106; 88 WT, 18 MUT), EGFR (N=106; 72 WT, 34 MUT), KRAS (N=106; 74 WT, 32 MUT), and TP53 driver mutation (N=106; 55 MUT, 51 WT).

\paragraph{CPTAC LUSC (testing)}
We collected N=108 primary cases from the CPTAC Lung Squamous Cell Carcinoma (LUSC) cohort and the aforementioned N=106 primary cases from the CPTAC LUAD cohort.

\section{Results}
\label{sec:supp-results}

\subsection{Full Classification}
\label{sec:supp-mlp-results}
Here, we provide the complete classification results of our experiments for the metrics \ac{auroc}, AUPRC, F1 score and balanced accuracy. \cref{tab:mlp_AUC-20x,tab:mlp_AUPRC-20x,tab:mlp_F1-20x,tab:mlp_Balanced Acc-20x} compare all models at 20$\times$ including \ac{cobra}-ENC, which was computed using the encoded embeddings ($\bm{H}_{S}$) with Virchow2 patch embeddings as shown in \cref{eq:aggregation_module}. In line with Wang et al.~\cite{Wang2024Chief}, using the original patch embeddings ($\bm{H}^{fe_n}$) is beneficial. \cref{tab:mlp_AUC-5x,tab:mlp_AUC-9x} show the complete AUC results at 5$\times$ and 9$\times$.

\subsection{Linear probing few-shot classification}
\label{sec:supp-lp-results}
\cref{tab:lp_auroc_5,tab:lp_auroc_10,tab:lp_auroc_25,tab:lp_auprc_5,tab:lp_auprc_10,tab:lp_auprc_25,tab:lp_f1_5,tab:lp_f1_10,tab:lp_f1_25,tab:lp_accuracy_5,tab:lp_accuracy_10,tab:lp_accuracy_25} show the complete results of our linear probing few-shot classification experiments for the metrics AUC, AUPRC, F1 score and balanced accuracy with k=5,10 and 25 samples per class.

\section{Heatmaps} \label{sec:supp-heatmaps}
\textsc{Cobra}’s approach to interpretability in WSI analysis is based on an aggregation method where each tile embedding is assigned a weight through a softmax-normalized attention score. These attention scores are used directly to compute a weighted average of the tile embeddings, yielding a slide-level representation that reflects the importance of each tile without requiring complex, non-linear transformations. Unlike GradCam\cite{Selvaraju_2019gradCam}-based interpretability methods used with tile embedding MIL approaches, \textsc{Cobra}’s attention scores are linearly applied to aggregate tile embeddings. This means that the attention scores correspond precisely to the actual weights used in generating the final slide embedding, allowing for direct interpretability without any intermediate non-linearities that might distort the contribution of each tile.

\begin{figure*}[ht!]
    \centering
    \includegraphics[width=0.75\linewidth]{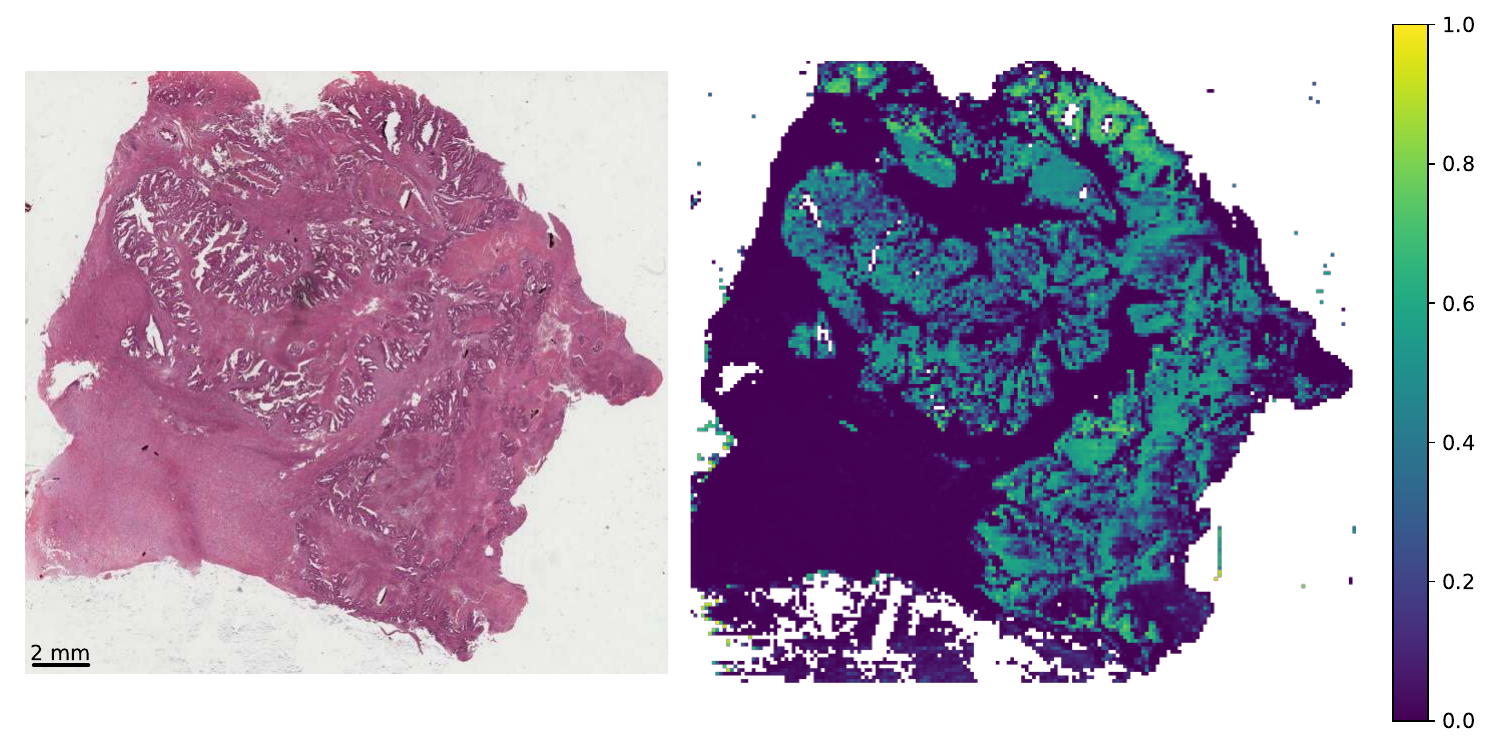}
    \caption{\textbf{\ac{cobra} Unsupervised Heatmap}. Patient: TCGA-CA-6715}
    \label{fig:heatmap-TCGA-CA-6715}
\end{figure*}

\begin{figure*}[ht!]
    \centering
    \includegraphics[width=0.75\linewidth]{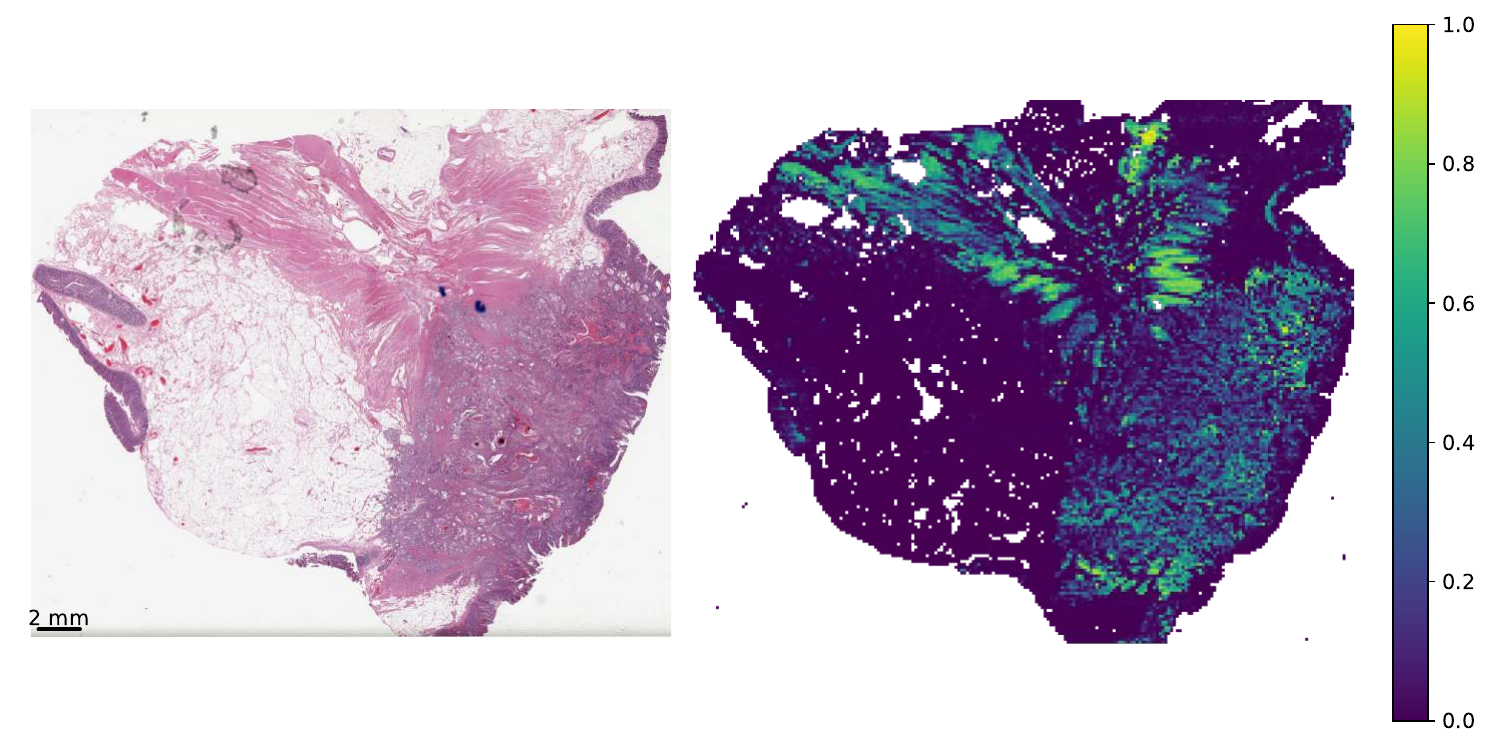}
    \caption{\textbf{\ac{cobra} Unsupervised Heatmap}. Patient: TCGA-CM-5349}
    \label{fig:heatmap-TCGA-CM-5349}
\end{figure*}

\begin{figure*}[ht!]
    \centering
    \includegraphics[width=0.75\linewidth]{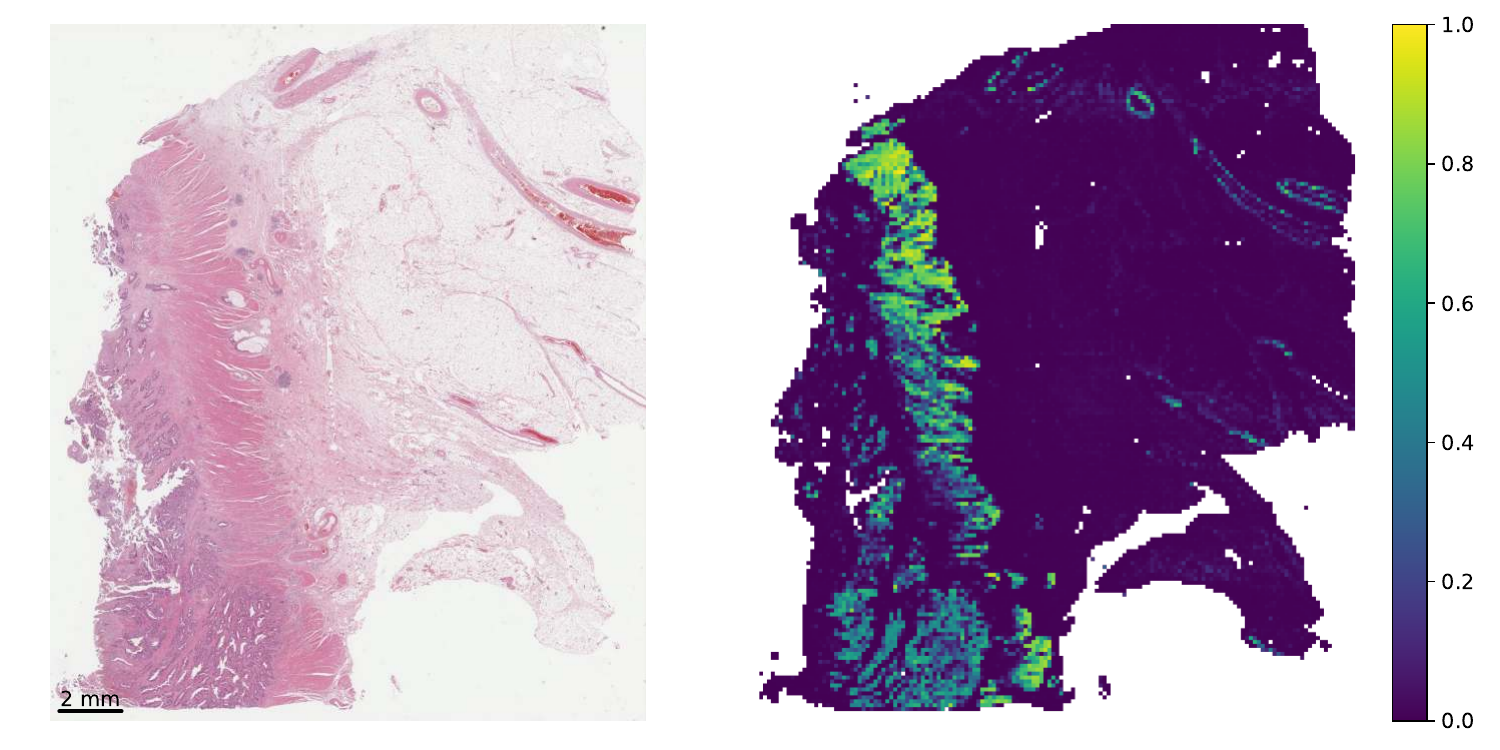}
    \caption{\textbf{\ac{cobra} Unsupervised Heatmap}. Patient: TCGA-EI-6508}
    \label{fig:heatmap-TCGA-EI-6508}
\end{figure*}

\begin{figure*}[ht!]
    \centering
    \includegraphics[width=0.75\linewidth]{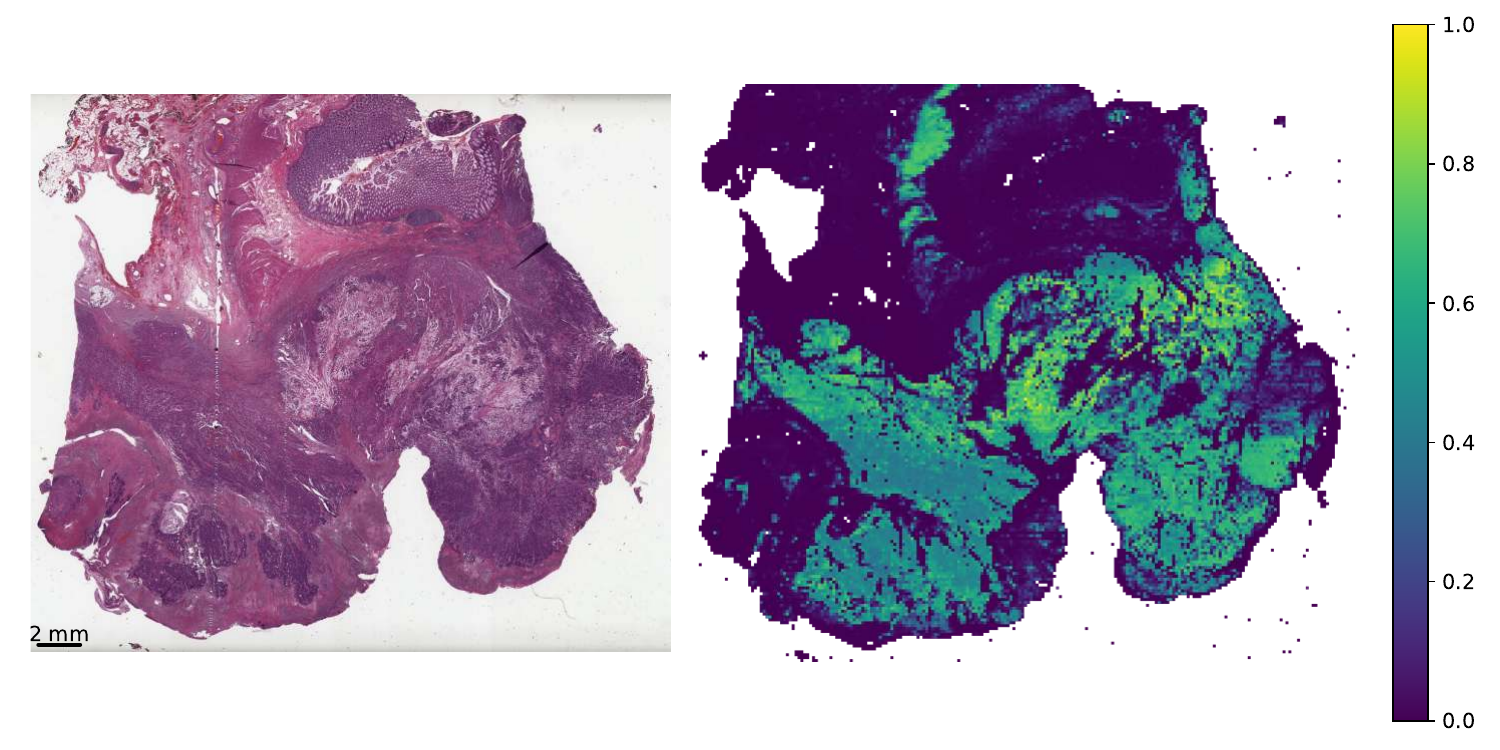}
    \caption{\textbf{\ac{cobra} Unsupervised Heatmap}. Patient: TCGA-CM-4743}
    \label{fig:heatmap-TCGA-CM-4743}
\end{figure*}

\begin{figure*}[ht!]
    \centering
    \includegraphics[width=0.75\linewidth]{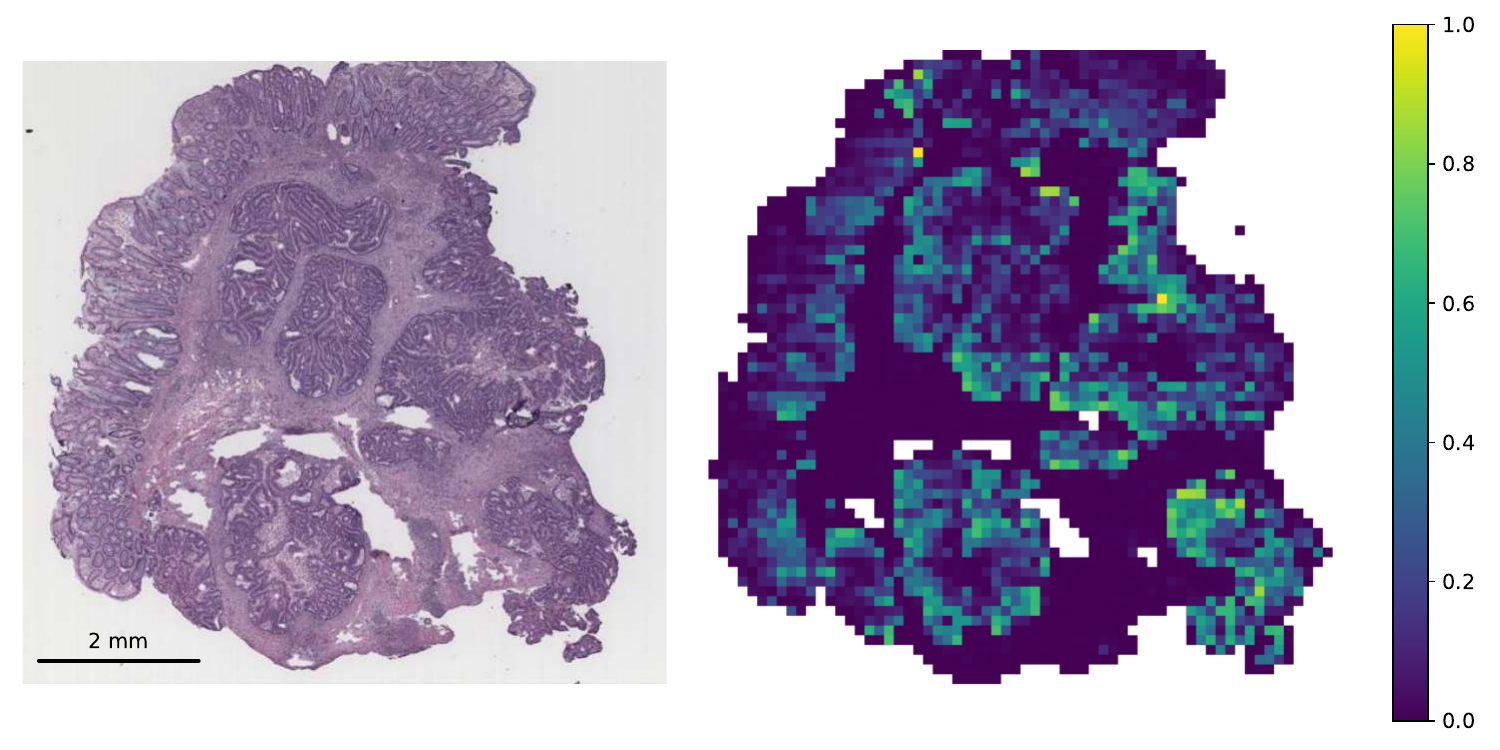}
    \caption{\textbf{\ac{cobra} Unsupervised Heatmap}. Patient: CPTAC-20CO007}
    \label{fig:heatmap-CPTAC-20CO007}
\end{figure*}

\begin{figure*}[ht!]
    \centering
    \includegraphics[width=0.75\linewidth]{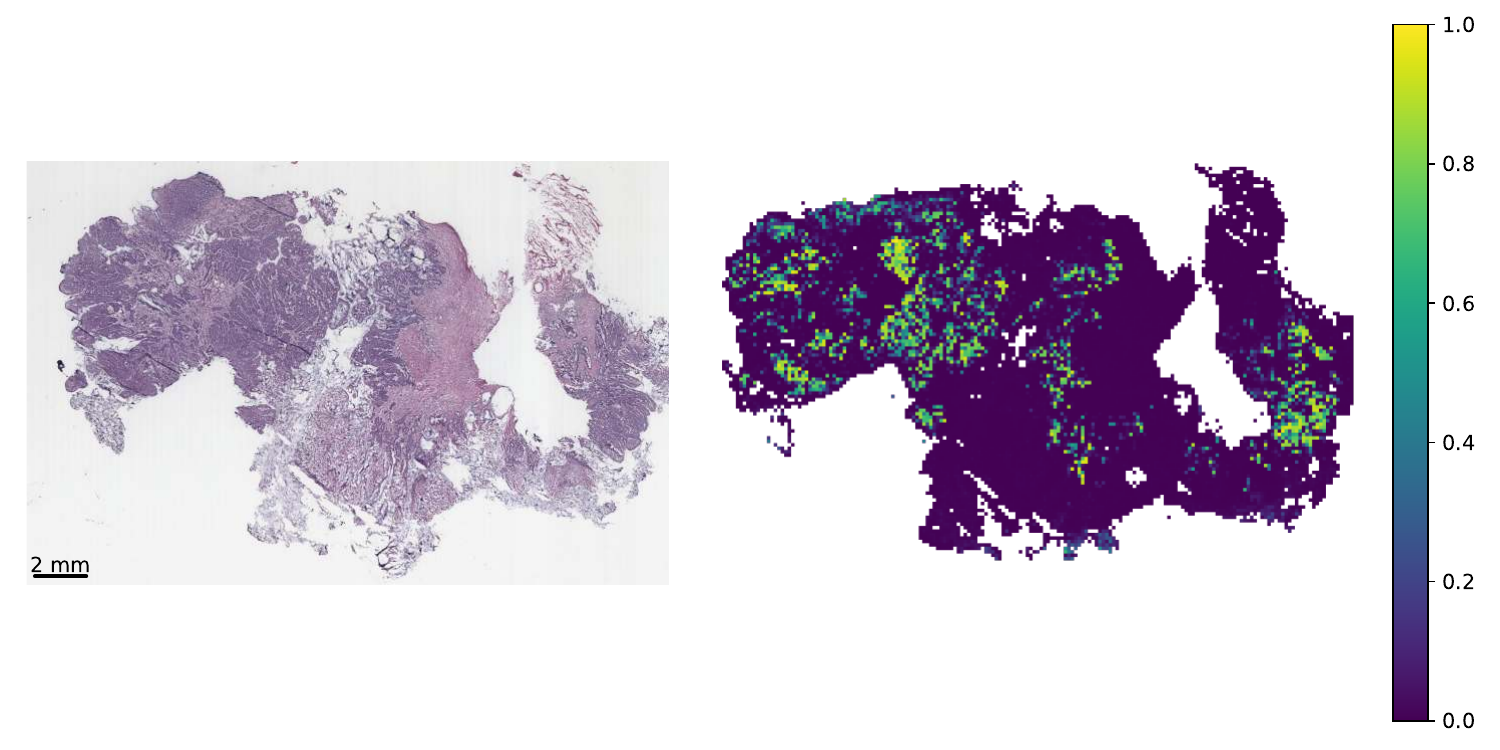}
    \caption{\textbf{\ac{cobra} Unsupervised Heatmap}. Patient: CPTAC-11CO062}
    \label{fig:heatmap-CPTAC-11CO062}
\end{figure*}

In \cref{fig:heatmap-TCGA-CA-6715,fig:heatmap-TCGA-CM-5349,fig:heatmap-TCGA-EI-6508,fig:heatmap-TCGA-CM-4743}, we provide interpretability heatmaps for slides from TCGA-CRC and in \cref{fig:heatmap-CPTAC-20CO007,fig:heatmap-CPTAC-11CO062}, we show interpretability heatmaps for slides from CPTAC-COAD. These heatmaps display the attention values across the slide, with tiles associated with higher attention scores consistently aligning with tumor regions. In contrast, non-tumorous areas and background regions receive lower attention values. This pattern demonstrates \textsc{Cobra}’s capability to emphasize diagnostically relevant areas based solely on the unsupervised training with tile embeddings.

While this tile-based attention approach lacks the spatial precision of pixel-level methods, it offers a computationally efficient way to highlight regions of model focus. By operating directly on tile embeddings, \ac{cobra} can produce interpretable heatmaps that outline primary areas of interest, indicating its utility in scenarios where rapid, general interpretability is more practical than fine-grained spatial resolution.

\begin{table*}[ht!]
\centering
\caption{\textbf{Classification performance comparison.} AUC score of models trained on TCGA deployed on CPTAC datasets. $\overline{\text{Overline}}$ indicates mean over patch embeddings, $^\dag$ indicates that embeddings of all four training FMs were used to generate the weighting vector (\cref{eq:cobra-dagger}). For the other \ac{cobra} entries, we used the inference mode from (\cref{eq:aggregation_module_inference}). \textbf{Bold} indicates the best performance, and $\underline{\text{underline}}$ indicates the second-best performance. The abbreviations are as follows: ST: Subtyping, CTP: CTransPath~\cite{wang2022ctp}, H0: H-Optimus-0~\cite{hoptimus0}, V2: Virchow-2~\cite{zimmermann2024virchow2}, GP: GigaPath~\cite{xu2024gigapath}, SE: Slide Encoder.}
\label{tab:mlp_AUC-20x}
\resizebox{\textwidth}{!}{%
\begin{tabular}{l|l|llll|llll|llllll|l}
\toprule
AUC-20$\times$[\%] & NSCLC & \multicolumn{4}{c|}{LUAD} & \multicolumn{4}{c|}{BRCA} & \multicolumn{6}{c|}{COAD} & Average \\
Model & ST & STK11 & EGFR & TP53 & KRAS & ESR1 & PGR & ERBB2 & PIK3CA & MSI & BRAF & LN & KRAS & Side & PIK3CA &  \\
\midrule
$\overline{\text{CTransPath}}$~\cite{wang2022ctp} & $87.2_{1.5}$ & $62.8_{2.5}$ & $59.3_{7.4}$ & $70.1_{2.3}$ & $52.4_{5.9}$ & $68.1_{2.5}$ & $66.5_{2.1}$ & $48.6_{1.5}$ & $56.3_{3.0}$ & $76.1_{4.6}$ & $59.8_{2.3}$ & $59.8_{1.0}$ & $55.9_{7.7}$ & $52.5_{2.6}$ & $56.3_{6.3}$ & $62.1_{4.2}$ \\
$\overline{\text{Virchow}}$~\cite{vorontsov2024virchow} & $89.4_{0.6}$ & $76.5_{6.8}$ & $60.3_{2.0}$ & $70.7_{1.7}$ & $54.3_{6.9}$ & $66.9_{4.8}$ & $60.5_{3.9}$ & $51.3_{5.7}$ & $63.5_{3.7}$ & $62.1_{6.7}$ & $65.0_{2.6}$ & $58.2_{5.1}$ & $53.9_{7.1}$ & $52.3_{3.1}$ & $52.4_{5.7}$ & $62.5_{4.9}$ \\
$\overline{\text{CONCH}}$~\cite{lu2024conch} & $96.5_{0.3}$ & $66.0_{10.3}$ & $62.0_{7.6}$ & $74.6_{1.6}$ & $59.0_{7.4}$ & $85.3_{1.5}$ & $80.3_{2.0}$ & $58.8_{11.0}$ & $63.2_{3.1}$ & $79.2_{0.5}$ & $57.5_{3.4}$ & $\mathbf{67.3}_{2.0}$ & $55.7_{8.6}$ & $53.4_{2.4}$ & $63.2_{5.3}$ & $68.1_{5.7}$ \\
$\overline{\text{UNI}}$~\cite{chen2024uni} & $95.8_{1.1}$ & $69.4_{2.4}$ & $70.1_{12.1}$ & $73.9_{0.8}$ & $50.7_{4.7}$ & $87.4_{3.1}$ & $74.9_{2.3}$ & $64.0_{3.4}$ & $62.3_{5.0}$ & $89.0_{1.6}$ & $73.0_{3.4}$ & $62.0_{7.8}$ & $63.5_{2.8}$ & $59.4_{3.7}$ & $63.5_{6.2}$ & $70.6_{4.9}$ \\
$\overline{\text{H-Optimus}}$~\cite{hoptimus0} & $97.2_{0.4}$ & $78.5_{2.7}$ & $78.2_{3.5}$ & $71.3_{1.1}$ & $58.1_{3.7}$ & $85.2_{2.7}$ & $74.9_{3.3}$ & $51.4_{4.5}$ & $59.5_{4.9}$ & $94.7_{0.7}$ & $77.1_{7.1}$ & $55.5_{3.6}$ & $59.2_{4.2}$ & $\mathbf{62.2}_{3.4}$ & $62.4_{9.0}$ & $71.0_{4.3}$ \\
$\overline{\text{GigaPath}}$~\cite{xu2024gigapath} & $96.6_{0.7}$ & $71.3_{1.9}$ & $75.7_{6.8}$ & $75.4_{1.3}$ & $56.9_{6.2}$ & $85.7_{1.1}$ & $75.9_{1.8}$ & $64.5_{2.7}$ & $62.2_{5.9}$ & $93.3_{1.6}$ & $77.5_{2.6}$ & $61.7_{2.9}$ & $56.1_{5.0}$ & $60.0_{1.5}$ & $59.4_{7.9}$ & $71.5_{4.0}$ \\
$\overline{\text{Ensemble Prediction}}$ & $97.2_{0.3}$ & $77.2_{3.7}$ & $78.5_{4.1}$ & $73.3_{0.6}$ & $\underline{59.5}_{5.1}$ & $87.6_{2.8}$ & $77.2_{2.7}$ & $65.6_{2.0}$ & $63.3_{4.1}$ & $94.7_{1.0}$ & $78.9_{5.4}$ & $62.5_{3.9}$ & $\underline{64.1}_{3.0}$ & $60.5_{2.1}$ & $\underline{64.5}_{9.1}$ & $73.6_{4.0}$ \\
$\overline{\text{Virchow2}}$~\cite{zimmermann2024virchow2} & $95.8_{0.7}$ & $79.6_{5.5}$ & $78.3_{4.6}$ & $72.1_{0.7}$ & $\mathbf{60.9}_{5.6}$ & $89.2_{2.8}$ & $79.3_{2.7}$ & $71.3_{1.8}$ & $63.2_{4.8}$ & $94.9_{1.2}$ & $81.6_{4.5}$ & $63.0_{1.9}$ & $59.3_{6.2}$ & $56.3_{3.8}$ & $62.7_{11.6}$ & $73.8_{4.7}$ \\
$\overline{\text{Concatenated}}$ & $97.4_{0.4}$ & $75.7_{3.0}$ & $\underline{80.2}_{2.2}$ & $72.5_{0.8}$ & $57.6_{4.8}$ & $\underline{89.6}_{1.4}$ & $79.1_{3.4}$ & $67.5_{3.9}$ & $61.8_{4.0}$ & $95.0_{1.1}$ & $82.2_{4.3}$ & $61.6_{2.5}$ & $59.7_{5.8}$ & $\underline{62.0}_{2.4}$ & $\mathbf{70.2}_{4.1}$ & $74.1_{3.3}$ \\
\hline
\textsc{{Cobra}}-ENC & $93.1_{0.3}$ & $65.6_{2.4}$ & $68.7_{2.3}$ & $72.0_{1.9}$ & $53.8_{2.8}$ & $71.1_{0.9}$ & $68.1_{2.9}$ & $62.9_{4.1}$ & $62.3_{4.1}$ & $54.9_{6.5}$ & $60.8_{9.0}$ & $50.0_{2.7}$ & $45.6_{1.6}$ & $45.6_{2.5}$ & $52.1_{2.1}$ & $61.8_{3.7}$ \\
GigaPath-SE~\cite{xu2024gigapath} & $90.9_{1.3}$ & $67.0_{4.4}$ & $65.4_{4.4}$ & $73.7_{1.4}$ & $57.1_{5.2}$ & $72.9_{0.9}$ & $71.9_{3.3}$ & $55.4_{4.7}$ & $60.5_{4.6}$ & $66.2_{2.1}$ & $56.7_{4.5}$ & $54.6_{5.2}$ & $51.3_{2.9}$ & $45.8_{3.1}$ & $53.2_{5.7}$ & $62.8_{3.9}$ \\
MADELEINE~\cite{jaume2024madeleine} & $94.0_{0.6}$ & $72.2_{8.7}$ & $64.0_{6.7}$ & $72.0_{2.8}$ & $51.9_{3.9}$ & $80.1_{1.7}$ & $73.7_{1.3}$ & $66.7_{2.7}$ & $64.9_{1.6}$ & $68.6_{9.1}$ & $54.2_{6.7}$ & $60.3_{7.3}$ & $58.9_{6.6}$ & $50.5_{1.6}$ & $59.5_{8.6}$ & $66.1_{5.5}$ \\
CHIEF~\cite{Wang2024Chief} & $93.6_{0.8}$ & $64.2_{10.7}$ & $62.8_{10.9}$ & $73.4_{1.5}$ & $50.1_{5.0}$ & $83.0_{0.5}$ & $77.5_{0.3}$ & $63.4_{2.3}$ & $\underline{65.4}_{1.5}$ & $75.1_{4.8}$ & $63.6_{4.3}$ & $58.0_{1.7}$ & $58.4_{3.8}$ & $48.2_{4.2}$ & $56.6_{3.2}$ & $66.2_{4.9}$ \\
\textsc{{Cobra}}$^\dag$-CTP & $95.9_{0.6}$ & $68.1_{5.1}$ & $69.2_{4.7}$ & $75.1_{1.6}$ & $46.7_{4.1}$ & $77.9_{0.8}$ & $71.3_{1.3}$ & $59.3_{1.5}$ & $59.2_{1.3}$ & $80.3_{1.5}$ & $73.5_{3.4}$ & $60.6_{2.5}$ & $55.2_{4.5}$ & $48.3_{3.4}$ & $54.3_{5.2}$ & $66.3_{3.2}$ \\
\textsc{{Cobra}}-CTP & $95.9_{0.6}$ & $65.0_{10.5}$ & $66.0_{5.5}$ & $74.8_{1.8}$ & $49.2_{4.4}$ & $78.6_{0.7}$ & $72.2_{0.6}$ & $62.0_{1.4}$ & $60.9_{3.6}$ & $80.3_{2.2}$ & $73.2_{3.0}$ & $61.3_{1.6}$ & $52.4_{4.2}$ & $48.1_{2.3}$ & $56.3_{4.3}$ & $66.4_{4.0}$ \\
PRISM~\cite{shaikovski2024prism} & $99.2_{0.1}$ & $\mathbf{87.6}_{1.6}$ & $70.7_{2.4}$ & $78.2_{0.5}$ & $52.9_{8.5}$ & $\mathbf{92.2}_{0.7}$ & $\mathbf{84.2}_{0.5}$ & $64.5_{6.0}$ & $\mathbf{69.4}_{2.1}$ & $79.1_{1.5}$ & $59.9_{1.4}$ & $\underline{67.2}_{2.4}$ & $54.6_{6.2}$ & $52.2_{1.8}$ & $52.1_{6.8}$ & $70.9_{3.8}$ \\
\textsc{{Cobra}}$^\dag$-UNI & $99.1_{0.2}$ & $79.1_{2.8}$ & $76.2_{4.6}$ & $\underline{80.2}_{0.7}$ & $55.0_{5.7}$ & $86.0_{1.7}$ & $78.1_{3.2}$ & $60.3_{4.9}$ & $62.3_{3.1}$ & $89.1_{0.7}$ & $83.5_{1.5}$ & $65.7_{2.1}$ & $\mathbf{65.3}_{4.2}$ & $57.2_{2.1}$ & $61.9_{2.1}$ & $73.3_{3.1}$ \\
\textsc{{Cobra}}$^\dag$-H0 & $\underline{\mathbf{99.4}}_{0.1}$ & $\underline{86.9}_{2.0}$ & $\mathbf{80.9}_{3.4}$ & $79.9_{1.8}$ & $56.7_{3.6}$ & $87.8_{1.2}$ & $72.8_{3.4}$ & $59.9_{2.1}$ & $58.0_{0.9}$ & $\underline{95.2}_{1.1}$ & $84.9_{3.7}$ & $58.1_{2.6}$ & $59.7_{7.2}$ & $58.5_{2.4}$ & $61.2_{3.4}$ & $73.3_{3.1}$ \\
\textsc{{Cobra}}-UNI & $98.8_{0.3}$ & $79.4_{2.5}$ & $76.5_{5.3}$ & $78.9_{1.2}$ & $52.2_{4.2}$ & $88.1_{1.7}$ & $\underline{80.5}_{3.0}$ & $65.1_{4.3}$ & $63.9_{5.2}$ & $89.1_{1.1}$ & $82.8_{1.5}$ & $64.6_{2.2}$ & $59.0_{8.4}$ & $57.4_{2.1}$ & $64.3_{5.8}$ & $73.4_{3.9}$ \\
\textsc{{Cobra}}-H0 & $\underline{\mathbf{99.4}}_{0.2}$ & $86.5_{1.8}$ & $79.9_{2.9}$ & $80.1_{2.4}$ & $54.3_{4.7}$ & $87.1_{1.0}$ & $74.0_{4.2}$ & $64.2_{4.9}$ & $55.7_{2.3}$ & $\mathbf{96.0}_{0.6}$ & $86.2_{3.3}$ & $58.2_{2.5}$ & $62.2_{4.4}$ & $57.2_{1.9}$ & $62.9_{4.7}$ & $73.6_{3.2}$ \\
\textsc{{Cobra}}$^\dag$-GP & $98.9_{0.3}$ & $81.5_{2.3}$ & $78.7_{4.2}$ & $\mathbf{80.9}_{1.1}$ & $56.9_{5.3}$ & $87.8_{1.2}$ & $77.5_{1.1}$ & $65.4_{1.3}$ & $64.6_{3.8}$ & $93.5_{1.2}$ & $85.6_{2.0}$ & $64.7_{2.4}$ & $59.2_{6.6}$ & $57.4_{2.1}$ & $56.9_{9.4}$ & $74.0_{3.8}$ \\
\textsc{{Cobra}}-V2 & $98.1_{0.2}$ & $84.0_{2.9}$ & $80.0_{2.4}$ & $78.4_{2.9}$ & $59.2_{6.2}$ & $\underline{89.6}_{2.0}$ & $79.2_{2.4}$ & $\underline{71.6}_{2.2}$ & $63.6_{6.2}$ & $94.1_{0.5}$ & $\underline{87.8}_{2.0}$ & $65.7_{2.5}$ & $62.1_{10.4}$ & $58.3_{1.9}$ & $57.6_{7.5}$ & $\underline{75.3}_{4.4}$ \\
\textsc{{Cobra}}$^\dag$-V2 & $98.4_{0.2}$ & $84.6_{1.9}$ & $78.9_{3.6}$ & $78.4_{2.6}$ & $55.9_{7.1}$ & $\underline{89.6}_{1.7}$ & $80.0_{2.3}$ & $\mathbf{72.2}_{1.7}$ & $65.1_{4.5}$ & $94.2_{0.6}$ & $\mathbf{88.7}_{1.6}$ & $64.8_{2.3}$ & $60.6_{5.7}$ & $58.6_{1.7}$ & $61.5_{4.1}$ & $\mathbf{75.4}_{3.3}$ \\
\bottomrule
\end{tabular}
}
\end{table*}

\begin{table*}[ht!]
\centering
\caption{\textbf{Classification performance comparison.} AUPRC score of models trained on TCGA deployed on CPTAC datasets. $\overline{\text{Overline}}$ indicates mean over patch embeddings, $^\dag$ indicates that embeddings of all four training FMs were used to generate the weighting vector (\cref{eq:cobra-dagger}). For the other \ac{cobra} entries, we used the inference mode from (\cref{eq:aggregation_module_inference}). \textbf{Bold} indicates the best performance, and $\underline{\text{underline}}$ indicates the second-best performance. The abbreviations are as follows: ST: Subtyping, CTP: CTransPath~\cite{wang2022ctp}, H0: H-Optimus-0~\cite{hoptimus0}, V2: Virchow-2~\cite{zimmermann2024virchow2}, GP: GigaPath~\cite{xu2024gigapath}, SE: Slide Encoder.}
\label{tab:mlp_AUPRC-20x}
\resizebox{\textwidth}{!}{%
\begin{tabular}{l|l|llll|llll|llllll|l}
\toprule
AUPRC-20$\times$[\%] & NSCLC & \multicolumn{4}{c|}{LUAD} & \multicolumn{4}{c|}{BRCA} & \multicolumn{6}{c|}{COAD} & Average \\
Model & ST & STK11 & EGFR & TP53 & KRAS & ESR1 & PGR & ERBB2 & PIK3CA & MSI & BRAF & LN & KRAS & Side & PIK3CA &  \\
\midrule
$\overline{\text{Virchow}}$~\cite{vorontsov2024virchow} & $90.3_{0.6}$ & $36.3_{10.2}$ & $43.1_{3.6}$ & $68.2_{0.8}$ & $40.1_{3.2}$ & $78.4_{2.8}$ & $68.4_{2.8}$ & $17.8_{1.5}$ & $46.1_{6.4}$ & $35.7_{8.0}$ & $25.8_{4.1}$ & $57.8_{3.5}$ & $40.2_{8.8}$ & $57.4_{4.0}$ & $21.4_{4.4}$ & $48.5_{5.1}$ \\
$\overline{\text{CTransPath}}$~\cite{wang2022ctp} & $87.9_{1.5}$ & $25.8_{2.0}$ & $46.4_{8.3}$ & $71.2_{2.3}$ & $35.5_{2.5}$ & $79.3_{1.6}$ & $73.8_{1.6}$ & $11.4_{0.3}$ & $39.7_{3.3}$ & $59.0_{8.7}$ & $27.7_{1.2}$ & $59.2_{1.3}$ & $45.2_{8.4}$ & $57.7_{2.3}$ & $25.2_{4.0}$ & $49.7_{4.3}$ \\
$\overline{\text{CONCH}}$~\cite{lu2024conch} & $96.7_{0.3}$ & $32.3_{10.0}$ & $44.7_{8.3}$ & $78.0_{1.0}$ & $39.0_{4.5}$ & $92.5_{0.7}$ & $\underline{84.4}_{1.0}$ & $18.6_{4.8}$ & $47.9_{3.5}$ & $62.5_{0.8}$ & $28.2_{5.2}$ & $\mathbf{70.6}_{3.0}$ & $40.4_{7.5}$ & $59.1_{2.1}$ & $30.4_{7.3}$ & $55.0_{5.0}$ \\
$\overline{\text{UNI}}$~\cite{chen2024uni} & $96.1_{0.9}$ & $26.9_{2.4}$ & $58.9_{13.5}$ & $72.6_{1.7}$ & $36.9_{5.3}$ & $93.1_{1.9}$ & $79.8_{1.8}$ & $20.2_{4.5}$ & $40.0_{3.5}$ & $75.5_{2.8}$ & $35.5_{3.3}$ & $63.1_{7.8}$ & $50.4_{4.8}$ & $64.0_{3.2}$ & $35.4_{4.8}$ & $56.6_{5.1}$ \\
$\overline{\text{H-Optimus}}$~\cite{hoptimus0} & $97.3_{0.3}$ & $38.3_{4.5}$ & $68.2_{5.0}$ & $68.5_{1.1}$ & $40.4_{2.1}$ & $91.2_{1.7}$ & $81.2_{3.1}$ & $13.6_{2.2}$ & $38.3_{5.0}$ & $89.2_{1.7}$ & $38.7_{8.2}$ & $53.2_{5.9}$ & $46.6_{7.0}$ & $64.2_{2.0}$ & $30.7_{6.4}$ & $57.3_{4.4}$ \\
$\overline{\text{GigaPath}}$~\cite{xu2024gigapath} & $96.6_{0.7}$ & $29.9_{3.1}$ & $68.4_{7.6}$ & $72.6_{2.3}$ & $40.4_{2.9}$ & $92.4_{0.6}$ & $80.2_{1.1}$ & $20.5_{5.1}$ & $43.9_{7.5}$ & $82.8_{2.2}$ & $39.3_{2.5}$ & $62.4_{4.5}$ & $42.6_{6.8}$ & $64.4_{1.3}$ & $28.7_{9.5}$ & $57.7_{4.7}$ \\
$\overline{\text{Concatenated}}$ & $97.4_{0.4}$ & $34.2_{4.9}$ & $69.7_{2.6}$ & $69.6_{1.8}$ & $42.1_{2.9}$ & $94.4_{1.0}$ & $83.1_{2.6}$ & $21.0_{3.6}$ & $38.7_{2.8}$ & $86.5_{2.9}$ & $40.9_{5.2}$ & $59.7_{4.8}$ & $47.4_{9.5}$ & $\mathbf{65.2}_{0.8}$ & $\mathbf{39.3}_{5.6}$ & $59.3_{4.1}$ \\
$\overline{\text{Ensemble Prediction}}$ & $97.3_{0.3}$ & $36.4_{5.6}$ & $69.9_{5.1}$ & $71.4_{1.4}$ & $\underline{42.9}_{2.4}$ & $93.0_{2.0}$ & $82.4_{2.5}$ & $20.0_{2.7}$ & $40.0_{3.4}$ & $85.6_{1.6}$ & $38.3_{3.6}$ & $61.4_{6.2}$ & $\mathbf{50.7}_{6.5}$ & $63.9_{1.4}$ & $35.7_{6.0}$ & $59.3_{3.9}$ \\
$\overline{\text{Virchow2}}$~\cite{zimmermann2024virchow2} & $95.9_{0.5}$ & $40.4_{8.9}$ & $\underline{70.2}_{5.4}$ & $70.7_{1.5}$ & $\mathbf{44.7}_{5.6}$ & $94.6_{2.0}$ & $83.0_{2.1}$ & $26.2_{5.9}$ & $40.5_{6.0}$ & $84.9_{1.6}$ & $41.5_{7.3}$ & $65.7_{2.7}$ & $47.3_{8.8}$ & $62.0_{4.6}$ & $32.7_{10.5}$ & $60.0_{5.7}$ \\
\hline
GigaPath-SE~\cite{xu2024gigapath} & $91.3_{1.1}$ & $28.8_{2.7}$ & $51.4_{1.5}$ & $69.9_{2.0}$ & $40.0_{3.0}$ & $80.5_{1.0}$ & $73.0_{2.6}$ & $17.2_{2.2}$ & $42.6_{3.2}$ & $39.9_{4.2}$ & $18.4_{1.9}$ & $55.3_{5.2}$ & $38.3_{3.3}$ & $51.8_{2.5}$ & $22.7_{5.0}$ & $48.1_{3.0}$ \\
\textsc{{Cobra}}-ENC & $92.7_{0.5}$ & $26.5_{3.4}$ & $50.2_{3.4}$ & $73.2_{3.0}$ & $35.5_{2.3}$ & $82.4_{0.8}$ & $77.8_{2.5}$ & $26.5_{2.1}$ & $41.1_{2.9}$ & $31.2_{6.8}$ & $31.2_{9.4}$ & $52.0_{3.5}$ & $31.0_{1.2}$ & $51.4_{2.5}$ & $21.2_{3.1}$ & $48.3_{3.8}$ \\
CHIEF~\cite{Wang2024Chief} & $94.7_{0.5}$ & $26.5_{7.3}$ & $52.5_{11.2}$ & $73.2_{1.7}$ & $32.7_{1.0}$ & $90.0_{0.5}$ & $82.1_{0.5}$ & $17.7_{1.5}$ & $\underline{51.4}_{2.9}$ & $56.3_{6.7}$ & $30.9_{3.4}$ & $56.5_{2.3}$ & $45.3_{5.8}$ & $54.3_{4.3}$ & $24.9_{3.8}$ & $52.6_{4.6}$ \\
\textsc{{Cobra}}-CTP & $96.4_{0.4}$ & $27.1_{6.0}$ & $55.7_{6.3}$ & $74.2_{1.4}$ & $33.0_{3.5}$ & $87.6_{0.4}$ & $78.9_{0.7}$ & $17.9_{1.1}$ & $46.2_{5.2}$ & $64.2_{3.8}$ & $37.1_{5.1}$ & $60.6_{1.5}$ & $41.0_{4.0}$ & $54.0_{2.1}$ & $23.4_{2.4}$ & $53.2_{3.5}$ \\
\textsc{{Cobra}}$^\dag$-CTP & $96.6_{0.4}$ & $27.7_{3.0}$ & $57.2_{4.8}$ & $74.2_{1.1}$ & $34.0_{2.7}$ & $86.7_{0.6}$ & $78.3_{0.7}$ & $18.4_{0.5}$ & $44.4_{2.2}$ & $66.0_{3.8}$ & $40.4_{3.7}$ & $61.0_{2.6}$ & $43.9_{6.3}$ & $55.2_{2.9}$ & $23.8_{3.5}$ & $53.9_{3.1}$ \\
MADELEINE~\cite{jaume2024madeleine} & $94.6_{0.6}$ & $46.3_{8.6}$ & $48.7_{9.5}$ & $74.3_{1.2}$ & $34.3_{2.9}$ & $88.7_{1.0}$ & $81.0_{0.8}$ & $26.0_{2.5}$ & $\mathbf{52.1}_{1.3}$ & $50.1_{13.5}$ & $28.0_{8.3}$ & $59.5_{7.4}$ & $44.0_{5.6}$ & $55.1_{3.3}$ & $30.4_{6.5}$ & $54.2_{6.2}$ \\
PRISM~\cite{shaikovski2024prism} & $99.3_{0.0}$ & $\mathbf{51.3}_{3.5}$ & $61.0_{2.7}$ & $70.8_{0.8}$ & $36.5_{6.7}$ & $\mathbf{95.3}_{0.4}$ & $\mathbf{86.9}_{1.0}$ & $19.1_{3.3}$ & $47.2_{3.5}$ & $58.0_{3.3}$ & $29.3_{2.0}$ & $65.5_{3.7}$ & $39.5_{8.7}$ & $60.4_{0.9}$ & $25.1_{4.4}$ & $56.3_{3.8}$ \\
\textsc{{Cobra}}$^\dag$-UNI & $99.1_{0.2}$ & $35.3_{3.6}$ & $65.6_{5.2}$ & $\mathbf{80.7}_{1.2}$ & $36.9_{0.9}$ & $91.7_{1.5}$ & $82.5_{2.5}$ & $19.4_{2.8}$ & $41.6_{2.5}$ & $77.4_{1.0}$ & $45.5_{4.3}$ & $\underline{67.4}_{2.6}$ & $50.4_{6.6}$ & $61.6_{2.2}$ & $32.1_{2.8}$ & $59.1_{3.1}$ \\
\textsc{{Cobra}}-UNI & $98.9_{0.3}$ & $35.7_{3.3}$ & $64.4_{5.5}$ & $78.9_{1.9}$ & $36.0_{3.6}$ & $93.2_{1.3}$ & $\underline{84.4}_{2.9}$ & $21.2_{3.8}$ & $44.8_{6.6}$ & $77.0_{2.2}$ & $44.8_{2.8}$ & $66.4_{3.9}$ & $44.1_{10.5}$ & $63.0_{1.6}$ & $\underline{36.4}_{4.1}$ & $59.3_{4.3}$ \\
\textsc{{Cobra}}$^\dag$-H0 & $\underline{99.4}_{0.1}$ & $\underline{50.3}_{2.8}$ & $\mathbf{70.8}_{3.3}$ & $79.6_{2.4}$ & $41.0_{2.5}$ & $92.6_{0.7}$ & $80.1_{2.2}$ & $14.9_{0.7}$ & $36.3_{1.0}$ & $\underline{89.3}_{1.8}$ & $46.4_{4.8}$ & $56.9_{3.7}$ & $47.0_{10.8}$ & $62.0_{2.1}$ & $27.9_{2.9}$ & $59.6_{3.7}$ \\
\textsc{{Cobra}}-H0 & $\mathbf{99.5}_{0.2}$ & $49.2_{2.2}$ & $69.8_{3.0}$ & $79.1_{3.1}$ & $38.2_{1.7}$ & $91.9_{1.0}$ & $80.6_{3.3}$ & $17.6_{2.6}$ & $34.9_{2.8}$ & $\mathbf{91.0}_{1.2}$ & $48.9_{5.3}$ & $57.7_{3.0}$ & $\underline{50.6}_{7.0}$ & $59.9_{1.1}$ & $29.2_{5.6}$ & $59.9_{3.4}$ \\
\textsc{{Cobra}}$^\dag$-GP & $98.9_{0.3}$ & $40.2_{3.4}$ & $70.1_{4.4}$ & $\underline{79.8}_{0.7}$ & $38.6_{4.2}$ & $93.5_{0.6}$ & $81.8_{0.6}$ & $26.6_{4.2}$ & $48.8_{4.8}$ & $87.1_{2.2}$ & $49.0_{3.7}$ & $65.4_{2.8}$ & $44.8_{10.9}$ & $62.2_{2.3}$ & $28.4_{7.1}$ & $61.0_{4.4}$ \\
\textsc{{Cobra}}-V2 & $98.1_{0.2}$ & $44.3_{6.1}$ & $70.1_{1.9}$ & $78.7_{2.0}$ & $41.0_{5.0}$ & $94.5_{1.2}$ & $82.1_{2.1}$ & $\underline{30.0}_{4.0}$ & $46.4_{9.9}$ & $85.2_{2.0}$ & $\underline{54.2}_{6.2}$ & $66.4_{2.9}$ & $50.1_{12.9}$ & $64.0_{1.1}$ & $25.4_{6.7}$ & $\underline{62.0}_{5.5}$ \\
\textsc{{Cobra}}$^\dag$-V2 & $98.4_{0.2}$ & $46.7_{4.8}$ & $69.5_{2.7}$ & $77.9_{1.7}$ & $39.3_{4.5}$ & $\underline{94.8}_{0.8}$ & $83.0_{1.5}$ & $\mathbf{30.7}_{5.5}$ & $47.6_{8.5}$ & $86.6_{1.4}$ & $\mathbf{54.3}_{4.7}$ & $66.8_{2.1}$ & $\underline{50.6}_{10.4}$ & $\underline{64.8}_{1.1}$ & $32.2_{5.9}$ & $\mathbf{62.9}_{4.7}$ \\
\bottomrule
\end{tabular}
}
\end{table*}

\begin{table*}[ht!]
\centering
\caption{\textbf{Classification performance comparison.} F1 score of models trained on TCGA deployed on CPTAC datasets. $\overline{\text{Overline}}$ indicates mean over patch embeddings, $^\dag$ indicates that embeddings of all four training FMs were used to generate the weighting vector (\cref{eq:cobra-dagger}). For the other \ac{cobra} entries, we used the inference mode from (\cref{eq:aggregation_module_inference}). \textbf{Bold} indicates the best performance, and $\underline{\text{underline}}$ indicates the second-best performance. The abbreviations are as follows: ST: Subtyping, CTP: CTransPath~\cite{wang2022ctp}, H0: H-Optimus-0~\cite{hoptimus0}, V2: Virchow-2~\cite{zimmermann2024virchow2}, GP: GigaPath~\cite{xu2024gigapath}, SE: Slide Encoder.}
\label{tab:mlp_F1-20x}
\resizebox{\textwidth}{!}{%
\begin{tabular}{l|l|llll|llll|llllll|l}
\toprule
F1-20$\times$[\%] & NSCLC & \multicolumn{4}{c|}{LUAD} & \multicolumn{4}{c|}{BRCA} & \multicolumn{6}{c|}{COAD} & Average \\
Model & ST & STK11 & EGFR & TP53 & KRAS & ESR1 & PGR & ERBB2 & PIK3CA & MSI & BRAF & LN & KRAS & Side & PIK3CA &  \\
\midrule
$\overline{\text{Virchow}}$~\cite{vorontsov2024virchow} & $78.2_{2.0}$ & $2.1_{4.2}$ & $3.3_{4.4}$ & $4.6_{5.9}$ & $2.0_{4.0}$ & $53.0_{9.6}$ & $48.8_{6.4}$ & $9.2_{8.4}$ & $17.2_{20.4}$ & $40.5_{1.3}$ & $26.3_{2.2}$ & $34.1_{24.1}$ & $4.6_{4.9}$ & $51.7_{22.8}$ & $19.5_{11.7}$ & $26.3_{11.5}$ \\
$\overline{\text{CTransPath}}$~\cite{wang2022ctp} & $75.9_{2.7}$ & $3.5_{4.3}$ & $9.0_{13.2}$ & $32.9_{12.7}$ & $9.9_{12.4}$ & $41.0_{10.4}$ & $42.8_{5.0}$ & $0.0_{0.0}$ & $4.7_{9.4}$ & $39.7_{1.1}$ & $25.5_{2.4}$ & $52.5_{14.9}$ & $12.5_{18.6}$ & $\underline{54.4}_{27.3}$ & $0.0_{0.0}$ & $27.0_{11.7}$ \\
$\overline{\text{H-Optimus}}$~\cite{hoptimus0} & $85.4_{1.8}$ & $29.4_{7.0}$ & $56.3_{5.3}$ & $53.7_{15.0}$ & $25.2_{7.5}$ & $78.7_{10.1}$ & $65.7_{14.2}$ & $0.0_{0.0}$ & $27.3_{22.2}$ & $46.4_{7.7}$ & $31.6_{6.3}$ & $12.3_{22.8}$ & $11.5_{10.8}$ & $38.0_{24.3}$ & $17.8_{19.6}$ & $38.6_{13.8}$ \\
$\overline{\text{CONCH}}$~\cite{lu2024conch} & $89.4_{0.8}$ & $11.1_{14.9}$ & $36.2_{16.5}$ & $58.3_{8.6}$ & $36.5_{7.6}$ & $67.5_{7.8}$ & $54.2_{11.5}$ & $18.8_{11.9}$ & $16.5_{10.4}$ & $42.6_{3.1}$ & $21.5_{11.3}$ & $54.9_{8.2}$ & $15.6_{19.8}$ & $34.1_{14.8}$ & $26.2_{13.3}$ & $38.9_{11.7}$ \\
$\overline{\text{Ensemble Prediction}}$ & $81.2_{2.4}$ & $18.6_{11.2}$ & $56.9_{7.0}$ & $59.5_{12.0}$ & $19.0_{7.2}$ & $83.7_{6.6}$ & $66.0_{14.3}$ & $1.5_{3.0}$ & $24.1_{16.6}$ & $52.1_{11.1}$ & $36.1_{8.0}$ & $20.2_{23.3}$ & $15.8_{17.6}$ & $37.0_{22.5}$ & $14.4_{18.4}$ & $39.1_{13.6}$ \\
$\overline{\text{UNI}}$~\cite{chen2024uni} & $74.6_{3.0}$ & $7.4_{9.4}$ & $49.2_{10.6}$ & $63.7_{11.8}$ & $22.9_{16.4}$ & $82.9_{5.1}$ & $67.2_{10.8}$ & $12.0_{7.8}$ & $27.6_{11.4}$ & $49.8_{9.0}$ & $30.8_{2.5}$ & $27.0_{20.4}$ & $26.6_{19.6}$ & $45.6_{15.0}$ & $22.5_{14.2}$ & $40.7_{12.3}$ \\
$\overline{\text{Concatenated}}$ & $78.0_{3.8}$ & $20.7_{7.9}$ & $43.4_{15.7}$ & $62.0_{12.4}$ & $27.7_{3.1}$ & $\mathbf{86.8}_{1.8}$ & $72.3_{12.5}$ & $4.0_{4.9}$ & $16.8_{18.1}$ & $\mathbf{74.3}_{10.4}$ & $36.7_{6.6}$ & $19.2_{15.0}$ & $21.4_{15.0}$ & $45.4_{20.6}$ & $15.2_{18.7}$ & $41.6_{12.6}$ \\
$\overline{\text{Virchow2}}$~\cite{zimmermann2024virchow2} & $80.8_{1.9}$ & $25.7_{14.0}$ & $56.5_{6.3}$ & $59.4_{10.7}$ & $24.1_{7.5}$ & $85.2_{6.1}$ & $67.8_{11.4}$ & $9.6_{11.7}$ & $26.1_{15.8}$ & $60.4_{13.8}$ & $45.1_{8.0}$ & $37.6_{30.9}$ & $32.2_{20.7}$ & $27.2_{18.4}$ & $15.7_{14.6}$ & $43.6_{14.5}$ \\
$\overline{\text{GigaPath}}$~\cite{xu2024gigapath} & $84.3_{2.7}$ & $16.7_{9.2}$ & $56.1_{8.6}$ & $45.3_{9.1}$ & $23.1_{10.2}$ & $83.7_{2.4}$ & $\underline{75.0}_{6.1}$ & $19.7_{9.1}$ & $34.8_{17.4}$ & $67.1_{8.5}$ & $38.3_{6.8}$ & $23.4_{19.8}$ & $28.4_{16.2}$ & $44.9_{18.7}$ & $\mathbf{30.8}_{16.2}$ & $44.8_{12.0}$ \\
\hline
MADELEINE~\cite{jaume2024madeleine} & $84.9_{0.6}$ & $11.1_{14.5}$ & $1.1_{2.2}$ & $62.3_{4.0}$ & $15.2_{12.4}$ & $53.4_{8.1}$ & $45.5_{8.3}$ & $22.7_{11.7}$ & $5.4_{8.6}$ & $22.0_{19.9}$ & $17.7_{15.0}$ & $23.5_{24.2}$ & $5.0_{6.1}$ & $26.7_{15.6}$ & $26.0_{14.0}$ & $28.2_{12.7}$ \\
CHIEF~\cite{Wang2024Chief} & $82.3_{1.0}$ & $12.5_{11.3}$ & $36.3_{18.5}$ & $51.4_{7.8}$ & $22.8_{4.9}$ & $69.3_{7.0}$ & $62.7_{4.5}$ & $3.6_{7.3}$ & $18.7_{19.4}$ & $46.1_{7.8}$ & $25.4_{2.2}$ & $62.1_{4.6}$ & $18.1_{20.6}$ & $41.3_{33.7}$ & $8.5_{12.8}$ & $37.4_{13.8}$ \\
GigaPath-SE~\cite{xu2024gigapath} & $70.6_{2.4}$ & $19.3_{2.7}$ & $41.0_{9.1}$ & $58.7_{7.4}$ & $\mathbf{40.4}_{5.4}$ & $66.6_{11.9}$ & $65.0_{17.3}$ & $9.4_{5.6}$ & $\mathbf{42.3}_{6.9}$ & $40.8_{2.8}$ & $24.6_{3.2}$ & $19.8_{20.9}$ & $\underline{35.7}_{9.2}$ & $54.3_{15.6}$ & $18.0_{13.4}$ & $40.4_{10.5}$ \\
\textsc{{Cobra}}$^\dag$-CTP & $85.4_{0.7}$ & $18.8_{12.1}$ & $51.9_{8.3}$ & $55.5_{3.3}$ & $15.4_{12.8}$ & $77.4_{2.5}$ & $74.1_{1.6}$ & $12.4_{7.0}$ & $37.8_{10.7}$ & $45.1_{18.6}$ & $31.4_{3.4}$ & $56.4_{9.8}$ & $30.8_{15.8}$ & $52.8_{26.6}$ & $1.9_{3.8}$ & $43.1_{11.5}$ \\
\textsc{{Cobra}}-CTP & $85.5_{0.9}$ & $19.2_{10.5}$ & $38.5_{20.1}$ & $58.7_{3.7}$ & $24.9_{13.4}$ & $77.6_{1.2}$ & $70.8_{2.8}$ & $9.8_{5.1}$ & $37.4_{18.9}$ & $49.5_{3.1}$ & $32.3_{6.8}$ & $\underline{62.9}_{3.9}$ & $25.2_{21.2}$ & $\mathbf{66.5}_{3.8}$ & $7.4_{11.3}$ & $44.4_{10.8}$ \\
\textsc{{Cobra}}$^\dag$-H0 & $94.2_{0.6}$ & $\underline{55.3}_{4.3}$ & $\mathbf{65.9}_{2.6}$ & $64.8_{5.0}$ & $\underline{37.7}_{7.4}$ & $\underline{86.4}_{2.2}$ & $65.4_{8.9}$ & $0.0_{0.0}$ & $14.1_{10.8}$ & $67.1_{15.0}$ & $37.9_{7.0}$ & $19.0_{18.4}$ & $21.8_{16.2}$ & $34.4_{19.2}$ & $7.9_{6.8}$ & $44.8_{10.3}$ \\
\textsc{{Cobra}}-ENC & $84.9_{0.7}$ & $33.7_{5.7}$ & $47.3_{6.8}$ & $62.7_{1.9}$ & $35.8_{5.0}$ & $79.0_{1.4}$ & $71.1_{2.3}$ & $17.9_{3.3}$ & $28.7_{11.0}$ & $36.7_{1.6}$ & $28.9_{5.6}$ & $45.6_{5.8}$ & $\mathbf{37.4}_{4.1}$ & $49.6_{6.9}$ & $\underline{29.5}_{6.1}$ & $45.9_{5.3}$ \\
\textsc{{Cobra}}-H0 & $\underline{94.6}_{0.3}$ & $52.2_{4.8}$ & $65.1_{1.7}$ & $68.2_{5.0}$ & $35.4_{4.3}$ & $81.5_{6.5}$ & $70.1_{6.8}$ & $6.1_{8.7}$ & $16.5_{15.9}$ & $72.5_{10.9}$ & $40.6_{7.2}$ & $20.7_{17.9}$ & $20.4_{14.3}$ & $32.0_{19.0}$ & $17.6_{18.4}$ & $46.2_{11.2}$ \\
\textsc{{Cobra}}$^\dag$-UNI & $90.5_{1.2}$ & $34.9_{13.6}$ & $56.8_{7.8}$ & $\underline{73.1}_{5.8}$ & $30.9_{16.5}$ & $82.3_{3.6}$ & $71.9_{6.8}$ & $21.2_{4.6}$ & $22.8_{21.1}$ & $58.9_{6.4}$ & $\underline{46.4}_{5.0}$ & $35.1_{10.9}$ & $30.7_{20.6}$ & $33.3_{15.0}$ & $15.9_{16.0}$ & $47.0_{12.0}$ \\
\textsc{{Cobra}}-UNI & $90.0_{3.0}$ & $32.5_{17.3}$ & $58.4_{7.6}$ & $71.8_{2.1}$ & $31.1_{12.4}$ & $83.9_{2.6}$ & $72.8_{6.2}$ & $\underline{25.4}_{3.9}$ & $22.5_{19.0}$ & $62.7_{4.4}$ & $41.0_{7.0}$ & $37.6_{10.6}$ & $33.9_{10.3}$ & $37.6_{15.2}$ & $16.5_{20.8}$ & $47.8_{11.2}$ \\
\textsc{{Cobra}}-V2 & $89.9_{1.1}$ & $46.9_{8.0}$ & $63.3_{5.7}$ & $69.1_{5.0}$ & $29.3_{15.3}$ & $83.7_{2.5}$ & $67.8_{18.5}$ & $23.7_{2.8}$ & $35.3_{22.4}$ & $\underline{73.0}_{6.2}$ & $\mathbf{49.8}_{4.2}$ & $50.3_{10.3}$ & $24.7_{21.6}$ & $22.3_{14.7}$ & $0.0_{0.0}$ & $48.6_{11.7}$ \\
PRISM~\cite{shaikovski2024prism} & $\mathbf{96.2}_{0.3}$ & $\mathbf{56.1}_{6.3}$ & $52.7_{1.3}$ & $\mathbf{75.0}_{1.8}$ & $27.5_{12.5}$ & $74.6_{2.7}$ & $62.9_{13.8}$ & $24.3_{4.4}$ & $23.5_{11.0}$ & $54.1_{2.9}$ & $29.2_{3.2}$ & $\mathbf{66.6}_{2.9}$ & $20.0_{20.5}$ & $46.0_{5.7}$ & $27.9_{4.7}$ & $49.1_{8.3}$ \\
\textsc{{Cobra}}$^\dag$-V2 & $89.7_{0.6}$ & $47.9_{4.6}$ & $63.1_{5.3}$ & $69.4_{3.4}$ & $28.9_{14.7}$ & $84.3_{2.0}$ & $69.3_{10.0}$ & $24.0_{2.1}$ & $30.9_{21.3}$ & $72.3_{7.4}$ & $46.3_{6.8}$ & $49.7_{18.2}$ & $34.8_{18.6}$ & $24.4_{17.5}$ & $9.1_{11.6}$ & $\underline{49.6}_{11.7}$ \\
\textsc{{Cobra}}$^\dag$-GP & $92.2_{0.5}$ & $34.2_{6.2}$ & $\underline{65.2}_{5.1}$ & $67.7_{5.5}$ & $34.1_{9.2}$ & $84.2_{1.3}$ & $\mathbf{77.7}_{1.9}$ & $\mathbf{28.1}_{4.0}$ & $\underline{42.0}_{10.5}$ & $68.7_{7.8}$ & $46.2_{3.3}$ & $35.7_{16.0}$ & $27.1_{23.1}$ & $43.8_{13.9}$ & $6.5_{12.9}$ & $\mathbf{50.2}_{10.1}$ \\
\bottomrule
\end{tabular}
}
\end{table*}

\begin{table*}[ht!]
\centering
\caption{\textbf{Classification performance comparison.} Balanced accuracy score of models trained on TCGA deployed on CPTAC datasets. $\overline{\text{Overline}}$ indicates mean over patch embeddings, $^\dag$ indicates that embeddings of all four training FMs were used to generate the weighting vector (\cref{eq:cobra-dagger}). For the other \ac{cobra} entries, we used the inference mode from (\cref{eq:aggregation_module_inference}). \textbf{Bold} indicates the best performance, and $\underline{\text{underline}}$ indicates the second-best performance. The abbreviations are as follows: ST: Subtyping, CTP: CTransPath~\cite{wang2022ctp}, H0: H-Optimus-0~\cite{hoptimus0}, V2: Virchow-2~\cite{zimmermann2024virchow2}, GP: GigaPath~\cite{xu2024gigapath}, SE: Slide Encoder.}
\label{tab:mlp_Balanced Acc-20x}
\resizebox{\textwidth}{!}{%
\begin{tabular}{l|l|llll|llll|llllll|l}
\toprule
Balanced Acc-20$\times$[\%] & NSCLC & \multicolumn{4}{c|}{LUAD} & \multicolumn{4}{c|}{BRCA} & \multicolumn{6}{c|}{COAD} & Average \\
Model & ST & STK11 & EGFR & TP53 & KRAS & ESR1 & PGR & ERBB2 & PIK3CA & MSI & BRAF & LN & KRAS & Side & PIK3CA &  \\
\midrule
$\overline{\text{Virchow}}$~\cite{vorontsov2024virchow} & $80.5_{1.4}$ & $50.6_{1.1}$ & $50.6_{1.1}$ & $50.3_{0.9}$ & $49.8_{0.4}$ & $58.2_{4.0}$ & $53.5_{3.2}$ & $50.4_{3.5}$ & $52.6_{4.3}$ & $56.8_{1.9}$ & $55.5_{1.9}$ & $52.8_{4.7}$ & $48.6_{1.0}$ & $51.2_{0.8}$ & $49.5_{4.2}$ & $54.1_{2.7}$ \\
$\overline{\text{CTransPath}}$~\cite{wang2022ctp} & $77.3_{2.0}$ & $50.2_{0.2}$ & $52.1_{3.2}$ & $57.8_{3.2}$ & $48.1_{3.1}$ & $57.6_{3.5}$ & $59.1_{1.1}$ & $48.4_{1.7}$ & $50.7_{1.5}$ & $55.6_{2.1}$ & $53.3_{3.6}$ & $54.7_{2.9}$ & $52.6_{4.2}$ & $50.4_{0.9}$ & $50.0_{0.0}$ & $54.5_{2.5}$ \\
$\overline{\text{CONCH}}$~\cite{lu2024conch} & $89.7_{0.7}$ & $52.2_{5.6}$ & $55.6_{5.4}$ & $67.4_{4.0}$ & $54.8_{6.5}$ & $73.5_{2.9}$ & $65.6_{3.7}$ & $56.8_{5.3}$ & $53.4_{2.2}$ & $60.4_{3.8}$ & $54.1_{5.0}$ & $\underline{60.6}_{4.8}$ & $53.2_{4.3}$ & $51.5_{1.6}$ & $54.9_{4.4}$ & $60.2_{4.3}$ \\
$\overline{\text{H-Optimus}}$~\cite{hoptimus0} & $87.0_{1.3}$ & $58.1_{3.0}$ & $68.8_{3.2}$ & $62.5_{4.8}$ & $52.5_{1.8}$ & $74.4_{3.1}$ & $63.4_{3.0}$ & $50.0_{0.0}$ & $54.9_{5.9}$ & $64.0_{9.8}$ & $59.7_{5.0}$ & $50.8_{1.6}$ & $52.4_{3.1}$ & $53.3_{3.5}$ & $54.9_{8.8}$ & $60.4_{4.6}$ \\
$\overline{\text{UNI}}$~\cite{chen2024uni} & $79.6_{1.9}$ & $51.0_{2.2}$ & $62.7_{8.4}$ & $65.2_{5.0}$ & $50.4_{4.6}$ & $75.4_{2.4}$ & $66.1_{4.0}$ & $51.6_{2.5}$ & $54.3_{2.2}$ & $67.8_{9.9}$ & $61.6_{3.3}$ & $54.8_{5.8}$ & $55.8_{4.5}$ & $54.4_{3.2}$ & $55.7_{5.7}$ & $60.4_{4.9}$ \\
$\overline{\text{Ensemble Prediction}}$ & $84.1_{1.7}$ & $54.3_{4.1}$ & $69.5_{4.4}$ & $64.8_{4.1}$ & $52.8_{1.2}$ & $78.3_{3.3}$ & $67.1_{4.3}$ & $49.6_{0.8}$ & $53.8_{4.4}$ & $70.0_{11.4}$ & $65.9_{4.8}$ & $52.3_{3.2}$ & $53.9_{4.4}$ & $54.6_{3.6}$ & $53.8_{6.2}$ & $61.7_{4.8}$ \\
$\overline{\text{GigaPath}}$~\cite{xu2024gigapath} & $86.1_{2.2}$ & $52.6_{3.9}$ & $69.2_{5.5}$ & $60.7_{3.7}$ & $52.5_{2.2}$ & $74.0_{2.4}$ & $67.2_{2.0}$ & $56.9_{6.4}$ & $54.3_{3.1}$ & $82.9_{4.6}$ & $68.2_{5.5}$ & $54.2_{5.2}$ & $54.1_{3.7}$ & $56.0_{2.7}$ & $\mathbf{58.3}_{6.8}$ & $63.1_{4.3}$ \\
$\overline{\text{Concatenated}}$ & $81.9_{2.6}$ & $54.4_{3.0}$ & $64.1_{6.8}$ & $64.9_{3.9}$ & $54.4_{1.8}$ & $78.1_{3.1}$ & $69.5_{4.5}$ & $50.3_{0.9}$ & $53.2_{4.9}$ & $\mathbf{85.5}_{6.0}$ & $68.5_{6.1}$ & $53.3_{4.1}$ & $55.2_{4.3}$ & $\mathbf{58.7}_{5.5}$ & $55.2_{6.7}$ & $63.1_{4.6}$ \\
$\overline{\text{Virchow2}}$~\cite{zimmermann2024virchow2} & $83.6_{1.3}$ & $57.3_{5.5}$ & $68.5_{4.1}$ & $65.0_{3.5}$ & $54.5_{2.3}$ & $\mathbf{78.8}_{4.0}$ & $69.8_{4.9}$ & $52.6_{3.6}$ & $53.8_{5.0}$ & $77.4_{11.5}$ & $73.3_{4.7}$ & $54.4_{3.9}$ & $53.7_{4.3}$ & $53.7_{2.8}$ & $52.6_{5.4}$ & $63.3_{5.0}$ \\
\hline
GigaPath-SE~\cite{xu2024gigapath} & $76.5_{1.6}$ & $53.7_{1.0}$ & $61.1_{2.9}$ & $63.6_{2.3}$ & $\mathbf{56.1}_{2.8}$ & $65.2_{4.0}$ & $65.2_{7.4}$ & $48.5_{5.3}$ & $55.0_{3.0}$ & $60.2_{2.5}$ & $54.1_{2.2}$ & $53.7_{6.2}$ & $51.9_{2.1}$ & $47.6_{1.5}$ & $52.2_{4.5}$ & $57.6_{3.7}$ \\
MADELEINE~\cite{jaume2024madeleine} & $85.1_{0.8}$ & $53.3_{4.4}$ & $50.2_{0.3}$ & $65.0_{3.2}$ & $50.7_{1.5}$ & $66.7_{3.1}$ & $62.3_{2.1}$ & $57.4_{3.8}$ & $51.2_{2.1}$ & $56.9_{6.5}$ & $56.2_{5.8}$ & $53.5_{3.1}$ & $50.7_{1.1}$ & $50.4_{1.1}$ & $\underline{57.7}_{4.7}$ & $57.8_{3.4}$ \\
\textsc{{Cobra}}-ENC & $85.9_{0.7}$ & $60.2_{5.0}$ & $62.5_{3.5}$ & $66.1_{1.4}$ & $53.3_{2.1}$ & $61.7_{2.8}$ & $60.8_{4.7}$ & $53.5_{2.1}$ & $53.8_{4.4}$ & $51.9_{1.7}$ & $58.7_{7.9}$ & $48.8_{2.6}$ & $47.7_{2.9}$ & $46.0_{3.2}$ & $55.6_{2.8}$ & $57.8_{3.6}$ \\
CHIEF~\cite{Wang2024Chief} & $84.2_{0.8}$ & $52.9_{3.2}$ & $59.8_{5.5}$ & $62.5_{3.4}$ & $47.7_{3.7}$ & $72.4_{3.0}$ & $67.8_{1.7}$ & $50.5_{1.3}$ & $54.6_{4.9}$ & $63.2_{8.4}$ & $52.8_{4.2}$ & $52.5_{2.9}$ & $53.1_{3.5}$ & $50.0_{0.1}$ & $51.7_{2.4}$ & $58.4_{3.8}$ \\
\textsc{{Cobra}}-CTP & $86.7_{0.7}$ & $54.5_{2.7}$ & $60.8_{6.6}$ & $66.0_{2.4}$ & $50.7_{3.3}$ & $68.5_{1.9}$ & $62.1_{2.1}$ & $51.7_{1.2}$ & $56.8_{4.9}$ & $68.6_{3.3}$ & $62.1_{6.4}$ & $55.9_{3.5}$ & $51.2_{1.6}$ & $49.5_{2.7}$ & $51.1_{2.3}$ & $59.7_{3.5}$ \\
\textsc{{Cobra}}$^\dag$-CTP & $86.9_{0.6}$ & $54.7_{3.5}$ & $65.4_{5.6}$ & $64.6_{2.2}$ & $49.4_{1.8}$ & $66.0_{0.7}$ & $64.2_{1.3}$ & $52.9_{2.0}$ & $55.0_{2.4}$ & $68.3_{8.1}$ & $62.0_{4.4}$ & $54.5_{3.5}$ & $53.9_{1.6}$ & $49.1_{2.3}$ & $50.4_{0.8}$ & $59.8_{3.4}$ \\
PRISM~\cite{shaikovski2024prism} & $\mathbf{96.3}_{0.3}$ & $\mathbf{75.5}_{5.1}$ & $65.6_{0.7}$ & $\mathbf{74.1}_{2.0}$ & $52.0_{2.9}$ & $\underline{78.4}_{1.6}$ & $69.5_{5.0}$ & $57.8_{5.6}$ & $54.4_{2.5}$ & $71.4_{2.7}$ & $58.9_{1.9}$ & $\mathbf{61.9}_{3.3}$ & $52.0_{4.8}$ & $50.5_{2.5}$ & $53.7_{4.2}$ & $64.8_{3.4}$ \\
\textsc{{Cobra}}$^\dag$-H0 & $94.4_{0.6}$ & $\underline{74.6}_{3.0}$ & $\mathbf{75.0}_{2.2}$ & $69.4_{2.2}$ & $\underline{55.3}_{4.1}$ & $78.3_{3.9}$ & $64.4_{3.2}$ & $48.4_{1.1}$ & $50.1_{1.7}$ & $81.5_{10.0}$ & $67.7_{4.6}$ & $53.8_{4.0}$ & $55.2_{5.5}$ & $53.8_{3.8}$ & $51.1_{1.4}$ & $64.9_{4.1}$ \\
\textsc{{Cobra}}-H0 & $\underline{94.8}_{0.3}$ & $72.3_{3.0}$ & $74.4_{1.3}$ & $71.3_{3.1}$ & $52.8_{1.8}$ & $74.7_{6.1}$ & $66.1_{4.6}$ & $50.0_{3.1}$ & $50.2_{2.2}$ & $84.8_{5.6}$ & $69.9_{4.2}$ & $52.2_{3.6}$ & $54.9_{4.0}$ & $53.1_{3.0}$ & $55.3_{6.1}$ & $65.1_{3.8}$ \\
\textsc{{Cobra}}$^\dag$-UNI & $91.2_{0.9}$ & $61.3_{8.1}$ & $68.7_{5.2}$ & $\underline{73.5}_{2.1}$ & $51.8_{4.4}$ & $74.4_{1.5}$ & $70.2_{3.7}$ & $56.4_{3.5}$ & $53.9_{5.3}$ & $76.2_{4.6}$ & $72.9_{3.0}$ & $58.1_{2.6}$ & $\mathbf{57.9}_{6.0}$ & $55.2_{2.8}$ & $54.6_{4.8}$ & $65.1_{4.3}$ \\
\textsc{{Cobra}}-UNI & $90.9_{2.4}$ & $61.5_{8.9}$ & $69.9_{4.7}$ & $70.7_{1.1}$ & $51.6_{4.8}$ & $77.1_{2.1}$ & $\underline{71.3}_{3.2}$ & $\underline{58.9}_{3.7}$ & $53.8_{5.2}$ & $79.1_{3.5}$ & $69.8_{4.8}$ & $58.0_{2.3}$ & $53.1_{3.8}$ & $56.8_{4.0}$ & $55.8_{8.4}$ & $65.2_{4.7}$ \\
\textsc{{Cobra}}-V2 & $90.6_{0.9}$ & $68.2_{5.0}$ & $73.3_{3.9}$ & $70.2_{3.3}$ & $54.4_{3.0}$ & $67.3_{9.0}$ & $70.3_{6.1}$ & $56.6_{1.2}$ & $\mathbf{59.1}_{6.6}$ & $\underline{84.9}_{4.5}$ & $\underline{75.0}_{6.5}$ & $56.9_{2.6}$ & $\underline{56.9}_{6.5}$ & $53.6_{2.6}$ & $49.4_{0.6}$ & $\underline{65.8}_{4.8}$ \\
\textsc{{Cobra}}$^\dag$-GP & $92.6_{0.5}$ & $60.3_{3.5}$ & $\underline{74.5}_{3.5}$ & $71.0_{3.8}$ & $52.9_{6.3}$ & $73.0_{5.1}$ & $67.5_{2.4}$ & $\mathbf{60.1}_{2.8}$ & $\underline{57.7}_{3.1}$ & $81.3_{4.1}$ & $\mathbf{75.5}_{3.2}$ & $57.6_{3.9}$ & $55.3_{5.6}$ & $\underline{56.9}_{1.8}$ & $50.3_{2.8}$ & $\underline{65.8}_{3.8}$ \\
\textsc{{Cobra}}$^\dag$-V2 & $90.5_{0.5}$ & $68.7_{3.0}$ & $73.2_{3.6}$ & $71.0_{3.0}$ & $53.1_{2.1}$ & $68.3_{6.3}$ & $\mathbf{72.0}_{3.8}$ & $57.2_{1.8}$ & $57.5_{6.8}$ & $\underline{84.9}_{4.3}$ & $74.2_{8.7}$ & $56.6_{4.0}$ & $\underline{56.9}_{5.4}$ & $55.1_{3.5}$ & $51.6_{2.3}$ & $\mathbf{66.1}_{4.4}$ \\
\bottomrule
\end{tabular}
}
\end{table*}

\begin{table*}[ht!]
\centering
\caption{\textbf{Classification performance comparison.} AUC score of models trained on TCGA deployed on CPTAC datasets. $\overline{\text{Overline}}$ indicates mean over patch embeddings, $^\dag$ indicates that embeddings of all four training FMs were used to generate the weighting vector (\cref{eq:cobra-dagger}). For the other \ac{cobra} entries, we used the inference mode from (\cref{eq:aggregation_module_inference}). \textbf{Bold} indicates the best performance, and $\underline{\text{underline}}$ indicates the second-best performance. The abbreviations are as follows: ST: Subtyping, CTP: CTransPath~\cite{wang2022ctp}, H0: H-Optimus-0~\cite{hoptimus0}, V2: Virchow-2~\cite{zimmermann2024virchow2}, GP: GigaPath~\cite{xu2024gigapath}, SE: Slide Encoder.}
\label{tab:mlp_AUC-5x}
\resizebox{\textwidth}{!}{%
\begin{tabular}{l|l|llll|llll|llllll|l}
\toprule
AUC-5$\times$[\%] & NSCLC & \multicolumn{4}{c|}{LUAD} & \multicolumn{4}{c|}{BRCA} & \multicolumn{6}{c|}{COAD} & Average \\
Model & ST & STK11 & EGFR & TP53 & KRAS & ESR1 & PGR & ERBB2 & PIK3CA & MSI & BRAF & LN & KRAS & Side & PIK3CA &  \\
\midrule
$\overline{\text{Virchow}}$~\cite{vorontsov2024virchow} & $84.0_{1.9}$ & $69.0_{2.9}$ & $65.7_{2.9}$ & $61.2_{2.8}$ & $52.6_{10.7}$ & $76.1_{2.8}$ & $74.5_{1.1}$ & $54.2_{2.2}$ & $64.9_{8.2}$ & $81.7_{0.3}$ & $63.9_{2.2}$ & $46.9_{2.4}$ & $61.6_{7.3}$ & $47.8_{1.6}$ & $52.5_{6.8}$ & $63.8_{4.7}$ \\
$\overline{\text{CTransPath}}$~\cite{wang2022ctp} & $91.5_{0.8}$ & $67.9_{5.4}$ & $62.6_{6.1}$ & $68.3_{2.5}$ & $50.0_{5.2}$ & $79.1_{2.1}$ & $75.5_{2.9}$ & $52.4_{1.1}$ & $57.2_{6.6}$ & $79.1_{2.0}$ & $62.0_{1.5}$ & $59.4_{3.4}$ & $55.8_{6.3}$ & $50.5_{1.5}$ & $51.6_{4.8}$ & $64.2_{4.0}$ \\
$\overline{\text{UNI}}$~\cite{chen2024uni} & $92.2_{1.1}$ & $63.5_{7.8}$ & $71.7_{4.5}$ & $68.0_{1.7}$ & $53.0_{3.7}$ & $79.7_{1.3}$ & $73.9_{0.6}$ & $50.7_{2.9}$ & $63.0_{6.6}$ & $82.0_{2.3}$ & $68.2_{2.6}$ & $55.4_{2.7}$ & $53.0_{7.0}$ & $52.4_{3.2}$ & $50.5_{8.2}$ & $65.1_{4.5}$ \\
$\overline{\text{GigaPath}}$~\cite{xu2024gigapath} & $96.2_{0.8}$ & $62.9_{14.5}$ & $71.3_{4.6}$ & $70.5_{2.2}$ & $\underline{57.8}_{5.7}$ & $79.8_{1.4}$ & $76.0_{1.8}$ & $54.1_{3.6}$ & $56.7_{5.2}$ & $81.9_{2.7}$ & $58.4_{8.9}$ & $57.0_{3.2}$ & $54.9_{5.7}$ & $52.0_{3.1}$ & $53.6_{1.8}$ & $65.5_{5.5}$ \\
$\overline{\text{H-Optimus}}$~\cite{hoptimus0} & $92.9_{0.6}$ & $72.1_{4.3}$ & $70.7_{2.5}$ & $65.0_{1.8}$ & $53.6_{4.9}$ & $78.8_{1.5}$ & $73.1_{1.1}$ & $52.2_{1.2}$ & $60.9_{6.1}$ & $84.9_{1.9}$ & $65.4_{2.6}$ & $61.1_{2.8}$ & $54.6_{9.6}$ & $53.2_{3.3}$ & $52.3_{7.5}$ & $66.1_{4.3}$ \\
$\overline{\text{CONCH}}$~\cite{lu2024conch} & $97.8_{0.1}$ & $74.4_{2.7}$ & $67.9_{5.7}$ & $74.2_{0.5}$ & $57.5_{8.0}$ & $81.6_{0.7}$ & $79.0_{1.5}$ & $62.5_{2.2}$ & $55.4_{10.7}$ & $80.4_{2.2}$ & $63.9_{9.2}$ & $60.5_{1.1}$ & $64.3_{7.8}$ & $56.1_{3.4}$ & $58.8_{1.7}$ & $69.0_{5.1}$ \\
$\overline{\text{Virchow2}}$~\cite{zimmermann2024virchow2} & $98.3_{0.3}$ & $70.8_{17.1}$ & $\underline{74.8}_{4.8}$ & $74.9_{1.5}$ & $57.3_{6.6}$ & $\mathbf{90.9}_{0.8}$ & $\underline{80.1}_{1.3}$ & $\mathbf{70.2}_{1.7}$ & $66.0_{5.5}$ & $\mathbf{94.7}_{0.5}$ & $81.5_{2.5}$ & $\underline{61.3}_{2.6}$ & $65.2_{8.5}$ & $58.9_{2.7}$ & $62.7_{12.2}$ & $\underline{73.8}_{6.5}$ \\
\hline
GigaPath-SE~\cite{xu2024gigapath} & $90.2_{1.2}$ & $63.7_{4.4}$ & $66.7_{7.2}$ & $74.0_{2.0}$ & $49.9_{6.2}$ & $75.3_{5.4}$ & $73.8_{2.9}$ & $51.8_{6.9}$ & $65.7_{5.0}$ & $75.5_{7.1}$ & $61.1_{15.2}$ & $51.9_{3.5}$ & $57.5_{7.4}$ & $47.9_{3.5}$ & $55.3_{4.7}$ & $64.0_{6.4}$ \\
PRISM~\cite{shaikovski2024prism} & $91.6_{1.2}$ & $64.0_{4.9}$ & $62.5_{5.7}$ & $71.9_{0.9}$ & $53.2_{5.8}$ & $75.1_{1.5}$ & $71.4_{2.7}$ & $61.2_{2.5}$ & $64.3_{6.5}$ & $79.1_{1.0}$ & $66.3_{1.1}$ & $56.4_{4.6}$ & $62.8_{4.1}$ & $49.2_{3.2}$ & $\underline{64.0}_{4.4}$ & $66.2_{3.8}$ \\
MADELEINE~\cite{jaume2024madeleine} & $95.4_{0.4}$ & $71.7_{2.0}$ & $70.2_{2.3}$ & $72.1_{1.8}$ & $\mathbf{59.2}_{5.6}$ & $79.0_{0.7}$ & $77.7_{1.0}$ & $62.6_{3.8}$ & $59.8_{5.1}$ & $75.6_{4.4}$ & $60.3_{2.8}$ & $57.4_{3.2}$ & $\underline{66.6}_{8.4}$ & $51.2_{0.4}$ & $53.8_{6.3}$ & $67.5_{3.9}$ \\
\textsc{{Cobra}}-UNI & $97.1_{0.4}$ & $66.3_{15.4}$ & $71.9_{3.8}$ & $74.6_{1.0}$ & $54.7_{3.6}$ & $84.2_{0.9}$ & $73.6_{1.1}$ & $59.6_{5.0}$ & $65.7_{3.3}$ & $79.2_{1.9}$ & $68.7_{1.1}$ & $56.1_{3.8}$ & $51.2_{5.9}$ & $53.5_{3.7}$ & $59.7_{4.2}$ & $67.7_{5.1}$ \\
\textsc{{Cobra}}-H0 & $97.2_{0.3}$ & $78.2_{5.9}$ & $71.9_{3.6}$ & $72.4_{1.7}$ & $51.6_{3.6}$ & $82.5_{1.6}$ & $75.1_{1.1}$ & $56.4_{3.7}$ & $59.8_{3.0}$ & $83.7_{2.1}$ & $71.0_{1.6}$ & $58.3_{4.5}$ & $51.9_{5.8}$ & $50.6_{4.2}$ & $54.5_{8.7}$ & $67.7_{4.0}$ \\
\textsc{{Cobra}}$^\dag$-CTP & $96.6_{0.4}$ & $70.5_{3.9}$ & $70.5_{2.1}$ & $74.3_{1.0}$ & $53.2_{2.4}$ & $82.2_{0.9}$ & $77.1_{0.8}$ & $65.7_{2.9}$ & $66.0_{2.6}$ & $79.3_{1.3}$ & $67.9_{2.8}$ & $60.6_{3.2}$ & $53.0_{8.8}$ & $47.2_{2.4}$ & $51.6_{2.3}$ & $67.7_{3.2}$ \\
\textsc{{Cobra}}-CTP & $96.5_{0.3}$ & $71.9_{2.3}$ & $70.0_{1.7}$ & $74.7_{1.2}$ & $51.9_{3.1}$ & $82.8_{0.8}$ & $77.7_{0.8}$ & $61.1_{7.3}$ & $65.2_{2.2}$ & $79.9_{1.6}$ & $69.1_{2.4}$ & $58.5_{4.0}$ & $58.2_{6.2}$ & $48.8_{2.7}$ & $51.9_{3.7}$ & $67.9_{3.3}$ \\
\textsc{{Cobra}}$^\dag$-H0 & $97.6_{0.4}$ & $78.5_{4.0}$ & $71.3_{4.0}$ & $73.1_{1.3}$ & $51.8_{5.2}$ & $81.8_{1.1}$ & $74.9_{1.8}$ & $56.7_{2.0}$ & $63.1_{4.1}$ & $82.3_{1.6}$ & $71.4_{1.8}$ & $59.6_{3.1}$ & $56.0_{6.6}$ & $50.6_{2.4}$ & $55.7_{5.3}$ & $68.3_{3.5}$ \\
\textsc{{Cobra}}$^\dag$-UNI & $97.1_{0.4}$ & $77.1_{1.6}$ & $71.5_{2.5}$ & $75.2_{1.1}$ & $57.3_{2.0}$ & $82.7_{0.7}$ & $74.3_{1.5}$ & $57.1_{3.2}$ & $\underline{68.2}_{4.2}$ & $78.8_{1.9}$ & $70.5_{2.2}$ & $55.9_{4.3}$ & $54.6_{7.0}$ & $49.3_{6.8}$ & $58.4_{4.4}$ & $68.5_{3.5}$ \\
CHIEF~\cite{Wang2024Chief} & $95.8_{0.3}$ & $77.3_{2.4}$ & $68.1_{2.9}$ & $72.7_{1.3}$ & $51.9_{7.6}$ & $84.3_{0.5}$ & $\mathbf{81.0}_{0.4}$ & $68.7_{3.0}$ & $\mathbf{70.4}_{1.9}$ & $78.0_{0.7}$ & $67.5_{2.5}$ & $59.0_{8.0}$ & $58.2_{8.7}$ & $49.6_{1.6}$ & $52.8_{2.3}$ & $69.0_{4.0}$ \\
\textsc{{Cobra}}-V2 & $\underline{98.9}_{0.3}$ & $\mathbf{82.6}_{0.9}$ & $74.6_{2.6}$ & $\mathbf{80.1}_{2.7}$ & $55.9_{3.3}$ & $88.8_{1.0}$ & $78.2_{1.3}$ & $\underline{69.5}_{2.7}$ & $66.3_{3.4}$ & $\underline{94.3}_{1.2}$ & $\mathbf{83.5}_{1.5}$ & $\mathbf{61.5}_{1.9}$ & $59.8_{12.3}$ & $\mathbf{60.0}_{1.8}$ & $53.5_{6.8}$ & $\underline{73.8}_{4.1}$ \\
\textsc{{Cobra}}$^\dag$-V2 & $\mathbf{99.0}_{0.2}$ & $\underline{81.6}_{1.5}$ & $\mathbf{75.5}_{2.7}$ & $\underline{79.9}_{1.8}$ & $51.4_{7.8}$ & $\underline{89.0}_{1.3}$ & $79.0_{1.2}$ & $67.2_{2.9}$ & $62.1_{4.7}$ & $94.1_{0.7}$ & $\underline{82.9}_{2.7}$ & $\underline{61.3}_{1.5}$ & $\mathbf{68.1}_{8.5}$ & $\underline{59.4}_{3.1}$ & $\mathbf{69.6}_{3.4}$ & $\mathbf{74.7}_{3.7}$ \\
\bottomrule
\end{tabular}
}
\end{table*}

\begin{table*}[ht!]
\centering
\caption{\textbf{Classification performance comparison.} AUC score of models trained on TCGA deployed on CPTAC datasets. $\overline{\text{Overline}}$ indicates mean over patch embeddings, $^\dag$ indicates that embeddings of all four training FMs were used to generate the weighting vector (\cref{eq:cobra-dagger}). For the other \ac{cobra} entries, we used the inference mode from (\cref{eq:aggregation_module_inference}). \textbf{Bold} indicates the best performance, and $\underline{\text{underline}}$ indicates the second-best performance. The abbreviations are as follows: ST: Subtyping, CTP: CTransPath~\cite{wang2022ctp}, H0: H-Optimus-0~\cite{hoptimus0}, V2: Virchow-2~\cite{zimmermann2024virchow2}, GP: GigaPath~\cite{xu2024gigapath}, SE: Slide Encoder.}
\label{tab:mlp_AUC-9x}
\resizebox{\textwidth}{!}{%
\begin{tabular}{l|l|llll|llll|llllll|l}
\toprule
AUC-9$\times$[\%] & NSCLC & \multicolumn{4}{c|}{LUAD} & \multicolumn{4}{c|}{BRCA} & \multicolumn{6}{c|}{COAD} & Average \\
Model & ST & STK11 & EGFR & TP53 & KRAS & ESR1 & PGR & ERBB2 & PIK3CA & MSI & BRAF & LN & KRAS & Side & PIK3CA &  \\
\midrule
$\overline{\text{CTransPath}}$~\cite{wang2022ctp} & $90.1_{0.9}$ & $64.8_{10.4}$ & $63.4_{7.7}$ & $70.9_{2.3}$ & $56.5_{5.5}$ & $77.6_{2.7}$ & $72.8_{2.9}$ & $53.4_{2.0}$ & $59.4_{4.3}$ & $71.6_{18.8}$ & $63.1_{2.7}$ & $60.0_{1.5}$ & $59.1_{4.1}$ & $52.3_{1.1}$ & $53.8_{8.0}$ & $64.6_{6.8}$ \\
$\overline{\text{Virchow}}$~\cite{vorontsov2024virchow} & $90.9_{2.1}$ & $70.9_{5.2}$ & $71.0_{2.7}$ & $71.7_{1.2}$ & $52.6_{6.1}$ & $76.3_{1.7}$ & $72.5_{1.8}$ & $48.3_{3.0}$ & $60.6_{6.0}$ & $82.9_{1.1}$ & $62.2_{3.2}$ & $56.6_{2.2}$ & $50.3_{18.8}$ & $52.7_{1.9}$ & $58.2_{0.9}$ & $65.2_{5.8}$ \\
$\overline{\text{CONCH}}$~\cite{lu2024conch} & $97.8_{0.3}$ & $66.7_{20.5}$ & $63.5_{10.4}$ & $74.8_{1.4}$ & $\mathbf{60.2}_{7.0}$ & $81.6_{1.7}$ & $78.8_{0.8}$ & $59.0_{9.5}$ & $58.9_{11.3}$ & $81.5_{0.9}$ & $57.2_{12.7}$ & $\underline{\mathbf{64.2}}_{4.5}$ & $63.2_{4.7}$ & $58.3_{1.5}$ & $57.8_{4.1}$ & $68.2_{8.3}$ \\
$\overline{\text{GigaPath}}$~\cite{xu2024gigapath} & $97.5_{0.3}$ & $74.9_{5.9}$ & $75.9_{5.0}$ & $73.1_{2.3}$ & $\underline{57.6}_{4.3}$ & $84.9_{1.3}$ & $77.3_{2.2}$ & $57.1_{2.3}$ & $65.1_{6.6}$ & $84.7_{11.5}$ & $72.8_{8.4}$ & $60.3_{2.4}$ & $61.4_{5.9}$ & $56.1_{3.4}$ & $54.0_{7.9}$ & $70.2_{5.5}$ \\
$\overline{\text{H-Optimus}}$~\cite{hoptimus0} & $97.1_{0.6}$ & $81.2_{4.2}$ & $69.3_{5.8}$ & $73.5_{1.0}$ & $49.9_{8.2}$ & $83.9_{2.0}$ & $76.8_{1.8}$ & $58.5_{2.1}$ & $55.4_{9.0}$ & $90.8_{1.5}$ & $76.5_{6.0}$ & $63.0_{2.8}$ & $62.6_{5.1}$ & $59.4_{2.3}$ & $\underline{63.3}_{2.9}$ & $70.7_{4.5}$ \\
$\overline{\text{UNI}}$~\cite{chen2024uni} & $96.4_{0.6}$ & $68.6_{11.1}$ & $75.7_{5.3}$ & $73.4_{1.5}$ & $56.6_{5.2}$ & $85.4_{1.6}$ & $79.3_{0.6}$ & $62.1_{3.6}$ & $57.5_{16.2}$ & $89.3_{1.3}$ & $74.6_{3.7}$ & $61.8_{2.5}$ & $59.3_{9.7}$ & $59.8_{2.0}$ & $60.2_{7.6}$ & $70.7_{6.5}$ \\
$\overline{\text{Virchow2}}$~\cite{zimmermann2024virchow2} & $97.2_{0.8}$ & $82.6_{2.2}$ & $74.7_{3.2}$ & $73.3_{2.0}$ & $52.9_{5.1}$ & $\mathbf{92.1}_{1.7}$ & $\underline{80.0}_{2.6}$ & $69.6_{2.2}$ & $64.5_{5.1}$ & $93.0_{1.4}$ & $75.8_{5.7}$ & $60.5_{2.0}$ & $\mathbf{64.2}_{4.7}$ & $60.0_{2.0}$ & $\mathbf{63.5}_{5.9}$ & $73.6_{3.5}$ \\
\hline
GigaPath-SE~\cite{xu2024gigapath} & $90.5_{0.8}$ & $60.4_{8.9}$ & $68.0_{11.8}$ & $74.7_{1.5}$ & $49.4_{4.4}$ & $75.1_{2.6}$ & $68.7_{3.6}$ & $58.5_{4.1}$ & $60.9_{2.5}$ & $80.0_{2.4}$ & $58.4_{7.0}$ & $56.5_{3.4}$ & $58.1_{6.4}$ & $46.9_{1.7}$ & $50.8_{6.4}$ & $63.8_{5.4}$ \\
MADELEINE~\cite{jaume2024madeleine} & $95.6_{0.4}$ & $69.8_{5.9}$ & $72.0_{2.3}$ & $73.7_{2.0}$ & $44.6_{7.2}$ & $79.8_{1.7}$ & $76.4_{1.2}$ & $64.7_{2.0}$ & $60.9_{6.1}$ & $73.9_{2.5}$ & $60.0_{2.2}$ & $63.1_{3.1}$ & $49.7_{8.4}$ & $53.1_{0.4}$ & $59.6_{2.5}$ & $66.5_{4.0}$ \\
\textsc{{Cobra}}$^\dag$-CTP & $96.4_{0.3}$ & $75.5_{3.5}$ & $69.4_{6.7}$ & $74.4_{1.4}$ & $52.1_{3.8}$ & $81.6_{0.6}$ & $76.2_{0.2}$ & $65.7_{1.5}$ & $62.0_{2.1}$ & $81.3_{1.0}$ & $72.1_{2.5}$ & $59.9_{2.2}$ & $52.6_{8.1}$ & $49.4_{4.4}$ & $55.4_{8.0}$ & $68.3_{4.0}$ \\
\textsc{{Cobra}}-CTP & $96.4_{0.4}$ & $75.9_{3.4}$ & $71.4_{2.5}$ & $74.8_{1.5}$ & $51.2_{3.3}$ & $83.2_{0.5}$ & $78.0_{0.5}$ & $63.5_{4.9}$ & $63.6_{2.6}$ & $81.9_{1.4}$ & $74.0_{3.6}$ & $58.4_{5.9}$ & $56.2_{4.5}$ & $51.0_{2.2}$ & $52.0_{7.4}$ & $68.8_{3.6}$ \\
CHIEF~\cite{Wang2024Chief} & $95.4_{0.4}$ & $74.5_{3.0}$ & $68.8_{4.4}$ & $73.6_{1.2}$ & $55.1_{5.3}$ & $85.7_{0.9}$ & $\mathbf{81.0}_{0.1}$ & $68.1_{3.0}$ & $66.5_{1.9}$ & $76.2_{8.5}$ & $70.8_{1.9}$ & $62.2_{1.0}$ & $58.3_{10.4}$ & $49.5_{2.0}$ & $50.5_{3.9}$ & $69.1_{4.3}$ \\
PRISM~\cite{shaikovski2024prism} & $97.9_{0.4}$ & $80.0_{3.1}$ & $71.8_{2.0}$ & $74.4_{1.5}$ & $56.8_{5.6}$ & $84.8_{0.5}$ & $77.3_{0.9}$ & $65.4_{1.6}$ & $\mathbf{68.5}_{2.3}$ & $80.7_{2.0}$ & $61.2_{5.4}$ & $58.0_{3.3}$ & $50.5_{4.0}$ & $54.5_{5.6}$ & $57.3_{3.9}$ & $69.3_{3.3}$ \\
\textsc{{Cobra}}-H0 & $\mathbf{99.4}_{0.2}$ & $\mathbf{84.4}_{1.6}$ & $72.8_{3.5}$ & $79.2_{1.7}$ & $51.9_{5.4}$ & $84.5_{1.8}$ & $78.3_{2.2}$ & $63.7_{1.6}$ & $58.1_{3.6}$ & $93.2_{0.8}$ & $81.5_{2.7}$ & $\underline{\mathbf{64.2}}_{1.8}$ & $57.7_{6.4}$ & $56.5_{5.7}$ & $57.8_{7.8}$ & $72.2_{3.8}$ \\
\textsc{{Cobra}}$^\dag$-H0 & $\underline{99.3}_{0.2}$ & $83.4_{2.3}$ & $73.6_{3.3}$ & $78.7_{2.1}$ & $52.4_{5.5}$ & $84.1_{2.0}$ & $77.2_{0.9}$ & $66.7_{1.7}$ & $62.3_{3.6}$ & $91.4_{0.7}$ & $82.5_{3.3}$ & $63.8_{2.7}$ & $56.2_{4.5}$ & $57.3_{2.8}$ & $58.3_{2.5}$ & $72.5_{2.9}$ \\
\textsc{{Cobra}}$^\dag$-UNI & $98.9_{0.3}$ & $71.6_{16.2}$ & $74.8_{3.4}$ & $\mathbf{80.5}_{1.7}$ & $56.3_{4.1}$ & $87.2_{0.7}$ & $79.1_{0.9}$ & $65.5_{2.6}$ & $66.0_{3.8}$ & $89.5_{1.4}$ & $\mathbf{85.2}_{1.9}$ & $59.9_{4.9}$ & $62.0_{8.7}$ & $58.4_{3.9}$ & $56.5_{5.7}$ & $72.8_{5.6}$ \\
\textsc{{Cobra}}-UNI & $98.8_{0.2}$ & $79.3_{1.8}$ & $\underline{76.5}_{4.0}$ & $79.9_{1.7}$ & $56.0_{2.8}$ & $88.2_{0.6}$ & $79.7_{0.6}$ & $64.8_{2.7}$ & $66.1_{3.4}$ & $88.9_{1.0}$ & $\underline{84.7}_{2.3}$ & $61.2_{2.9}$ & $62.8_{4.4}$ & $56.7_{7.4}$ & $57.7_{6.8}$ & $73.4_{3.5}$ \\
\textsc{{Cobra}}-V2 & $98.8_{0.1}$ & $\underline{83.6}_{1.1}$ & $75.8_{2.7}$ & $79.7_{2.4}$ & $54.9_{6.9}$ & $88.8_{1.8}$ & $79.0_{0.8}$ & $\mathbf{70.9}_{3.0}$ & $\underline{66.6}_{3.8}$ & $\underline{94.7}_{1.3}$ & $83.7_{1.9}$ & $62.4_{0.6}$ & $\underline{63.5}_{10.6}$ & $\underline{61.0}_{2.8}$ & $51.8_{12.9}$ & $\underline{74.3}_{5.0}$ \\
\textsc{{Cobra}}$^\dag$-V2 & $98.9_{0.2}$ & $\underline{83.6}_{1.7}$ & $\mathbf{76.7}_{3.9}$ & $\underline{80.0}_{1.8}$ & $53.0_{4.4}$ & $\underline{89.6}_{1.6}$ & $79.5_{1.2}$ & $\underline{70.6}_{2.6}$ & $65.8_{4.8}$ & $\mathbf{95.1}_{0.9}$ & $82.5_{2.5}$ & $61.7_{0.5}$ & $58.4_{12.8}$ & $\mathbf{61.9}_{2.9}$ & $61.2_{3.2}$ & $\mathbf{74.6}_{4.2}$ \\
\bottomrule
\end{tabular}
}
\end{table*}

\begin{table*}[ht!]
\centering
\caption{\textbf{Few shot performance comparison.} AUC score of models on CPTAC datasets with k=5 positive samples during training on TCGA. $\overline{\text{Overline}}$ indicates mean over patch embeddings, $^\dag$ indicates that embeddings of all four training FMs were used to generate the weighting vector (\cref{eq:cobra-dagger}). \textbf{Bold} indicates the best performance, and $\underline{\text{underline}}$ indicates the second-best performance. The abbreviations are as follows: ST: Subtyping CTP: CTransPath~\cite{wang2022ctp}, H0: H-Optimus-0~\cite{hoptimus0}, V2: Virchow-2~\cite{zimmermann2024virchow2}, GP: GigaPath~\cite{xu2024gigapath}, SE: Slide Encoder.}
\label{tab:lp_auroc_5}
\resizebox{\textwidth}{!}{%
\begin{tabular}{l|l|lll|lll|lll|l}
\toprule
AUC[\%]-k=5 & LUNG & \multicolumn{3}{c|}{LUAD} & \multicolumn{3}{c|}{BRCA} & \multicolumn{3}{c|}{COAD} & Average \\
Model & ST & STK11 & EGFR & TP53 & ESR1 & PGR & ERBB2 & MSI & BRAF & Side &  \\
\midrule
$\overline{\text{Virchow}}$~\cite{vorontsov2024virchow} & $72.3_{15.4}$ & $60.8_{11.7}$ & $56.5_{7.7}$ & $51.1_{10.5}$ & $53.9_{7.6}$ & $49.1_{9.8}$ & $48.3_{5.7}$ & $52.8_{5.0}$ & $50.3_{6.9}$ & $50.2_{4.1}$ & $54.5_{9.1}$ \\
$\overline{\text{CTransPath}}$~\cite{wang2022ctp} & $64.1_{13.5}$ & $55.6_{8.5}$ & $56.6_{10.5}$ & $50.6_{10.2}$ & $58.2_{9.0}$ & $58.9_{5.4}$ & $49.3_{4.7}$ & $59.4_{5.1}$ & $47.8_{12.8}$ & $49.7_{5.1}$ & $55.0_{9.0}$ \\
$\overline{\text{H-Optimus}}$~\cite{hoptimus0} & $68.6_{17.1}$ & $63.9_{10.9}$ & $63.3_{10.6}$ & $51.3_{6.1}$ & $62.1_{9.7}$ & $51.1_{8.4}$ & $48.1_{5.2}$ & $71.9_{8.2}$ & $55.9_{15.9}$ & $51.6_{4.4}$ & $58.8_{10.4}$ \\
$\overline{\text{UNI}}$~\cite{chen2024uni} & $67.8_{15.5}$ & $60.8_{6.8}$ & $60.3_{12.2}$ & $53.3_{7.7}$ & $61.9_{11.0}$ & $59.8_{10.0}$ & $53.6_{7.2}$ & $67.4_{6.4}$ & $53.8_{9.6}$ & $50.7_{10.4}$ & $58.9_{10.0}$ \\
$\overline{\text{GigaPath}}$~\cite{xu2024gigapath} & $71.6_{13.9}$ & $58.1_{10.2}$ & $62.9_{11.2}$ & $54.1_{8.0}$ & $63.3_{11.0}$ & $58.5_{6.8}$ & $53.2_{7.1}$ & $69.9_{9.4}$ & $56.8_{10.1}$ & $\underline{52.6}_{6.6}$ & $60.1_{9.7}$ \\
$\overline{\text{CONCH}}$~\cite{lu2024conch} & $83.1_{8.8}$ & $61.8_{8.9}$ & $56.8_{8.7}$ & $54.8_{9.2}$ & $60.5_{12.7}$ & $64.8_{9.3}$ & $51.8_{9.0}$ & $66.2_{7.1}$ & $55.0_{5.8}$ & $52.3_{5.8}$ & $60.7_{8.7}$ \\
$\overline{\text{Virchow2}}$~\cite{zimmermann2024virchow2} & $72.4_{13.6}$ & $61.1_{7.3}$ & $62.6_{8.9}$ & $52.6_{9.8}$ & $65.6_{12.1}$ & $62.2_{7.2}$ & $56.9_{7.1}$ & $78.0_{6.9}$ & $59.7_{6.6}$ & $\mathbf{53.3}_{4.2}$ & $62.4_{8.8}$ \\
\hline
GigaPath-SE~\cite{xu2024gigapath} & $65.2_{9.4}$ & $57.2_{7.4}$ & $58.3_{4.7}$ & $52.5_{7.7}$ & $58.4_{8.0}$ & $54.0_{6.4}$ & $53.7_{11.0}$ & $54.4_{11.4}$ & $51.1_{11.4}$ & $47.2_{8.8}$ & $55.2_{8.9}$ \\
CHIEF~\cite{Wang2024Chief} & $73.5_{13.1}$ & $60.9_{7.7}$ & $58.7_{7.9}$ & $54.6_{8.5}$ & $63.1_{7.9}$ & $\mathbf{66.6}_{4.5}$ & $53.7_{7.1}$ & $64.2_{8.1}$ & $49.8_{12.8}$ & $48.7_{5.6}$ & $59.4_{8.7}$ \\
\textsc{{Cobra}}$^\dag$-CTP & $77.5_{11.3}$ & $62.0_{7.4}$ & $59.9_{9.6}$ & $60.6_{7.0}$ & $61.7_{6.6}$ & $60.2_{5.2}$ & $51.8_{5.0}$ & $61.7_{7.0}$ & $53.2_{14.2}$ & $47.7_{4.9}$ & $59.6_{8.3}$ \\
MADELEINE~\cite{jaume2024madeleine} & $87.8_{5.8}$ & $63.2_{7.4}$ & $59.5_{7.5}$ & $54.7_{8.6}$ & $62.6_{8.5}$ & $62.5_{11.0}$ & $\underline{59.3}_{7.6}$ & $68.3_{4.2}$ & $56.4_{7.1}$ & $52.2_{4.5}$ & $62.6_{7.5}$ \\
\textsc{{Cobra}}$^\dag$-UNI & $86.5_{8.4}$ & $\underline{71.8}_{6.2}$ & $62.8_{10.5}$ & $\underline{60.8}_{7.7}$ & $66.4_{10.2}$ & $61.7_{9.1}$ & $57.5_{11.6}$ & $71.7_{8.1}$ & $61.5_{10.5}$ & $49.1_{8.8}$ & $65.0_{9.2}$ \\
\textsc{{Cobra}}$^\dag$-H0 & $\underline{88.6}_{7.6}$ & $\mathbf{74.0}_{11.5}$ & $\mathbf{68.4}_{8.5}$ & $60.5_{7.5}$ & $64.9_{10.8}$ & $54.3_{7.9}$ & $52.8_{8.3}$ & $\underline{78.8}_{7.5}$ & $\underline{61.7}_{14.3}$ & $51.3_{5.7}$ & $65.5_{9.3}$ \\
PRISM~\cite{shaikovski2024prism} & $\mathbf{96.9}_{1.7}$ & $70.2_{9.6}$ & $59.0_{9.5}$ & $\mathbf{65.6}_{8.6}$ & $\mathbf{73.0}_{10.3}$ & $\underline{66.3}_{7.7}$ & $57.1_{9.7}$ & $71.2_{5.0}$ & $58.6_{5.3}$ & $52.1_{2.7}$ & $\underline{67.0}_{7.6}$ \\
\textsc{{Cobra}}$^\dag$-V2 & $86.7_{6.8}$ & $66.9_{9.0}$ & $\underline{63.4}_{7.6}$ & $59.4_{9.4}$ & $\underline{71.7}_{10.4}$ & $64.9_{6.3}$ & $\mathbf{59.8}_{9.7}$ & $\mathbf{82.2}_{8.5}$ & $\mathbf{66.6}_{9.9}$ & $51.0_{3.5}$ & $\mathbf{67.3}_{8.4}$ \\
\bottomrule
\end{tabular}
}
\end{table*}

\begin{table*}[ht!]
\centering
\caption{\textbf{Few shot performance comparison.} AUC score of models on CPTAC datasets with k=10 positive samples during training on TCGA. $\overline{\text{Overline}}$ indicates mean over patch embeddings, $^\dag$ indicates that embeddings of all four training FMs were used to generate the weighting vector (\cref{eq:cobra-dagger}). \textbf{Bold} indicates the best performance, and $\underline{\text{underline}}$ indicates the second-best performance. The abbreviations are as follows: ST: Subtyping CTP: CTransPath~\cite{wang2022ctp}, H0: H-Optimus-0~\cite{hoptimus0}, V2: Virchow-2~\cite{zimmermann2024virchow2}, GP: GigaPath~\cite{xu2024gigapath}, SE: Slide Encoder.}
\label{tab:lp_auroc_10}
\resizebox{\textwidth}{!}{%
\begin{tabular}{l|l|lll|lll|lll|l}
\toprule
AUC[\%]-k=10 & LUNG & \multicolumn{3}{c|}{LUAD} & \multicolumn{3}{c|}{BRCA} & \multicolumn{3}{c|}{COAD} & Average \\
Model & ST & STK11 & EGFR & TP53 & ESR1 & PGR & ERBB2 & MSI & BRAF & Side &  \\
\midrule
$\overline{\text{CTransPath}}$~\cite{wang2022ctp} & $63.8_{13.8}$ & $55.9_{7.1}$ & $56.1_{10.1}$ & $53.9_{8.4}$ & $66.0_{7.1}$ & $60.9_{5.6}$ & $49.7_{6.8}$ & $66.6_{9.3}$ & $54.6_{13.5}$ & $50.4_{4.7}$ & $57.8_{9.1}$ \\
$\overline{\text{Virchow}}$~\cite{vorontsov2024virchow} & $75.9_{8.8}$ & $62.7_{11.7}$ & $56.7_{8.6}$ & $56.4_{10.6}$ & $59.6_{12.4}$ & $58.9_{8.4}$ & $48.6_{5.7}$ & $62.6_{8.1}$ & $53.5_{4.0}$ & $50.5_{2.4}$ & $58.5_{8.6}$ \\
$\overline{\text{H-Optimus}}$~\cite{hoptimus0} & $74.6_{13.1}$ & $66.8_{7.2}$ & $63.5_{6.4}$ & $57.9_{9.0}$ & $71.6_{8.2}$ & $59.8_{7.7}$ & $51.1_{7.2}$ & $77.7_{8.9}$ & $62.5_{8.4}$ & $53.8_{5.0}$ & $63.9_{8.4}$ \\
$\overline{\text{CONCH}}$~\cite{lu2024conch} & $85.8_{7.4}$ & $60.9_{10.9}$ & $59.1_{6.2}$ & $60.6_{7.0}$ & $73.9_{7.9}$ & $66.9_{9.5}$ & $53.6_{7.5}$ & $68.1_{6.0}$ & $61.8_{5.8}$ & $\mathbf{55.4}_{5.8}$ & $64.6_{7.6}$ \\
$\overline{\text{UNI}}$~\cite{chen2024uni} & $73.4_{14.8}$ & $63.3_{9.2}$ & $62.9_{9.0}$ & $60.9_{8.3}$ & $70.6_{10.8}$ & $66.4_{5.6}$ & $55.6_{8.7}$ & $76.3_{7.4}$ & $63.5_{8.3}$ & $54.2_{7.3}$ & $64.7_{9.2}$ \\
$\overline{\text{GigaPath}}$~\cite{xu2024gigapath} & $78.7_{10.1}$ & $59.6_{8.3}$ & $65.2_{8.9}$ & $59.7_{8.1}$ & $72.1_{9.6}$ & $62.6_{5.1}$ & $56.6_{9.0}$ & $78.1_{9.2}$ & $65.2_{9.6}$ & $53.7_{4.5}$ & $65.2_{8.4}$ \\
$\overline{\text{Virchow2}}$~\cite{zimmermann2024virchow2} & $76.5_{7.1}$ & $60.2_{6.8}$ & $64.1_{7.1}$ & $59.2_{9.1}$ & $76.0_{8.3}$ & $67.4_{6.1}$ & $59.4_{5.1}$ & $\underline{82.6}_{8.5}$ & $70.0_{7.0}$ & $\underline{54.6}_{3.9}$ & $67.0_{7.1}$ \\
\hline
GigaPath-SE~\cite{xu2024gigapath} & $71.4_{7.8}$ & $60.0_{6.6}$ & $61.0_{7.1}$ & $57.6_{4.5}$ & $62.1_{3.1}$ & $57.7_{8.0}$ & $53.8_{10.3}$ & $55.2_{10.1}$ & $56.8_{9.4}$ & $48.9_{6.7}$ & $58.5_{7.7}$ \\
CHIEF~\cite{Wang2024Chief} & $76.2_{12.0}$ & $65.0_{4.4}$ & $60.2_{8.6}$ & $58.2_{8.0}$ & $70.8_{8.1}$ & $\underline{68.9}_{6.1}$ & $56.6_{8.8}$ & $71.8_{10.6}$ & $57.9_{13.5}$ & $50.1_{4.1}$ & $63.6_{8.9}$ \\
\textsc{{Cobra}}$^\dag$-CTP & $82.1_{9.7}$ & $67.0_{4.2}$ & $60.9_{7.9}$ & $64.1_{6.2}$ & $67.2_{6.0}$ & $61.0_{6.2}$ & $54.3_{6.0}$ & $71.3_{10.7}$ & $61.7_{12.7}$ & $47.7_{3.1}$ & $63.7_{7.8}$ \\
MADELEINE~\cite{jaume2024madeleine} & $90.0_{5.4}$ & $64.9_{7.5}$ & $60.9_{5.8}$ & $61.2_{7.7}$ & $74.5_{6.8}$ & $64.7_{10.2}$ & $\mathbf{63.0}_{6.2}$ & $71.0_{6.7}$ & $60.2_{4.7}$ & $54.0_{3.1}$ & $66.4_{6.7}$ \\
PRISM~\cite{shaikovski2024prism} & $\mathbf{97.8}_{0.7}$ & $\underline{74.9}_{9.3}$ & $63.0_{7.2}$ & $\mathbf{70.9}_{6.8}$ & $77.0_{7.7}$ & $\mathbf{72.5}_{6.6}$ & $58.7_{7.9}$ & $74.4_{3.8}$ & $62.0_{8.1}$ & $51.5_{3.8}$ & $70.3_{6.7}$ \\
\textsc{{Cobra}}$^\dag$-H0 & $\underline{92.7}_{4.3}$ & $\mathbf{78.8}_{3.9}$ & $\mathbf{72.6}_{4.2}$ & $67.5_{6.5}$ & $75.5_{5.9}$ & $59.4_{9.1}$ & $54.0_{8.8}$ & $\underline{82.6}_{7.4}$ & $67.5_{8.7}$ & $52.4_{5.9}$ & $70.3_{6.7}$ \\
\textsc{{Cobra}}$^\dag$-UNI & $91.0_{5.7}$ & $73.5_{6.2}$ & $\underline{69.7}_{5.3}$ & $\underline{69.4}_{7.1}$ & $\underline{77.1}_{6.3}$ & $63.6_{6.6}$ & $58.2_{8.4}$ & $78.9_{5.9}$ & $\underline{70.6}_{6.7}$ & $51.9_{5.9}$ & $\underline{70.4}_{6.5}$ \\
\textsc{{Cobra}}$^\dag$-V2 & $90.7_{4.0}$ & $71.4_{3.8}$ & $69.3_{6.2}$ & $68.8_{6.1}$ & $\mathbf{78.2}_{6.1}$ & $64.4_{7.8}$ & $\underline{62.7}_{7.2}$ & $\mathbf{85.3}_{5.5}$ & $\mathbf{76.6}_{7.7}$ & $53.2_{4.2}$ & $\mathbf{72.1}_{6.0}$ \\
\bottomrule
\end{tabular}
}
\end{table*}

\begin{table*}[ht!]
\centering
\caption{\textbf{Few shot performance comparison.} AUC score of models on CPTAC datasets with k=25 positive samples during training on TCGA. $\overline{\text{Overline}}$ indicates mean over patch embeddings, $^\dag$ indicates that embeddings of all four training FMs were used to generate the weighting vector (\cref{eq:cobra-dagger}). \textbf{Bold} indicates the best performance, and $\underline{\text{underline}}$ indicates the second-best performance. The abbreviations are as follows: ST: Subtyping CTP: CTransPath~\cite{wang2022ctp}, H0: H-Optimus-0~\cite{hoptimus0}, V2: Virchow-2~\cite{zimmermann2024virchow2}, GP: GigaPath~\cite{xu2024gigapath}, SE: Slide Encoder.}
\label{tab:lp_auroc_25}
\resizebox{\textwidth}{!}{%
\begin{tabular}{l|l|lll|lll|lll|l}
\toprule
AUC[\%]-k=25 & LUNG & \multicolumn{3}{c|}{LUAD} & \multicolumn{3}{c|}{BRCA} & \multicolumn{3}{c|}{COAD} & Average \\
Model & ST & STK11 & EGFR & TP53 & ESR1 & PGR & ERBB2 & MSI & BRAF & Side &  \\
\midrule
$\overline{\text{CTransPath}}$~\cite{wang2022ctp} & $71.8_{14.1}$ & $61.1_{6.7}$ & $60.6_{5.6}$ & $60.3_{8.2}$ & $67.6_{6.5}$ & $62.3_{7.2}$ & $50.3_{5.8}$ & $76.2_{6.2}$ & $63.9_{12.0}$ & $53.0_{4.0}$ & $62.7_{8.2}$ \\
$\overline{\text{Virchow}}$~\cite{vorontsov2024virchow} & $79.9_{9.3}$ & $68.0_{7.8}$ & $64.2_{8.0}$ & $60.4_{9.8}$ & $61.6_{13.1}$ & $58.1_{9.9}$ & $53.8_{6.9}$ & $72.4_{10.8}$ & $62.2_{5.8}$ & $52.6_{5.0}$ & $63.3_{8.9}$ \\
$\overline{\text{UNI}}$~\cite{chen2024uni} & $80.1_{10.6}$ & $65.3_{9.8}$ & $69.0_{6.8}$ & $65.1_{7.2}$ & $73.4_{11.2}$ & $63.9_{6.7}$ & $56.8_{6.8}$ & $83.5_{5.0}$ & $67.3_{8.3}$ & $57.7_{3.3}$ & $68.2_{7.9}$ \\
$\overline{\text{H-Optimus}}$~\cite{hoptimus0} & $82.4_{9.1}$ & $70.9_{9.4}$ & $72.1_{3.3}$ & $61.5_{6.4}$ & $74.5_{6.8}$ & $58.6_{10.8}$ & $49.1_{5.3}$ & $85.3_{7.5}$ & $70.5_{8.6}$ & $\mathbf{59.7}_{4.6}$ & $68.5_{7.5}$ \\
$\overline{\text{GigaPath}}$~\cite{xu2024gigapath} & $82.2_{10.0}$ & $66.2_{7.8}$ & $73.7_{4.0}$ & $63.9_{8.2}$ & $75.2_{8.2}$ & $62.6_{8.5}$ & $59.5_{7.1}$ & $82.5_{9.2}$ & $68.8_{7.8}$ & $56.4_{2.7}$ & $69.1_{7.7}$ \\
$\overline{\text{CONCH}}$~\cite{lu2024conch} & $91.3_{5.3}$ & $70.1_{8.8}$ & $66.0_{7.3}$ & $64.8_{6.8}$ & $76.4_{7.3}$ & $66.7_{10.9}$ & $56.4_{6.2}$ & $77.8_{5.3}$ & $68.1_{6.8}$ & $58.4_{5.8}$ & $69.6_{7.2}$ \\
$\overline{\text{Virchow2}}$~\cite{zimmermann2024virchow2} & $83.9_{7.7}$ & $69.5_{7.6}$ & $71.0_{4.2}$ & $63.0_{8.2}$ & $79.0_{5.5}$ & $66.3_{10.2}$ & $63.9_{6.2}$ & $89.1_{4.6}$ & $74.5_{6.1}$ & $\underline{58.8}_{3.4}$ & $71.9_{6.7}$ \\
\hline
GigaPath-SE~\cite{xu2024gigapath} & $77.8_{9.0}$ & $64.0_{6.4}$ & $63.0_{4.0}$ & $61.4_{6.8}$ & $56.9_{7.7}$ & $59.4_{8.9}$ & $55.8_{8.8}$ & $61.9_{6.3}$ & $61.6_{9.3}$ & $49.5_{5.1}$ & $61.1_{7.4}$ \\
CHIEF~\cite{Wang2024Chief} & $84.3_{11.0}$ & $69.4_{6.2}$ & $67.1_{6.0}$ & $65.6_{7.1}$ & $74.3_{5.9}$ & $\mathbf{70.6}_{5.7}$ & $55.5_{7.5}$ & $78.0_{7.3}$ & $65.0_{13.6}$ & $50.9_{3.5}$ & $68.1_{7.9}$ \\
\textsc{{Cobra}}$^\dag$-CTP & $88.6_{7.6}$ & $70.2_{6.1}$ & $69.0_{5.3}$ & $70.3_{6.0}$ & $72.5_{4.4}$ & $63.7_{7.4}$ & $51.9_{6.6}$ & $80.1_{6.5}$ & $68.6_{9.6}$ & $49.8_{3.4}$ & $68.5_{6.5}$ \\
MADELEINE~\cite{jaume2024madeleine} & $93.4_{4.4}$ & $70.8_{6.9}$ & $67.3_{6.0}$ & $66.7_{6.4}$ & $77.7_{6.5}$ & $65.2_{9.6}$ & $\mathbf{66.3}_{3.2}$ & $77.1_{4.0}$ & $60.5_{3.4}$ & $56.1_{4.0}$ & $70.1_{5.8}$ \\
PRISM~\cite{shaikovski2024prism} & $\mathbf{98.1}_{0.6}$ & $\mathbf{82.6}_{5.0}$ & $73.2_{5.3}$ & $\underline{72.2}_{4.4}$ & $\underline{79.1}_{6.8}$ & $\underline{70.5}_{4.2}$ & $59.7_{7.4}$ & $78.2_{3.5}$ & $62.9_{6.0}$ & $51.1_{3.5}$ & $72.8_{5.0}$ \\
\textsc{{Cobra}}$^\dag$-UNI & $94.2_{3.7}$ & $73.1_{5.6}$ & $74.2_{6.2}$ & $\mathbf{73.3}_{5.0}$ & $77.6_{8.6}$ & $66.6_{8.2}$ & $57.3_{7.9}$ & $84.5_{5.2}$ & $75.1_{7.8}$ & $55.6_{5.4}$ & $73.2_{6.6}$ \\
\textsc{{Cobra}}$^\dag$-H0 & $\underline{95.5}_{3.3}$ & $\underline{79.0}_{4.7}$ & $\mathbf{78.1}_{3.8}$ & $70.7_{5.8}$ & $75.7_{7.3}$ & $60.0_{11.4}$ & $51.8_{5.8}$ & $\underline{89.6}_{4.3}$ & $\underline{76.1}_{7.6}$ & $56.5_{6.6}$ & $\underline{73.3}_{6.5}$ \\
\textsc{{Cobra}}$^\dag$-V2 & $93.4_{4.9}$ & $73.9_{4.7}$ & $\underline{75.3}_{5.2}$ & $71.7_{7.4}$ & $\mathbf{81.6}_{4.9}$ & $65.7_{10.8}$ & $\underline{64.8}_{5.7}$ & $\mathbf{90.3}_{4.1}$ & $\mathbf{82.2}_{5.2}$ & $58.3_{4.8}$ & $\mathbf{75.7}_{6.1}$ \\
\bottomrule
\end{tabular}
}
\end{table*}

\begin{table*}[ht!]
\centering
\caption{\textbf{Few shot performance comparison.} AUPRC score of models on CPTAC datasets with k=5 positive samples during training on TCGA. $\overline{\text{Overline}}$ indicates mean over patch embeddings, $^\dag$ indicates that embeddings of all four training FMs were used to generate the weighting vector (\cref{eq:cobra-dagger}). \textbf{Bold} indicates the best performance, and $\underline{\text{underline}}$ indicates the second-best performance. The abbreviations are as follows: ST: Subtyping CTP: CTransPath~\cite{wang2022ctp}, H0: H-Optimus-0~\cite{hoptimus0}, V2: Virchow-2~\cite{zimmermann2024virchow2}, GP: GigaPath~\cite{xu2024gigapath}, SE: Slide Encoder.}
\label{tab:lp_auprc_5}
\resizebox{\textwidth}{!}{%
\begin{tabular}{l|l|lll|lll|lll|l}
\toprule
AUPRC[\%]-k=5 & LUNG & \multicolumn{3}{c|}{LUAD} & \multicolumn{3}{c|}{BRCA} & \multicolumn{3}{c|}{COAD} & Average \\
Model & ST & STK11 & EGFR & TP53 & ESR1 & PGR & ERBB2 & MSI & BRAF & Side &  \\
\midrule
$\overline{\text{Virchow}}$~\cite{vorontsov2024virchow} & $72.1_{14.8}$ & $25.3_{10.0}$ & $39.3_{6.7}$ & $55.0_{6.9}$ & $67.6_{3.9}$ & $59.8_{7.2}$ & $13.1_{2.1}$ & $25.8_{3.3}$ & $17.5_{5.3}$ & $54.4_{3.7}$ & $43.0_{7.3}$ \\
$\overline{\text{CTransPath}}$~\cite{wang2022ctp} & $63.7_{12.5}$ & $22.7_{5.9}$ & $40.8_{10.2}$ & $55.9_{9.0}$ & $73.0_{5.8}$ & $68.7_{4.8}$ & $13.6_{1.9}$ & $36.2_{6.2}$ & $17.9_{7.9}$ & $55.6_{4.1}$ & $44.8_{7.4}$ \\
$\overline{\text{UNI}}$~\cite{chen2024uni} & $70.5_{13.0}$ & $26.7_{7.3}$ & $46.0_{11.9}$ & $56.1_{6.1}$ & $75.7_{8.6}$ & $67.9_{6.9}$ & $16.4_{5.1}$ & $43.3_{6.3}$ & $22.1_{9.0}$ & $55.2_{7.2}$ & $48.0_{8.5}$ \\
$\overline{\text{H-Optimus}}$~\cite{hoptimus0} & $70.7_{15.0}$ & $31.8_{13.0}$ & $\underline{48.5}_{11.4}$ & $54.6_{4.3}$ & $75.1_{7.2}$ & $62.4_{6.4}$ & $12.7_{1.8}$ & $47.3_{11.7}$ & $25.4_{11.4}$ & $56.6_{3.7}$ & $48.5_{9.6}$ \\
$\overline{\text{GigaPath}}$~\cite{xu2024gigapath} & $71.0_{13.9}$ & $25.0_{7.8}$ & $48.1_{12.3}$ & $57.3_{6.4}$ & $77.5_{8.0}$ & $67.2_{5.7}$ & $14.2_{2.5}$ & $46.8_{10.4}$ & $21.5_{6.3}$ & $\mathbf{57.7}_{5.5}$ & $48.6_{8.5}$ \\
$\overline{\text{CONCH}}$~\cite{lu2024conch} & $83.7_{10.2}$ & $27.2_{10.8}$ & $39.0_{6.9}$ & $58.1_{7.9}$ & $75.0_{9.2}$ & $72.3_{6.3}$ & $14.5_{4.4}$ & $44.9_{7.2}$ & $21.1_{4.4}$ & $56.8_{3.9}$ & $49.3_{7.5}$ \\
$\overline{\text{Virchow2}}$~\cite{zimmermann2024virchow2} & $73.3_{13.5}$ & $24.7_{5.8}$ & $48.2_{10.7}$ & $56.5_{8.2}$ & $78.0_{8.4}$ & $69.7_{5.8}$ & $16.1_{3.3}$ & $54.6_{9.6}$ & $\underline{26.5}_{10.4}$ & $\underline{57.6}_{3.6}$ & $50.5_{8.5}$ \\
\hline
GigaPath-SE~\cite{xu2024gigapath} & $65.9_{9.6}$ & $25.2_{8.2}$ & $40.7_{5.0}$ & $57.2_{8.0}$ & $72.8_{5.7}$ & $63.1_{6.5}$ & $15.8_{5.8}$ & $31.7_{7.8}$ & $18.1_{7.8}$ & $52.2_{6.2}$ & $44.3_{7.2}$ \\
CHIEF~\cite{Wang2024Chief} & $73.2_{14.4}$ & $25.1_{5.5}$ & $42.4_{8.0}$ & $58.8_{8.3}$ & $75.4_{4.9}$ & $\mathbf{73.5}_{3.2}$ & $14.5_{2.5}$ & $40.8_{9.0}$ & $19.7_{8.5}$ & $54.7_{4.6}$ & $47.8_{7.7}$ \\
\textsc{{Cobra}}$^\dag$-CTP & $78.2_{13.1}$ & $26.6_{6.3}$ & $43.6_{8.7}$ & $63.1_{7.3}$ & $75.2_{3.8}$ & $69.3_{4.3}$ & $15.1_{2.6}$ & $38.7_{9.5}$ & $21.5_{9.9}$ & $54.3_{4.1}$ & $48.6_{7.6}$ \\
MADELEINE~\cite{jaume2024madeleine} & $88.6_{6.1}$ & $28.3_{9.0}$ & $41.7_{8.1}$ & $58.8_{8.5}$ & $77.5_{6.3}$ & $70.9_{9.2}$ & $\mathbf{20.5}_{6.6}$ & $50.6_{4.5}$ & $24.2_{6.8}$ & $56.1_{2.7}$ & $51.7_{7.1}$ \\
\textsc{{Cobra}}$^\dag$-UNI & $87.0_{11.1}$ & $\underline{33.4}_{9.2}$ & $47.1_{11.6}$ & $\underline{63.4}_{7.8}$ & $79.1_{7.2}$ & $69.8_{7.3}$ & $17.7_{6.4}$ & $49.0_{8.9}$ & $25.7_{9.3}$ & $55.2_{6.8}$ & $52.7_{8.7}$ \\
PRISM~\cite{shaikovski2024prism} & $\mathbf{96.8}_{2.1}$ & $31.9_{7.6}$ & $40.0_{9.1}$ & $\mathbf{64.9}_{5.9}$ & $\mathbf{83.3}_{7.1}$ & $\underline{72.4}_{7.4}$ & $18.1_{4.7}$ & $47.6_{4.5}$ & $24.7_{5.3}$ & $56.9_{3.2}$ & $53.7_{6.0}$ \\
\textsc{{Cobra}}$^\dag$-H0 & $\underline{88.9}_{9.0}$ & $\mathbf{40.5}_{17.1}$ & $\mathbf{53.4}_{10.2}$ & $62.9_{6.8}$ & $76.9_{7.6}$ & $63.7_{6.5}$ & $16.3_{4.5}$ & $\underline{59.6}_{12.5}$ & $25.9_{10.6}$ & $55.6_{4.9}$ & $\underline{54.4}_{9.7}$ \\
\textsc{{Cobra}}$^\dag$-V2 & $87.5_{7.7}$ & $29.5_{7.1}$ & $46.7_{7.8}$ & $61.8_{8.2}$ & $\underline{82.3}_{7.0}$ & $71.5_{5.7}$ & $\underline{19.7}_{5.0}$ & $\mathbf{64.0}_{10.3}$ & $\mathbf{31.4}_{9.2}$ & $56.9_{3.7}$ & $\mathbf{55.1}_{7.4}$ \\
\bottomrule
\end{tabular}
}
\end{table*}

\begin{table*}[ht!]
\centering
\caption{\textbf{Few shot performance comparison.} AUPRC score of models on CPTAC datasets with k=10 positive samples during training on TCGA. $\overline{\text{Overline}}$ indicates mean over patch embeddings, $^\dag$ indicates that embeddings of all four training FMs were used to generate the weighting vector (\cref{eq:cobra-dagger}). \textbf{Bold} indicates the best performance, and $\underline{\text{underline}}$ indicates the second-best performance. The abbreviations are as follows: ST: Subtyping CTP: CTransPath~\cite{wang2022ctp}, H0: H-Optimus-0~\cite{hoptimus0}, V2: Virchow-2~\cite{zimmermann2024virchow2}, GP: GigaPath~\cite{xu2024gigapath}, SE: Slide Encoder.}
\label{tab:lp_auprc_10}
\resizebox{\textwidth}{!}{%
\begin{tabular}{l|l|lll|lll|lll|l}
\toprule
AUPRC[\%]-k=10 & LUNG & \multicolumn{3}{c|}{LUAD} & \multicolumn{3}{c|}{BRCA} & \multicolumn{3}{c|}{COAD} & Average \\
Model & ST & STK11 & EGFR & TP53 & ESR1 & PGR & ERBB2 & MSI & BRAF & Side &  \\
\midrule
$\overline{\text{Virchow}}$~\cite{vorontsov2024virchow} & $77.4_{8.1}$ & $26.3_{9.9}$ & $40.3_{9.5}$ & $59.3_{8.2}$ & $74.4_{8.0}$ & $66.6_{5.6}$ & $13.3_{2.8}$ & $35.9_{9.7}$ & $19.5_{3.9}$ & $55.5_{2.7}$ & $46.9_{7.4}$ \\
$\overline{\text{CTransPath}}$~\cite{wang2022ctp} & $64.1_{12.9}$ & $21.2_{4.9}$ & $40.6_{9.4}$ & $57.7_{7.4}$ & $78.9_{5.2}$ & $70.1_{4.4}$ & $13.7_{2.8}$ & $43.1_{9.2}$ & $25.9_{11.6}$ & $57.1_{4.1}$ & $47.2_{7.9}$ \\
$\overline{\text{UNI}}$~\cite{chen2024uni} & $76.2_{12.3}$ & $25.6_{6.3}$ & $48.7_{10.3}$ & $62.9_{6.9}$ & $82.5_{7.1}$ & $73.1_{3.8}$ & $15.4_{4.4}$ & $55.7_{12.4}$ & $28.6_{9.5}$ & $\underline{59.2}_{5.6}$ & $52.8_{8.4}$ \\
$\overline{\text{CONCH}}$~\cite{lu2024conch} & $86.6_{7.4}$ & $27.3_{9.3}$ & $40.9_{7.5}$ & $62.8_{7.6}$ & $84.9_{5.1}$ & $73.9_{6.7}$ & $15.7_{3.7}$ & $44.7_{8.0}$ & $31.1_{6.8}$ & $\mathbf{59.6}_{3.4}$ & $52.8_{6.8}$ \\
$\overline{\text{H-Optimus}}$~\cite{hoptimus0} & $76.9_{11.2}$ & $32.7_{12.5}$ & $48.6_{7.4}$ & $61.2_{6.8}$ & $82.7_{5.4}$ & $68.3_{6.7}$ & $14.8_{3.9}$ & $59.6_{14.4}$ & $31.0_{9.1}$ & $58.5_{4.1}$ & $53.4_{8.8}$ \\
$\overline{\text{GigaPath}}$~\cite{xu2024gigapath} & $80.0_{9.1}$ & $24.7_{8.2}$ & $51.2_{10.7}$ & $62.0_{6.3}$ & $83.5_{6.5}$ & $69.7_{3.8}$ & $18.0_{5.4}$ & $59.1_{11.8}$ & $31.9_{9.7}$ & $58.9_{5.0}$ & $53.9_{8.1}$ \\
$\overline{\text{Virchow2}}$~\cite{zimmermann2024virchow2} & $77.6_{6.8}$ & $25.0_{4.5}$ & $48.9_{9.9}$ & $62.4_{7.9}$ & $\underline{86.5}_{5.0}$ & $74.0_{4.2}$ & $18.2_{3.2}$ & $62.1_{14.2}$ & $\underline{34.5}_{10.5}$ & $58.0_{3.3}$ & $54.7_{7.8}$ \\
\hline
GigaPath-SE~\cite{xu2024gigapath} & $72.6_{8.0}$ & $23.6_{4.9}$ & $43.6_{7.0}$ & $59.7_{4.6}$ & $74.7_{2.7}$ & $64.9_{6.0}$ & $15.4_{5.5}$ & $33.6_{8.3}$ & $19.9_{5.4}$ & $52.7_{4.1}$ & $46.1_{5.9}$ \\
CHIEF~\cite{Wang2024Chief} & $76.4_{13.5}$ & $26.6_{5.3}$ & $45.0_{9.0}$ & $61.6_{7.4}$ & $81.6_{4.7}$ & $\underline{75.1}_{5.9}$ & $17.6_{4.6}$ & $49.1_{14.3}$ & $26.1_{11.6}$ & $56.5_{3.9}$ & $51.6_{8.8}$ \\
\textsc{{Cobra}}$^\dag$-CTP & $83.3_{10.9}$ & $28.1_{5.7}$ & $46.0_{7.0}$ & $66.2_{5.9}$ & $79.1_{4.1}$ & $68.4_{5.1}$ & $16.2_{4.0}$ & $50.5_{14.8}$ & $29.0_{12.1}$ & $54.2_{3.2}$ & $52.1_{8.2}$ \\
MADELEINE~\cite{jaume2024madeleine} & $90.7_{5.3}$ & $26.9_{6.3}$ & $44.5_{8.1}$ & $63.6_{6.4}$ & $85.5_{3.8}$ & $72.0_{8.1}$ & $\underline{19.9}_{4.5}$ & $49.6_{6.9}$ & $30.7_{5.5}$ & $58.8_{2.1}$ & $54.2_{6.0}$ \\
PRISM~\cite{shaikovski2024prism} & $\mathbf{98.0}_{0.8}$ & $\underline{35.8}_{11.3}$ & $45.4_{8.3}$ & $69.2_{4.6}$ & $85.8_{4.9}$ & $\mathbf{76.1}_{5.4}$ & $\underline{19.9}_{5.9}$ & $50.8_{5.8}$ & $25.5_{3.9}$ & $56.8_{2.5}$ & $56.3_{6.0}$ \\
\textsc{{Cobra}}$^\dag$-UNI & $92.5_{4.3}$ & $32.6_{6.1}$ & $\underline{53.6}_{7.4}$ & $\mathbf{71.4}_{6.3}$ & $85.8_{4.5}$ & $71.1_{5.2}$ & $17.1_{4.4}$ & $60.5_{11.0}$ & $\underline{34.5}_{9.1}$ & $57.2_{4.4}$ & $57.6_{6.6}$ \\
\textsc{{Cobra}}$^\dag$-H0 & $\underline{93.6}_{3.7}$ & $\mathbf{42.3}_{6.6}$ & $\mathbf{56.9}_{5.9}$ & $69.1_{6.1}$ & $84.1_{4.5}$ & $68.5_{7.9}$ & $16.6_{4.9}$ & $\underline{65.3}_{12.5}$ & $31.3_{9.5}$ & $56.3_{5.6}$ & $\underline{58.4}_{7.2}$ \\
\textsc{{Cobra}}$^\dag$-V2 & $92.1_{3.2}$ & $30.3_{3.0}$ & $53.4_{7.2}$ & $\underline{70.9}_{5.5}$ & $\mathbf{87.3}_{3.8}$ & $72.1_{6.2}$ & $\mathbf{21.0}_{5.1}$ & $\mathbf{67.7}_{11.4}$ & $\mathbf{42.5}_{11.4}$ & $57.3_{3.3}$ & $\mathbf{59.5}_{6.7}$ \\
\bottomrule
\end{tabular}
}
\end{table*}

\begin{table*}[ht!]
\centering
\caption{\textbf{Few shot performance comparison.} AUPRC score of models on CPTAC datasets with k=25 positive samples during training on TCGA. $\overline{\text{Overline}}$ indicates mean over patch embeddings, $^\dag$ indicates that embeddings of all four training FMs were used to generate the weighting vector (\cref{eq:cobra-dagger}). \textbf{Bold} indicates the best performance, and $\underline{\text{underline}}$ indicates the second-best performance. The abbreviations are as follows: ST: Subtyping CTP: CTransPath~\cite{wang2022ctp}, H0: H-Optimus-0~\cite{hoptimus0}, V2: Virchow-2~\cite{zimmermann2024virchow2}, GP: GigaPath~\cite{xu2024gigapath}, SE: Slide Encoder.}
\label{tab:lp_auprc_25}
\resizebox{\textwidth}{!}{%
\begin{tabular}{l|l|lll|lll|lll|l}
\toprule
AUPRC[\%]-k=25 & LUNG & \multicolumn{3}{c|}{LUAD} & \multicolumn{3}{c|}{BRCA} & \multicolumn{3}{c|}{COAD} & Average \\
Model & ST & STK11 & EGFR & TP53 & ESR1 & PGR & ERBB2 & MSI & BRAF & Side &  \\
\midrule
$\overline{\text{Virchow}}$~\cite{vorontsov2024virchow} & $80.6_{8.2}$ & $30.2_{9.6}$ & $47.8_{8.3}$ & $62.5_{8.3}$ & $75.8_{8.2}$ & $65.9_{7.1}$ & $16.6_{3.9}$ & $48.7_{15.4}$ & $24.7_{5.7}$ & $57.0_{4.0}$ & $51.0_{8.5}$ \\
$\overline{\text{CTransPath}}$~\cite{wang2022ctp} & $72.9_{13.8}$ & $25.4_{4.7}$ & $44.5_{7.2}$ & $63.1_{7.3}$ & $79.6_{4.8}$ & $70.5_{6.1}$ & $13.3_{2.4}$ & $57.8_{9.5}$ & $31.2_{9.5}$ & $57.1_{3.1}$ & $51.5_{7.6}$ \\
$\overline{\text{UNI}}$~\cite{chen2024uni} & $81.1_{9.1}$ & $26.4_{6.6}$ & $53.5_{9.6}$ & $66.1_{5.9}$ & $83.6_{8.0}$ & $70.2_{5.1}$ & $18.0_{6.4}$ & $64.2_{7.8}$ & $30.0_{9.6}$ & $60.9_{3.8}$ & $55.4_{7.4}$ \\
$\overline{\text{CONCH}}$~\cite{lu2024conch} & $92.0_{4.6}$ & $35.6_{10.5}$ & $47.5_{10.7}$ & $67.0_{7.3}$ & $85.7_{4.7}$ & $72.7_{8.2}$ & $17.1_{4.7}$ & $56.2_{6.8}$ & $30.8_{8.5}$ & $60.9_{4.7}$ & $56.5_{7.4}$ \\
$\overline{\text{GigaPath}}$~\cite{xu2024gigapath} & $83.7_{8.8}$ & $27.5_{6.3}$ & $\underline{59.3}_{6.8}$ & $64.0_{5.9}$ & $85.3_{5.5}$ & $69.2_{5.7}$ & $18.9_{6.0}$ & $66.0_{12.3}$ & $33.7_{11.8}$ & $59.9_{3.1}$ & $56.8_{7.7}$ \\
$\overline{\text{H-Optimus}}$~\cite{hoptimus0} & $83.0_{8.6}$ & $34.4_{9.8}$ & $57.0_{5.0}$ & $63.5_{4.1}$ & $85.4_{4.1}$ & $66.2_{7.3}$ & $16.1_{4.0}$ & $70.9_{10.8}$ & $34.4_{9.4}$ & $\mathbf{62.0}_{3.9}$ & $57.3_{7.2}$ \\
$\overline{\text{Virchow2}}$~\cite{zimmermann2024virchow2} & $84.9_{6.5}$ & $30.2_{7.3}$ & $54.3_{7.5}$ & $64.6_{6.8}$ & $\underline{88.5}_{2.9}$ & $72.5_{7.9}$ & $22.3_{5.3}$ & $72.8_{10.2}$ & $32.8_{10.9}$ & $61.2_{3.7}$ & $58.4_{7.3}$ \\
\hline
GigaPath-SE~\cite{xu2024gigapath} & $80.1_{7.1}$ & $25.4_{5.2}$ & $46.2_{6.5}$ & $63.2_{5.5}$ & $70.9_{5.8}$ & $66.6_{6.4}$ & $17.9_{5.4}$ & $38.2_{7.3}$ & $23.0_{8.0}$ & $54.0_{3.8}$ & $48.5_{6.2}$ \\
CHIEF~\cite{Wang2024Chief} & $85.2_{11.0}$ & $30.9_{4.5}$ & $52.2_{7.0}$ & $67.3_{7.2}$ & $83.8_{4.2}$ & $\mathbf{76.3}_{5.1}$ & $16.6_{5.0}$ & $58.6_{12.7}$ & $29.9_{9.2}$ & $54.9_{3.2}$ & $55.6_{7.5}$ \\
\textsc{{Cobra}}$^\dag$-CTP & $90.2_{6.5}$ & $31.6_{5.2}$ & $53.1_{3.5}$ & $70.7_{6.1}$ & $83.2_{2.2}$ & $71.1_{6.2}$ & $16.7_{5.7}$ & $64.5_{10.6}$ & $34.9_{10.6}$ & $54.0_{2.4}$ & $57.0_{6.5}$ \\
MADELEINE~\cite{jaume2024madeleine} & $94.1_{3.8}$ & $34.1_{9.7}$ & $50.7_{9.4}$ & $69.2_{5.3}$ & $86.7_{3.6}$ & $72.6_{6.8}$ & $\mathbf{24.1}_{6.7}$ & $57.5_{4.0}$ & $29.5_{4.0}$ & $\underline{61.4}_{3.3}$ & $58.0_{6.1}$ \\
PRISM~\cite{shaikovski2024prism} & $\mathbf{98.3}_{0.8}$ & $\mathbf{45.9}_{8.9}$ & $54.0_{7.4}$ & $69.6_{4.4}$ & $86.9_{4.3}$ & $\underline{75.1}_{3.6}$ & $17.9_{3.6}$ & $53.0_{6.3}$ & $25.8_{6.3}$ & $57.4_{2.9}$ & $58.4_{5.3}$ \\
\textsc{{Cobra}}$^\dag$-UNI & $95.0_{2.9}$ & $32.5_{7.2}$ & $57.8_{6.2}$ & $\mathbf{74.5}_{5.0}$ & $86.2_{6.1}$ & $72.6_{7.1}$ & $18.9_{6.1}$ & $67.8_{7.5}$ & $36.3_{9.2}$ & $58.8_{4.4}$ & $60.0_{6.4}$ \\
\textsc{{Cobra}}$^\dag$-H0 & $\underline{96.0}_{2.7}$ & $\underline{41.8}_{7.4}$ & $\mathbf{63.8}_{4.2}$ & $71.5_{6.0}$ & $85.9_{5.1}$ & $68.8_{10.0}$ & $17.1_{4.7}$ & $\mathbf{79.0}_{6.9}$ & $\underline{38.9}_{7.1}$ & $59.7_{4.7}$ & $\underline{62.2}_{6.2}$ \\
\textsc{{Cobra}}$^\dag$-V2 & $94.4_{3.7}$ & $34.5_{5.8}$ & $59.2_{5.7}$ & $\underline{72.6}_{7.2}$ & $\mathbf{89.7}_{2.4}$ & $72.2_{8.9}$ & $\underline{24.0}_{7.8}$ & $\underline{75.0}_{8.8}$ & $\mathbf{43.2}_{8.4}$ & $59.6_{3.6}$ & $\mathbf{62.4}_{6.6}$ \\
\bottomrule
\end{tabular}
}
\end{table*}

\begin{table*}[ht!]
\centering
\caption{\textbf{Few shot performance comparison.} F1 score of models on CPTAC datasets with k=5 positive samples during training on TCGA. $\overline{\text{Overline}}$ indicates mean over patch embeddings, $^\dag$ indicates that embeddings of all four training FMs were used to generate the weighting vector (\cref{eq:cobra-dagger}). \textbf{Bold} indicates the best performance, and $\underline{\text{underline}}$ indicates the second-best performance. The abbreviations are as follows: ST: Subtyping CTP: CTransPath~\cite{wang2022ctp}, H0: H-Optimus-0~\cite{hoptimus0}, V2: Virchow-2~\cite{zimmermann2024virchow2}, GP: GigaPath~\cite{xu2024gigapath}, SE: Slide Encoder.}
\label{tab:lp_f1_5}
\resizebox{\textwidth}{!}{%
\begin{tabular}{l|l|lll|lll|lll|l}
\toprule
F1[\%]-k=5 & LUNG & \multicolumn{3}{c|}{LUAD} & \multicolumn{3}{c|}{BRCA} & \multicolumn{3}{c|}{COAD} & Average \\
Model & ST & STK11 & EGFR & TP53 & ESR1 & PGR & ERBB2 & MSI & BRAF & Side &  \\
\midrule
$\overline{\text{Virchow}}$~\cite{vorontsov2024virchow} & $50.4_{29.7}$ & $16.9_{12.5}$ & $33.0_{19.5}$ & $26.5_{27.4}$ & $50.0_{27.8}$ & $39.9_{25.3}$ & $9.9_{8.7}$ & $36.6_{2.3}$ & $\underline{24.0}_{3.2}$ & $41.1_{25.0}$ & $32.8_{20.7}$ \\
$\overline{\text{UNI}}$~\cite{chen2024uni} & $48.3_{21.2}$ & $23.1_{9.5}$ & $33.5_{16.3}$ & $30.4_{20.0}$ & $58.4_{22.9}$ & $53.2_{21.0}$ & $16.4_{8.5}$ & $36.5_{13.2}$ & $23.8_{1.9}$ & $41.6_{28.5}$ & $36.5_{18.0}$ \\
$\overline{\text{H-Optimus}}$~\cite{hoptimus0} & $58.4_{19.2}$ & $27.2_{10.4}$ & $41.1_{11.1}$ & $30.9_{22.5}$ & $54.4_{34.1}$ & $49.5_{29.0}$ & $10.5_{9.4}$ & $31.9_{16.7}$ & $22.5_{8.1}$ & $42.7_{28.0}$ & $36.9_{20.8}$ \\
$\overline{\text{CTransPath}}$~\cite{wang2022ctp} & $58.1_{18.8}$ & $25.0_{9.7}$ & $42.7_{12.3}$ & $30.8_{23.1}$ & $53.7_{21.7}$ & $53.8_{17.4}$ & $13.1_{8.9}$ & $33.8_{11.4}$ & $22.2_{5.0}$ & $40.5_{27.7}$ & $37.4_{17.1}$ \\
$\overline{\text{GigaPath}}$~\cite{xu2024gigapath} & $54.3_{18.1}$ & $24.7_{8.7}$ & $42.8_{12.9}$ & $39.7_{18.1}$ & $55.3_{35.3}$ & $45.1_{28.4}$ & $13.1_{9.7}$ & $30.6_{14.5}$ & $22.7_{6.0}$ & $48.5_{24.3}$ & $37.7_{19.7}$ \\
$\overline{\text{Virchow2}}$~\cite{zimmermann2024virchow2} & $59.4_{18.6}$ & $23.5_{11.3}$ & $46.2_{9.2}$ & $38.3_{21.9}$ & $57.6_{22.0}$ & $45.6_{27.5}$ & $15.7_{9.4}$ & $35.8_{12.6}$ & $\mathbf{26.6}_{2.7}$ & $42.4_{28.6}$ & $39.1_{18.3}$ \\
$\overline{\text{CONCH}}$~\cite{lu2024conch} & $70.4_{17.5}$ & $32.0_{11.6}$ & $37.8_{13.9}$ & $44.2_{21.4}$ & $45.5_{26.1}$ & $\mathbf{55.3}_{23.9}$ & $15.5_{9.1}$ & $38.2_{6.1}$ & $21.6_{7.8}$ & $46.9_{24.1}$ & $40.7_{17.6}$ \\
\hline
GigaPath-SE~\cite{xu2024gigapath} & $50.5_{14.3}$ & $27.5_{6.0}$ & $39.9_{8.9}$ & $41.0_{20.9}$ & $55.9_{25.4}$ & $40.4_{18.3}$ & $17.0_{8.2}$ & $28.7_{13.1}$ & $23.7_{7.1}$ & $\mathbf{54.3}_{19.1}$ & $37.9_{15.5}$ \\
CHIEF~\cite{Wang2024Chief} & $63.1_{17.8}$ & $27.5_{7.0}$ & $45.7_{6.0}$ & $35.4_{16.4}$ & $62.7_{20.5}$ & $48.4_{26.7}$ & $14.2_{10.0}$ & $34.1_{11.4}$ & $22.8_{3.8}$ & $36.2_{28.2}$ & $39.0_{16.9}$ \\
\textsc{{Cobra}}$^\dag$-UNI & $66.3_{20.6}$ & $32.1_{12.0}$ & $41.4_{10.4}$ & $37.5_{16.0}$ & $60.7_{29.4}$ & $51.9_{21.8}$ & $13.8_{12.2}$ & $28.7_{16.2}$ & $23.8_{4.5}$ & $44.1_{28.1}$ & $40.0_{18.7}$ \\
\textsc{{Cobra}}$^\dag$-CTP & $65.2_{19.0}$ & $30.3_{5.4}$ & $48.3_{7.7}$ & $39.5_{16.9}$ & $\underline{62.9}_{21.6}$ & $\underline{54.9}_{18.8}$ & $11.4_{8.5}$ & $34.0_{12.7}$ & $21.8_{8.8}$ & $39.9_{25.5}$ & $40.8_{15.9}$ \\
\textsc{{Cobra}}$^\dag$-H0 & $74.3_{14.4}$ & $\mathbf{39.2}_{8.9}$ & $\mathbf{49.7}_{8.2}$ & $41.2_{17.9}$ & $49.4_{35.3}$ & $48.3_{29.9}$ & $9.7_{9.9}$ & $32.9_{19.0}$ & $23.3_{9.9}$ & $41.8_{27.5}$ & $41.0_{20.3}$ \\
\textsc{{Cobra}}$^\dag$-V2 & $74.8_{10.1}$ & $36.0_{6.2}$ & $47.6_{7.5}$ & $\underline{45.5}_{10.8}$ & $\mathbf{64.3}_{21.8}$ & $43.4_{30.4}$ & $\underline{19.5}_{11.0}$ & $32.7_{22.4}$ & $23.9_{7.6}$ & $42.8_{26.0}$ & $43.0_{17.5}$ \\
MADELEINE~\cite{jaume2024madeleine} & $\underline{76.0}_{9.8}$ & $32.3_{6.0}$ & $45.3_{7.7}$ & $43.7_{14.3}$ & $55.0_{19.5}$ & $45.4_{24.5}$ & $\mathbf{21.2}_{3.3}$ & $\underline{38.9}_{4.3}$ & $22.6_{8.6}$ & $\underline{50.5}_{20.7}$ & $\underline{43.1}_{13.8}$ \\
PRISM~\cite{shaikovski2024prism} & $\mathbf{91.7}_{3.4}$ & $\underline{37.1}_{8.9}$ & $\underline{49.5}_{5.8}$ & $\mathbf{54.8}_{18.8}$ & $59.0_{27.1}$ & $41.2_{30.5}$ & $12.4_{10.3}$ & $\mathbf{41.9}_{12.9}$ & $23.4_{6.9}$ & $32.0_{22.5}$ & $\mathbf{44.3}_{17.2}$ \\
\bottomrule
\end{tabular}
}
\end{table*}

\begin{table*}[ht!]
\centering
\caption{\textbf{Few shot performance comparison.} F1 score of models on CPTAC datasets with k=10 positive samples during training on TCGA. $\overline{\text{Overline}}$ indicates mean over patch embeddings, $^\dag$ indicates that embeddings of all four training FMs were used to generate the weighting vector (\cref{eq:cobra-dagger}). \textbf{Bold} indicates the best performance, and $\underline{\text{underline}}$ indicates the second-best performance. The abbreviations are as follows: ST: Subtyping CTP: CTransPath~\cite{wang2022ctp}, H0: H-Optimus-0~\cite{hoptimus0}, V2: Virchow-2~\cite{zimmermann2024virchow2}, GP: GigaPath~\cite{xu2024gigapath}, SE: Slide Encoder.}
\label{tab:lp_f1_10}
\resizebox{\textwidth}{!}{%
\begin{tabular}{l|l|lll|lll|lll|l}
\toprule
F1[\%]-k=10 & LUNG & \multicolumn{3}{c|}{LUAD} & \multicolumn{3}{c|}{BRCA} & \multicolumn{3}{c|}{COAD} & Average \\
Model & ST & STK11 & EGFR & TP53 & ESR1 & PGR & ERBB2 & MSI & BRAF & Side &  \\
\midrule
$\overline{\text{CTransPath}}$~\cite{wang2022ctp} & $58.0_{15.7}$ & $24.9_{9.2}$ & $41.1_{12.9}$ & $24.9_{18.3}$ & $54.4_{22.5}$ & $53.9_{15.2}$ & $12.3_{9.2}$ & $40.6_{4.3}$ & $24.5_{3.5}$ & $39.9_{24.2}$ & $37.5_{15.1}$ \\
$\overline{\text{Virchow}}$~\cite{vorontsov2024virchow} & $58.5_{21.0}$ & $22.2_{11.8}$ & $32.4_{14.4}$ & $28.6_{28.3}$ & $53.9_{23.2}$ & $57.9_{21.2}$ & $13.8_{8.0}$ & $39.6_{1.7}$ & $23.6_{6.0}$ & $48.4_{22.1}$ & $37.9_{17.8}$ \\
$\overline{\text{H-Optimus}}$~\cite{hoptimus0} & $57.7_{21.8}$ & $28.7_{11.6}$ & $39.4_{15.6}$ & $28.9_{21.2}$ & $\underline{70.4}_{23.3}$ & $49.6_{30.2}$ & $10.7_{10.1}$ & $40.4_{9.3}$ & $21.9_{7.4}$ & $41.6_{29.2}$ & $38.9_{19.6}$ \\
$\overline{\text{UNI}}$~\cite{chen2024uni} & $55.8_{21.8}$ & $23.0_{9.9}$ & $42.7_{14.4}$ & $33.1_{22.3}$ & $62.3_{22.4}$ & $\mathbf{58.4}_{18.8}$ & $13.2_{12.3}$ & $42.1_{6.6}$ & $22.8_{7.6}$ & $40.3_{24.2}$ & $39.4_{17.2}$ \\
$\overline{\text{GigaPath}}$~\cite{xu2024gigapath} & $59.3_{15.9}$ & $19.9_{13.7}$ & $47.7_{8.2}$ & $38.4_{14.4}$ & $65.0_{22.3}$ & $54.5_{26.7}$ & $13.9_{10.6}$ & $40.2_{12.2}$ & $25.9_{7.0}$ & $\underline{53.9}_{22.9}$ & $41.9_{16.6}$ \\
$\overline{\text{Virchow2}}$~\cite{zimmermann2024virchow2} & $62.4_{15.1}$ & $25.9_{10.4}$ & $45.6_{10.5}$ & $42.0_{14.7}$ & $\mathbf{72.9}_{14.0}$ & $52.2_{23.5}$ & $18.4_{7.1}$ & $41.4_{8.6}$ & $22.3_{11.9}$ & $37.9_{27.2}$ & $42.1_{15.5}$ \\
$\overline{\text{CONCH}}$~\cite{lu2024conch} & $72.8_{15.3}$ & $31.3_{4.0}$ & $42.5_{8.9}$ & $44.1_{16.0}$ & $51.7_{26.9}$ & $49.0_{23.5}$ & $17.8_{7.4}$ & $40.0_{4.3}$ & $26.6_{2.5}$ & $48.4_{24.1}$ & $42.4_{15.9}$ \\
\hline
GigaPath-SE~\cite{xu2024gigapath} & $60.8_{9.1}$ & $25.4_{9.1}$ & $44.2_{9.3}$ & $46.1_{15.2}$ & $57.6_{22.6}$ & $47.7_{20.8}$ & $15.5_{7.9}$ & $34.0_{10.9}$ & $21.7_{8.1}$ & $\mathbf{57.3}_{10.0}$ & $41.0_{13.3}$ \\
CHIEF~\cite{Wang2024Chief} & $67.8_{13.1}$ & $32.7_{8.8}$ & $44.3_{12.4}$ & $35.7_{16.6}$ & $64.5_{18.5}$ & $48.7_{23.4}$ & $14.9_{13.6}$ & $41.5_{6.7}$ & $24.3_{5.2}$ & $40.1_{23.5}$ & $41.5_{15.4}$ \\
MADELEINE~\cite{jaume2024madeleine} & $\underline{80.2}_{5.2}$ & $31.9_{7.4}$ & $44.2_{9.4}$ & $46.5_{13.9}$ & $56.9_{24.9}$ & $29.9_{22.7}$ & $\mathbf{22.7}_{2.5}$ & $40.3_{3.6}$ & $24.7_{2.5}$ & $47.7_{21.5}$ & $42.5_{14.1}$ \\
\textsc{{Cobra}}$^\dag$-H0 & $78.9_{8.3}$ & $\mathbf{42.9}_{7.0}$ & $48.6_{8.6}$ & $39.3_{20.3}$ & $54.8_{32.3}$ & $49.4_{26.9}$ & $11.8_{10.3}$ & $41.7_{14.9}$ & $22.8_{7.7}$ & $45.0_{25.4}$ & $43.5_{18.4}$ \\
\textsc{{Cobra}}$^\dag$-CTP & $72.5_{13.3}$ & $34.0_{5.4}$ & $47.4_{11.3}$ & $38.8_{18.9}$ & $67.0_{16.2}$ & $\underline{58.1}_{22.6}$ & $13.0_{11.2}$ & $39.3_{15.3}$ & $25.3_{6.8}$ & $40.4_{21.2}$ & $43.6_{15.2}$ \\
\textsc{{Cobra}}$^\dag$-UNI & $75.5_{16.7}$ & $38.5_{9.8}$ & $49.9_{8.7}$ & $40.2_{15.0}$ & $59.6_{29.2}$ & $54.1_{24.6}$ & $12.8_{10.7}$ & $42.1_{16.5}$ & $\mathbf{28.1}_{5.5}$ & $41.1_{19.9}$ & $44.2_{17.2}$ \\
\textsc{{Cobra}}$^\dag$-V2 & $80.0_{7.9}$ & $37.2_{8.7}$ & $\underline{50.3}_{11.1}$ & $\underline{48.2}_{11.9}$ & $67.1_{18.0}$ & $45.4_{29.2}$ & $\underline{19.8}_{8.7}$ & $\underline{44.8}_{17.4}$ & $\underline{26.8}_{13.7}$ & $41.3_{25.9}$ & $\underline{46.1}_{16.8}$ \\
PRISM~\cite{shaikovski2024prism} & $\mathbf{92.8}_{3.4}$ & $\underline{40.1}_{8.4}$ & $\mathbf{52.4}_{4.7}$ & $\mathbf{56.3}_{9.6}$ & $69.6_{17.5}$ & $52.6_{23.1}$ & $17.2_{10.3}$ & $\mathbf{48.0}_{6.6}$ & $24.3_{6.9}$ & $41.6_{19.7}$ & $\mathbf{49.5}_{12.7}$ \\
\bottomrule
\end{tabular}
}
\end{table*}

\begin{table*}[ht!]
\centering
\caption{\textbf{Few shot performance comparison.} F1 score of models on CPTAC datasets with k=25 positive samples during training on TCGA. $\overline{\text{Overline}}$ indicates mean over patch embeddings, $^\dag$ indicates that embeddings of all four training FMs were used to generate the weighting vector (\cref{eq:cobra-dagger}). \textbf{Bold} indicates the best performance, and $\underline{\text{underline}}$ indicates the second-best performance. The abbreviations are as follows: ST: Subtyping CTP: CTransPath~\cite{wang2022ctp}, H0: H-Optimus-0~\cite{hoptimus0}, V2: Virchow-2~\cite{zimmermann2024virchow2}, GP: GigaPath~\cite{xu2024gigapath}, SE: Slide Encoder.}
\label{tab:lp_f1_25}
\resizebox{\textwidth}{!}{%
\begin{tabular}{l|l|lll|lll|lll|l}
\toprule
F1[\%]-k=25 & LUNG & \multicolumn{3}{c|}{LUAD} & \multicolumn{3}{c|}{BRCA} & \multicolumn{3}{c|}{COAD} & Average \\
Model & ST & STK11 & EGFR & TP53 & ESR1 & PGR & ERBB2 & MSI & BRAF & Side &  \\
\midrule
$\overline{\text{Virchow}}$~\cite{vorontsov2024virchow} & $62.8_{16.5}$ & $26.7_{13.0}$ & $46.7_{13.4}$ & $38.2_{27.2}$ & $40.4_{25.5}$ & $50.2_{24.6}$ & $17.0_{5.0}$ & $44.1_{7.8}$ & $26.0_{3.0}$ & $41.8_{22.3}$ & $39.4_{17.9}$ \\
$\overline{\text{CTransPath}}$~\cite{wang2022ctp} & $64.1_{12.3}$ & $25.5_{11.6}$ & $47.3_{9.5}$ & $34.7_{18.7}$ & $60.0_{21.2}$ & $49.1_{21.0}$ & $15.6_{8.8}$ & $41.2_{5.7}$ & $25.7_{1.4}$ & $42.3_{25.9}$ & $40.5_{15.5}$ \\
$\overline{\text{H-Optimus}}$~\cite{hoptimus0} & $65.4_{14.7}$ & $27.4_{14.3}$ & $46.6_{17.7}$ & $42.6_{23.9}$ & $66.2_{26.6}$ & $49.3_{29.6}$ & $13.5_{9.6}$ & $43.7_{10.8}$ & $27.1_{3.0}$ & $25.1_{28.7}$ & $40.7_{19.8}$ \\
$\overline{\text{GigaPath}}$~\cite{xu2024gigapath} & $64.4_{15.0}$ & $22.1_{14.3}$ & $54.1_{7.6}$ & $36.4_{20.0}$ & $67.9_{22.9}$ & $49.5_{29.6}$ & $19.5_{8.2}$ & $39.9_{11.6}$ & $27.0_{4.7}$ & $37.6_{23.0}$ & $41.8_{17.4}$ \\
$\overline{\text{CONCH}}$~\cite{lu2024conch} & $81.8_{8.5}$ & $37.9_{9.4}$ & $44.7_{11.6}$ & $42.6_{15.4}$ & $50.3_{27.6}$ & $52.4_{19.5}$ & $18.9_{7.2}$ & $49.9_{9.1}$ & $29.0_{10.0}$ & $\underline{46.8}_{24.9}$ & $45.4_{15.9}$ \\
$\overline{\text{UNI}}$~\cite{chen2024uni} & $64.5_{15.4}$ & $27.7_{10.1}$ & $49.2_{10.4}$ & $45.6_{25.1}$ & $68.3_{22.9}$ & $\underline{62.0}_{22.3}$ & $18.0_{6.7}$ & $47.4_{10.4}$ & $26.9_{2.2}$ & $45.2_{18.9}$ & $45.5_{16.2}$ \\
$\overline{\text{Virchow2}}$~\cite{zimmermann2024virchow2} & $71.4_{9.2}$ & $30.7_{11.5}$ & $52.6_{5.5}$ & $46.1_{18.0}$ & $71.6_{17.8}$ & $51.5_{22.8}$ & $16.9_{9.5}$ & $\underline{54.5}_{19.0}$ & $29.3_{6.5}$ & $30.6_{25.4}$ & $45.5_{16.0}$ \\
\hline
GigaPath-SE~\cite{xu2024gigapath} & $64.4_{9.7}$ & $21.1_{11.8}$ & $42.4_{9.9}$ & $44.5_{16.7}$ & $48.9_{25.2}$ & $43.1_{24.4}$ & $20.2_{7.6}$ & $36.3_{6.9}$ & $25.0_{5.3}$ & $\mathbf{50.9}_{16.3}$ & $39.7_{15.0}$ \\
CHIEF~\cite{Wang2024Chief} & $76.1_{9.8}$ & $33.9_{6.1}$ & $52.8_{7.4}$ & $45.1_{17.0}$ & $68.2_{18.4}$ & $53.5_{27.2}$ & $18.1_{8.1}$ & $42.7_{10.2}$ & $24.5_{1.7}$ & $34.4_{25.0}$ & $44.9_{15.3}$ \\
MADELEINE~\cite{jaume2024madeleine} & $85.1_{5.2}$ & $37.9_{7.3}$ & $45.0_{10.7}$ & $50.3_{13.5}$ & $57.2_{24.2}$ & $40.0_{21.2}$ & $\mathbf{23.2}_{1.5}$ & $47.5_{7.4}$ & $24.8_{2.7}$ & $42.5_{22.8}$ & $45.4_{14.1}$ \\
\textsc{{Cobra}}$^\dag$-CTP & $79.6_{7.3}$ & $33.9_{6.7}$ & $54.2_{4.6}$ & $48.4_{15.3}$ & $\mathbf{75.1}_{8.1}$ & $58.4_{23.0}$ & $14.1_{6.6}$ & $47.9_{13.4}$ & $26.0_{2.3}$ & $33.7_{26.9}$ & $47.1_{13.8}$ \\
\textsc{{Cobra}}$^\dag$-H0 & $\underline{85.2}_{6.7}$ & $\underline{42.9}_{6.0}$ & $\mathbf{58.4}_{13.1}$ & $50.2_{13.9}$ & $65.8_{30.6}$ & $49.8_{28.4}$ & $13.2_{9.9}$ & $52.5_{12.5}$ & $31.0_{5.4}$ & $30.9_{26.0}$ & $48.0_{17.7}$ \\
\textsc{{Cobra}}$^\dag$-UNI & $83.9_{6.6}$ & $40.4_{7.5}$ & $56.9_{7.9}$ & $53.5_{12.9}$ & $\underline{74.2}_{13.5}$ & $\mathbf{62.2}_{23.8}$ & $18.0_{9.0}$ & $50.2_{17.1}$ & $\underline{33.3}_{6.7}$ & $24.5_{20.9}$ & $49.7_{13.9}$ \\
\textsc{{Cobra}}$^\dag$-V2 & $84.9_{6.0}$ & $40.7_{5.3}$ & $56.9_{3.6}$ & $\underline{53.8}_{14.5}$ & $73.2_{15.1}$ & $55.7_{22.7}$ & $18.5_{7.7}$ & $\mathbf{58.0}_{24.2}$ & $\mathbf{38.6}_{8.5}$ & $24.4_{21.0}$ & $\underline{50.5}_{14.8}$ \\
PRISM~\cite{shaikovski2024prism} & $\mathbf{93.4}_{1.8}$ & $\mathbf{49.0}_{4.9}$ & $\underline{57.6}_{2.9}$ & $\mathbf{63.8}_{6.6}$ & $66.0_{17.8}$ & $55.7_{22.8}$ & $\underline{21.1}_{6.3}$ & $49.2_{4.1}$ & $29.5_{4.4}$ & $45.6_{17.7}$ & $\mathbf{53.1}_{11.4}$ \\
\bottomrule
\end{tabular}
}
\end{table*}

\begin{table*}[ht!]
\centering
\caption{\textbf{Few shot performance comparison.} Balanced accuracy score of models on CPTAC datasets with k=5 positive samples during training on TCGA. $\overline{\text{Overline}}$ indicates mean over patch embeddings, $^\dag$ indicates that embeddings of all four training FMs were used to generate the weighting vector (\cref{eq:cobra-dagger}). \textbf{Bold} indicates the best performance, and $\underline{\text{underline}}$ indicates the second-best performance. The abbreviations are as follows: ST: Subtyping CTP: CTransPath~\cite{wang2022ctp}, H0: H-Optimus-0~\cite{hoptimus0}, V2: Virchow-2~\cite{zimmermann2024virchow2}, GP: GigaPath~\cite{xu2024gigapath}, SE: Slide Encoder.}
\label{tab:lp_accuracy_5}
\resizebox{\textwidth}{!}{%
\begin{tabular}{l|l|lll|lll|lll|l}
\toprule
Balanced Acc.[\%]-k=5 & LUNG & \multicolumn{3}{c|}{LUAD} & \multicolumn{3}{c|}{BRCA} & \multicolumn{3}{c|}{COAD} & Average \\
Model & ST & STK11 & EGFR & TP53 & ESR1 & PGR & ERBB2 & MSI & BRAF & Side &  \\
\midrule
$\overline{\text{Virchow}}$~\cite{vorontsov2024virchow} & $55.2_{8.3}$ & $49.5_{5.3}$ & $51.7_{2.1}$ & $49.3_{3.5}$ & $53.5_{6.3}$ & $48.6_{3.8}$ & $46.5_{3.7}$ & $52.6_{2.1}$ & $51.0_{3.7}$ & $50.2_{2.3}$ & $50.8_{4.5}$ \\
$\overline{\text{CTransPath}}$~\cite{wang2022ctp} & $58.7_{9.3}$ & $52.5_{4.3}$ & $54.1_{6.8}$ & $50.6_{4.9}$ & $53.5_{7.7}$ & $54.3_{3.9}$ & $51.0_{1.8}$ & $51.3_{2.9}$ & $48.4_{2.6}$ & $\underline{51.3}_{1.5}$ & $52.6_{5.2}$ \\
$\overline{\text{UNI}}$~\cite{chen2024uni} & $58.9_{7.7}$ & $53.5_{3.4}$ & $52.8_{7.4}$ & $52.1_{5.3}$ & $53.4_{6.1}$ & $53.3_{4.1}$ & $51.9_{2.1}$ & $54.9_{7.7}$ & $48.9_{4.0}$ & $51.1_{2.6}$ & $53.1_{5.4}$ \\
$\overline{\text{GigaPath}}$~\cite{xu2024gigapath} & $62.2_{9.7}$ & $52.7_{5.3}$ & $57.8_{7.0}$ & $52.8_{6.7}$ & $51.5_{4.0}$ & $51.8_{3.6}$ & $50.9_{4.7}$ & $53.7_{4.9}$ & $51.0_{6.0}$ & $51.2_{3.0}$ & $53.6_{5.8}$ \\
$\overline{\text{H-Optimus}}$~\cite{hoptimus0} & $63.0_{9.9}$ & $56.0_{5.7}$ & $56.8_{5.5}$ & $50.9_{5.4}$ & $54.2_{5.5}$ & $49.7_{3.6}$ & $50.2_{3.2}$ & $54.8_{7.6}$ & $50.7_{5.8}$ & $\underline{51.3}_{2.6}$ & $53.8_{5.8}$ \\
$\overline{\text{Virchow2}}$~\cite{zimmermann2024virchow2} & $64.3_{9.9}$ & $53.7_{4.4}$ & $58.8_{7.3}$ & $51.6_{6.9}$ & $56.0_{8.0}$ & $54.7_{3.3}$ & $51.2_{4.4}$ & $55.1_{6.5}$ & $\underline{54.3}_{4.6}$ & $50.5_{0.9}$ & $55.0_{6.1}$ \\
$\overline{\text{CONCH}}$~\cite{lu2024conch} & $74.9_{10.9}$ & $58.3_{8.2}$ & $52.9_{3.6}$ & $53.4_{6.2}$ & $54.3_{7.0}$ & $\mathbf{58.0}_{7.3}$ & $50.2_{6.7}$ & $54.3_{6.4}$ & $50.7_{4.6}$ & $50.0_{3.2}$ & $55.7_{6.8}$ \\
\hline
GigaPath-SE~\cite{xu2024gigapath} & $59.2_{4.3}$ & $54.2_{5.4}$ & $54.9_{3.8}$ & $52.0_{6.8}$ & $53.7_{5.2}$ & $51.4_{4.9}$ & $50.7_{7.7}$ & $50.0_{4.7}$ & $52.0_{6.3}$ & $49.2_{4.3}$ & $52.7_{5.5}$ \\
CHIEF~\cite{Wang2024Chief} & $65.4_{10.7}$ & $54.4_{5.0}$ & $55.8_{4.7}$ & $51.7_{5.5}$ & $54.5_{6.6}$ & $57.1_{5.4}$ & $51.7_{3.9}$ & $52.1_{2.5}$ & $48.9_{4.2}$ & $49.5_{2.7}$ & $54.1_{5.6}$ \\
\textsc{{Cobra}}$^\dag$-CTP & $67.7_{10.4}$ & $56.6_{5.7}$ & $57.8_{6.5}$ & $55.4_{4.0}$ & $55.1_{5.4}$ & $53.5_{4.2}$ & $50.9_{2.3}$ & $54.7_{5.4}$ & $50.4_{6.2}$ & $48.5_{2.2}$ & $55.1_{5.7}$ \\
\textsc{{Cobra}}$^\dag$-UNI & $73.1_{10.6}$ & $61.6_{7.2}$ & $56.3_{6.0}$ & $54.8_{4.7}$ & $54.2_{7.7}$ & $56.4_{5.7}$ & $\underline{53.0}_{7.0}$ & $52.6_{4.9}$ & $51.8_{4.8}$ & $50.8_{4.9}$ & $56.5_{6.6}$ \\
MADELEINE~\cite{jaume2024madeleine} & $\underline{78.1}_{6.9}$ & $58.2_{6.6}$ & $57.2_{6.1}$ & $52.7_{5.2}$ & $57.4_{5.9}$ & $55.9_{5.9}$ & $52.2_{3.7}$ & $54.5_{6.1}$ & $51.8_{7.4}$ & $\mathbf{51.8}_{2.6}$ & $57.0_{5.8}$ \\
\textsc{{Cobra}}$^\dag$-H0 & $77.2_{9.2}$ & $\mathbf{65.3}_{7.6}$ & $\mathbf{62.2}_{6.1}$ & $\underline{56.4}_{5.8}$ & $54.6_{6.9}$ & $52.7_{4.4}$ & $51.4_{4.4}$ & $54.5_{8.3}$ & $52.1_{8.2}$ & $50.3_{2.0}$ & $57.7_{6.6}$ \\
\textsc{{Cobra}}$^\dag$-V2 & $77.2_{7.5}$ & $62.6_{6.8}$ & $\underline{59.3}_{5.4}$ & $55.5_{4.2}$ & $\underline{58.0}_{6.4}$ & $55.4_{4.7}$ & $\mathbf{56.0}_{5.4}$ & $\underline{56.6}_{11.8}$ & $54.2_{6.7}$ & $50.0_{1.4}$ & $\underline{58.5}_{6.5}$ \\
PRISM~\cite{shaikovski2024prism} & $\mathbf{91.7}_{3.5}$ & $\underline{64.9}_{8.5}$ & $57.2_{7.5}$ & $\mathbf{60.3}_{6.2}$ & $\mathbf{60.8}_{7.2}$ & $\underline{57.7}_{7.2}$ & $50.5_{3.8}$ & $\mathbf{62.0}_{7.7}$ & $\mathbf{55.5}_{3.6}$ & $50.2_{2.4}$ & $\mathbf{61.1}_{6.1}$ \\
\bottomrule
\end{tabular}
}
\end{table*}

\begin{table*}[ht!]
\centering
\caption{\textbf{Few shot performance comparison.} Balanced accuracy score of models on CPTAC datasets with k=10 positive samples during training on TCGA. $\overline{\text{Overline}}$ indicates mean over patch embeddings, $^\dag$ indicates that embeddings of all four training FMs were used to generate the weighting vector (\cref{eq:cobra-dagger}). \textbf{Bold} indicates the best performance, and $\underline{\text{underline}}$ indicates the second-best performance. The abbreviations are as follows: ST: Subtyping CTP: CTransPath~\cite{wang2022ctp}, H0: H-Optimus-0~\cite{hoptimus0}, V2: Virchow-2~\cite{zimmermann2024virchow2}, GP: GigaPath~\cite{xu2024gigapath}, SE: Slide Encoder.}
\label{tab:lp_accuracy_10}
\resizebox{\textwidth}{!}{%
\begin{tabular}{l|l|lll|lll|lll|l}
\toprule
Balanced Acc.[\%]-k=10 & LUNG & \multicolumn{3}{c|}{LUAD} & \multicolumn{3}{c|}{BRCA} & \multicolumn{3}{c|}{COAD} & Average \\
Model & ST & STK11 & EGFR & TP53 & ESR1 & PGR & ERBB2 & MSI & BRAF & Side &  \\
\midrule
$\overline{\text{Virchow}}$~\cite{vorontsov2024virchow} & $59.9_{7.2}$ & $51.7_{3.9}$ & $51.1_{3.8}$ & $51.4_{5.2}$ & $53.3_{7.2}$ & $54.6_{4.6}$ & $48.2_{3.1}$ & $56.0_{3.5}$ & $52.3_{4.6}$ & $49.9_{3.2}$ & $52.8_{4.8}$ \\
$\overline{\text{CTransPath}}$~\cite{wang2022ctp} & $58.5_{9.8}$ & $53.1_{4.1}$ & $53.0_{6.2}$ & $49.9_{4.8}$ & $56.7_{3.3}$ & $56.4_{4.6}$ & $49.6_{3.2}$ & $56.8_{6.6}$ & $50.3_{5.9}$ & $50.1_{2.0}$ & $53.4_{5.5}$ \\
$\overline{\text{H-Optimus}}$~\cite{hoptimus0} & $64.7_{10.8}$ & $58.2_{7.2}$ & $55.2_{3.9}$ & $52.2_{4.9}$ & $57.4_{8.2}$ & $52.8_{3.8}$ & $51.1_{3.8}$ & $57.0_{9.7}$ & $50.3_{1.8}$ & $51.2_{2.2}$ & $55.0_{6.4}$ \\
$\overline{\text{UNI}}$~\cite{chen2024uni} & $64.9_{9.9}$ & $53.6_{4.7}$ & $57.0_{6.3}$ & $55.6_{4.8}$ & $58.2_{7.4}$ & $58.6_{6.2}$ & $51.9_{8.4}$ & $57.7_{8.1}$ & $51.3_{2.6}$ & $\mathbf{53.3}_{4.8}$ & $56.2_{6.6}$ \\
$\overline{\text{GigaPath}}$~\cite{xu2024gigapath} & $67.1_{8.1}$ & $52.6_{4.7}$ & $58.3_{5.2}$ & $53.7_{7.0}$ & $58.6_{8.5}$ & $54.1_{4.8}$ & $52.7_{4.9}$ & $60.2_{8.8}$ & $54.9_{4.9}$ & $\underline{52.3}_{3.6}$ & $56.5_{6.3}$ \\
$\overline{\text{Virchow2}}$~\cite{zimmermann2024virchow2} & $67.1_{7.7}$ & $55.4_{5.6}$ & $57.5_{5.3}$ & $55.4_{6.2}$ & $\underline{62.8}_{10.2}$ & $57.2_{4.5}$ & $53.3_{3.6}$ & $60.0_{7.8}$ & $53.7_{5.1}$ & $51.4_{1.8}$ & $57.4_{6.2}$ \\
$\overline{\text{CONCH}}$~\cite{lu2024conch} & $75.9_{8.7}$ & $56.0_{5.8}$ & $54.1_{4.8}$ & $56.0_{5.6}$ & $61.3_{6.0}$ & $58.9_{7.2}$ & $52.3_{4.5}$ & $57.1_{5.8}$ & $54.3_{4.5}$ & $49.9_{4.4}$ & $57.6_{5.9}$ \\
\hline
GigaPath-SE~\cite{xu2024gigapath} & $64.6_{5.9}$ & $54.6_{6.1}$ & $57.8_{4.6}$ & $55.2_{4.3}$ & $56.8_{3.6}$ & $53.1_{6.4}$ & $50.2_{5.5}$ & $52.5_{8.6}$ & $51.9_{4.2}$ & $51.4_{4.8}$ & $54.8_{5.6}$ \\
CHIEF~\cite{Wang2024Chief} & $68.9_{10.9}$ & $59.0_{6.9}$ & $56.1_{6.1}$ & $54.2_{5.2}$ & $62.3_{5.3}$ & $\underline{59.7}_{5.4}$ & $52.3_{7.5}$ & $58.1_{8.8}$ & $50.4_{7.7}$ & $50.2_{4.5}$ & $57.1_{7.1}$ \\
\textsc{{Cobra}}$^\dag$-CTP & $74.3_{9.7}$ & $59.9_{3.8}$ & $58.4_{7.2}$ & $56.1_{4.4}$ & $58.2_{6.0}$ & $55.4_{4.6}$ & $52.2_{4.8}$ & $58.4_{9.8}$ & $52.0_{9.8}$ & $49.5_{1.9}$ & $57.4_{6.7}$ \\
MADELEINE~\cite{jaume2024madeleine} & $80.6_{5.9}$ & $59.2_{6.2}$ & $57.4_{5.8}$ & $57.1_{6.2}$ & $60.2_{7.7}$ & $55.2_{5.0}$ & $54.3_{4.7}$ & $56.6_{5.4}$ & $50.6_{4.4}$ & $51.0_{3.5}$ & $58.2_{5.6}$ \\
\textsc{{Cobra}}$^\dag$-H0 & $80.9_{5.6}$ & $\mathbf{69.8}_{7.2}$ & $\underline{61.5}_{5.1}$ & $58.8_{6.1}$ & $55.7_{6.1}$ & $51.4_{3.3}$ & $51.4_{4.1}$ & $60.9_{10.1}$ & $52.1_{2.3}$ & $50.7_{2.4}$ & $59.3_{5.7}$ \\
\textsc{{Cobra}}$^\dag$-UNI & $79.7_{8.7}$ & $65.6_{7.7}$ & $61.0_{6.5}$ & $58.9_{4.4}$ & $58.6_{6.2}$ & $55.3_{3.8}$ & $51.6_{6.5}$ & $62.1_{10.9}$ & $\underline{57.0}_{4.9}$ & $51.1_{3.2}$ & $60.1_{6.7}$ \\
\textsc{{Cobra}}$^\dag$-V2 & $\underline{81.6}_{6.2}$ & $64.0_{7.2}$ & $\mathbf{61.9}_{8.2}$ & $\underline{59.7}_{5.2}$ & $59.2_{5.8}$ & $56.5_{6.1}$ & $\mathbf{56.2}_{4.5}$ & $\underline{64.6}_{12.1}$ & $\mathbf{59.2}_{9.2}$ & $51.3_{2.4}$ & $\underline{61.4}_{7.2}$ \\
PRISM~\cite{shaikovski2024prism} & $\mathbf{92.9}_{3.3}$ & $\underline{67.5}_{8.6}$ & $60.4_{6.2}$ & $\mathbf{62.0}_{4.4}$ & $\mathbf{64.5}_{7.8}$ & $\mathbf{62.8}_{6.8}$ & $\underline{54.4}_{6.6}$ & $\mathbf{65.7}_{6.9}$ & $55.7_{4.6}$ & $50.4_{2.9}$ & $\mathbf{63.6}_{6.1}$ \\
\bottomrule
\end{tabular}
}
\end{table*}

\begin{table*}[ht!]
\centering
\caption{\textbf{Few shot performance comparison.} Balanced accuracy score of models on CPTAC datasets with k=25 positive samples during training on TCGA. $\overline{\text{Overline}}$ indicates mean over patch embeddings, $^\dag$ indicates that embeddings of all four training FMs were used to generate the weighting vector (\cref{eq:cobra-dagger}). \textbf{Bold} indicates the best performance, and $\underline{\text{underline}}$ indicates the second-best performance. The abbreviations are as follows: ST: Subtyping CTP: CTransPath~\cite{wang2022ctp}, H0: H-Optimus-0~\cite{hoptimus0}, V2: Virchow-2~\cite{zimmermann2024virchow2}, GP: GigaPath~\cite{xu2024gigapath}, SE: Slide Encoder.}
\label{tab:lp_accuracy_25}
\resizebox{\textwidth}{!}{%
\begin{tabular}{l|l|lll|lll|lll|l}
\toprule
Balanced Acc.[\%]-k=25 & LUNG & \multicolumn{3}{c|}{LUAD} & \multicolumn{3}{c|}{BRCA} & \multicolumn{3}{c|}{COAD} & Average \\
Model & ST & STK11 & EGFR & TP53 & ESR1 & PGR & ERBB2 & MSI & BRAF & Side &  \\
\midrule
$\overline{\text{Virchow}}$~\cite{vorontsov2024virchow} & $64.8_{7.0}$ & $55.0_{8.0}$ & $56.6_{6.6}$ & $54.5_{6.9}$ & $54.1_{7.4}$ & $53.8_{5.2}$ & $49.7_{5.1}$ & $60.3_{7.9}$ & $54.3_{4.3}$ & $52.0_{2.9}$ & $55.5_{6.3}$ \\
$\overline{\text{CTransPath}}$~\cite{wang2022ctp} & $64.2_{9.7}$ & $54.6_{6.2}$ & $56.1_{6.4}$ & $56.6_{5.3}$ & $59.9_{6.0}$ & $55.9_{3.7}$ & $50.4_{6.0}$ & $57.9_{7.6}$ & $52.2_{3.2}$ & $50.0_{2.3}$ & $55.8_{6.0}$ \\
$\overline{\text{H-Optimus}}$~\cite{hoptimus0} & $71.4_{7.3}$ & $59.5_{9.0}$ & $62.4_{5.3}$ & $57.7_{6.1}$ & $54.0_{6.9}$ & $51.1_{5.0}$ & $52.2_{2.8}$ & $61.8_{9.4}$ & $54.9_{5.9}$ & $50.5_{1.0}$ & $57.5_{6.4}$ \\
$\overline{\text{GigaPath}}$~\cite{xu2024gigapath} & $71.3_{9.7}$ & $55.6_{7.0}$ & $64.8_{4.7}$ & $57.2_{6.7}$ & $59.6_{7.7}$ & $54.0_{4.2}$ & $54.5_{6.3}$ & $61.8_{8.3}$ & $55.6_{7.4}$ & $51.6_{1.8}$ & $58.6_{6.7}$ \\
$\overline{\text{UNI}}$~\cite{chen2024uni} & $70.4_{8.8}$ & $56.6_{8.1}$ & $60.6_{5.6}$ & $58.9_{6.5}$ & $61.1_{8.9}$ & $58.1_{5.3}$ & $51.9_{3.7}$ & $65.1_{10.3}$ & $54.9_{4.6}$ & $\mathbf{53.6}_{2.0}$ & $59.1_{6.8}$ \\
$\overline{\text{Virchow2}}$~\cite{zimmermann2024virchow2} & $73.9_{6.3}$ & $60.1_{8.6}$ & $63.4_{3.2}$ & $58.5_{7.0}$ & $63.0_{10.0}$ & $55.6_{4.9}$ & $53.1_{4.3}$ & $\underline{72.3}_{12.5}$ & $58.5_{8.0}$ & $\underline{52.8}_{2.6}$ & $61.1_{7.4}$ \\
$\overline{\text{CONCH}}$~\cite{lu2024conch} & $83.1_{6.6}$ & $62.8_{9.3}$ & $57.4_{4.9}$ & $58.1_{6.0}$ & $62.3_{8.0}$ & $58.3_{6.1}$ & $52.5_{4.6}$ & $67.0_{8.3}$ & $58.6_{9.6}$ & $52.0_{3.6}$ & $61.2_{7.0}$ \\
\hline
GigaPath-SE~\cite{xu2024gigapath} & $68.6_{6.1}$ & $53.5_{4.9}$ & $56.9_{5.4}$ & $55.5_{5.5}$ & $54.4_{5.3}$ & $53.6_{6.5}$ & $53.6_{6.9}$ & $57.3_{4.5}$ & $52.7_{5.3}$ & $49.8_{4.5}$ & $55.6_{5.5}$ \\
CHIEF~\cite{Wang2024Chief} & $77.0_{8.9}$ & $60.4_{5.5}$ & $62.1_{5.7}$ & $60.1_{6.6}$ & $63.3_{6.8}$ & $\underline{60.3}_{6.6}$ & $53.1_{5.6}$ & $61.5_{7.9}$ & $50.0_{3.4}$ & $49.5_{2.2}$ & $59.7_{6.2}$ \\
\textsc{{Cobra}}$^\dag$-CTP & $80.7_{6.6}$ & $60.5_{6.9}$ & $63.5_{4.2}$ & $61.6_{6.0}$ & $62.8_{6.5}$ & $56.4_{3.9}$ & $50.0_{3.8}$ & $67.1_{8.0}$ & $53.2_{4.3}$ & $49.4_{2.5}$ & $60.5_{5.5}$ \\
MADELEINE~\cite{jaume2024madeleine} & $85.6_{4.7}$ & $64.1_{6.0}$ & $59.3_{5.8}$ & $61.4_{5.1}$ & $64.7_{7.2}$ & $57.1_{4.1}$ & $\mathbf{55.7}_{3.2}$ & $64.6_{6.9}$ & $50.9_{4.8}$ & $50.5_{2.7}$ & $61.4_{5.2}$ \\
\textsc{{Cobra}}$^\dag$-H0 & $\underline{86.5}_{4.7}$ & $\underline{69.4}_{6.1}$ & $\mathbf{70.2}_{6.1}$ & $62.1_{6.1}$ & $58.3_{7.6}$ & $51.5_{5.5}$ & $51.5_{4.8}$ & $68.9_{10.9}$ & $60.8_{8.6}$ & $50.9_{2.3}$ & $63.0_{6.7}$ \\
\textsc{{Cobra}}$^\dag$-UNI & $85.5_{5.3}$ & $67.0_{7.1}$ & $\underline{67.0}_{6.4}$ & $\underline{63.8}_{5.0}$ & $\underline{67.3}_{6.8}$ & $58.7_{5.4}$ & $53.9_{4.3}$ & $69.6_{10.9}$ & $\underline{64.6}_{8.7}$ & $50.3_{3.1}$ & $64.8_{6.7}$ \\
\textsc{{Cobra}}$^\dag$-V2 & $86.0_{4.8}$ & $67.5_{5.4}$ & $66.8_{3.6}$ & $63.6_{6.8}$ & $64.1_{10.1}$ & $58.6_{7.8}$ & $54.8_{2.6}$ & $\mathbf{76.3}_{13.2}$ & $\mathbf{67.4}_{8.4}$ & $50.7_{1.7}$ & $\underline{65.6}_{7.3}$ \\
PRISM~\cite{shaikovski2024prism} & $\mathbf{93.4}_{1.8}$ & $\mathbf{74.7}_{4.9}$ & $66.1_{3.5}$ & $\mathbf{65.7}_{2.8}$ & $\mathbf{68.0}_{8.2}$ & $\mathbf{60.6}_{5.4}$ & $\underline{55.0}_{6.2}$ & $67.5_{3.1}$ & $58.8_{4.6}$ & $50.1_{3.0}$ & $\mathbf{66.0}_{4.7}$ \\
\bottomrule
\end{tabular}
}
\end{table*}


\section*{Limitations}
While \ac{cobra} has demonstrated promising results, several limitations exist that warrant further investigation. First, the pretraining process involves a limited number of tissue types, which may restrict its generalizability to other histopathological contexts. Second, the diversity of downstream tasks and evaluation datasets is currently narrow, potentially limiting the framework's applicability across varied clinical scenarios. Third, the self-supervised learning (SSL) strategy exclusively employs a contrastive loss function based on MoCo-v3, leaving room for exploration of alternative or complementary loss functions that could enhance representation quality. Finally, the resulting patient-level embedding is formulated as a linear combination of patch embeddings, which may not fully capture the complex, non-linear relationships inherent in histopathological data. Addressing these limitations will be a focus of future research to improve the robustness and versatility of the proposed framework.

\section*{Competing Interests}
JNK declares consulting services for Bioptimus, France; Panakeia, UK; AstraZeneca, UK; and MultiplexDx, Slovakia. Furthermore, he holds shares in StratifAI, Germany, Synagen, Germany, Ignition Lab, Germany; has received an institutional research grant by GSK; and has received honoraria by AstraZeneca, Bayer, Daiichi Sankyo, Eisai, Janssen, Merck, MSD, BMS, Roche, Pfizer, and Fresenius. GW declares consulting services for Synagen. TL declares consulting services for StratifAI. The remaining authors have no competing interests to declare.

\section*{Funding}
JNK is supported by the German Cancer Aid (DECADE, 70115166), the German Federal Ministry of Education and Research (PEARL, 01KD2104C; CAMINO, 01EO2101; TRANSFORM LIVER, 031L0312A; TANGERINE, 01KT2302 through ERA-NET Transcan; Come2Data, 16DKZ2044A; DEEP-HCC, 031L0315A; DECIPHER-M, 01KD2420A; NextBIG, 01ZU2402A), the German Academic Exchange Service (SECAI, 57616814), the German Federal Joint Committee (TransplantKI, 01VSF21048), the European Union’s Horizon Europe research and innovation programme (ODELIA, 101057091; GENIAL, 101096312), the European Research Council (ERC; NADIR, 101114631), the National Institutes of Health (EPICO, R01 CA263318) and the National Institute for Health and Care Research (NIHR, NIHR203331) Leeds Biomedical Research Centre. The views expressed are those of the author(s) and not necessarily those of the NHS, the NIHR or the Department of Health and Social Care. This work was funded by the European Union. Views and opinions expressed are however those of the author(s) only and do not necessarily reflect those of the European Union. Neither the European Union nor the granting authority can be held responsible for them.

\end{document}